\documentclass[letterpaper, 10 pt, conference]{ieeeconf}  %

\IEEEoverridecommandlockouts                              %

\usepackage{graphics} %
\usepackage{epsfig} %
\usepackage{mathptmx} %
\usepackage{times} %
\usepackage{amsmath} %
\usepackage{amssymb}  %
\usepackage[colorinlistoftodos,prependcaption,textsize=tiny]{todonotes}
\usepackage{hyperref}

\usepackage{xcolor}%
\usepackage{colortbl}
\usepackage{tikz,array,collcell}
\usepackage{pifont}
\usepackage{ragged2e}
\usepackage{booktabs}
\usepackage{xifthen}
\usepackage[nomessages]{fp}%

\usepackage{float}
\usepackage{todonotes}
\usepackage{mathtools,nccmath}

\usepackage{xtab}
\usepackage{etoolbox}
\usepackage[binary-units=true]{siunitx}
\robustify\bfseries

\newif\ifextended
\extendedtrue   %

\newcommand{\websiteurl}{\url{https://robot-motion.github.io/mpb/}}
\newcommand{\codeurl}{\url{https://github.com/robot-motion/mpb}}

\def \setReal {\mathbb{R}}
\def \setReach{\boldsymbol{\mathfrak{R}}_{\mathrm{dist}}}

\def \state {\mathbf{x}}
\def \setState {\boldsymbol{\mathfrak{X}}}
\def \setFree {\boldsymbol{\mathfrak{X}}_\mathrm{free}}
\def \setObs {\boldsymbol{\mathfrak{X}}_\mathrm{obs}}

\def \setControl  {\boldsymbol{\mathfrak{U}}}
\def \control {\mathbf{u}}
\def \start {\mathbf{x}_\mathrm{start}}
\def \goal {\mathbf{x}_\mathrm{goal}}

\def \traj {\sigma(t)}

\def \costFunction {c}

\def \traj {\sigma}

\title{\LARGE \bf
Experimental Comparison of Global Motion Planning Algorithms\\ for Wheeled Mobile Robots
}

\author{Eric Heiden${}^*{}^1$\thanks{${}^*$ Equal contribution}, Luigi Palmieri${}^*{}^{2}$, Kai O. Arras$^{2}$, Gaurav S. Sukhatme$^{2}$ and Sven Koenig$^{1}$%
\thanks{$^{1}$E. Heiden, G. S. Sukhatme, S. Koenig are with the Department of Computer Science, University of Southern California, Los Angeles, USA
        {\tt\small \{heiden, gaurav, skoenig\}@usc.edu.}}%
\thanks{$^{2}$L. Palmieri and K. O. Arras are with Robert Bosch GmbH, Corporate Research, Stuttgart, Germany
        {\tt\small \{Luigi.Palmieri, KaiOliver.Arras\}@de.bosch.com}}%
}

\newcommand{\maxnum}{1.00}
\newlength{\maxlen}
\newcommand{\databar}[2][orange!30]{%
  \addtolength{\maxlen}{\dimexpr2\tabcolsep-\arrayrulewidth}%
  \FPeval\result{round(#2/\maxnum:4)}%
  \rlap{\color{black!7}\hspace*{\dimexpr-\tabcolsep+.5\arrayrulewidth}%
  \rule[-.2\ht\strutbox]{\maxlen}{1.2\ht\strutbox}}%
  \rlap{\color{#1}\hspace*{\dimexpr-\tabcolsep+.5\arrayrulewidth}%
  \rule[-.2\ht\strutbox]{\result\maxlen}{1.2\ht\strutbox}}%
}
\newcommand{\databartwo}[2]{%
\setlength{\maxlen}{1.2\maxlen}
  \makebox[0pt][l]{\color{black!7}\hspace*{\dimexpr-\tabcolsep+.5\arrayrulewidth}%
  \rule[-.2\ht\strutbox]{\maxlen}{1.2\ht\strutbox}}%
  \makebox[0pt][l]{\color{orange!30}\hspace*{\dimexpr-\tabcolsep+.5\arrayrulewidth}%
  \FPeval\result{round(#1/\maxnum:4)}%
  \rule[-.2\ht\strutbox]{\result\maxlen}{1.2\ht\strutbox}}%
  \makebox[0pt][l]{\color{green!70!black!30}\hspace*{\dimexpr-\tabcolsep+.5\arrayrulewidth}%
  \FPeval\result{round(#2/\maxnum:4)}%
  \rule[-.2\ht\strutbox]{\result\maxlen}{1.2\ht\strutbox}}%
}

\newcommand\mcc[1]{\multicolumn{1}{>{\Centering\bf}p{\dimexpr1\maxlen}}{#1}}
\newcommand\T{\rule{0pt}{2.6ex}}       %
\newcommand\B{\rule[-1.2ex]{0pt}{0pt}} %
\newcommand\rowlabel[1]{\multicolumn{6}{>{}l}{#1}\T\B}

\begin{document}

\maketitle
\thispagestyle{empty}
\pagestyle{empty}

\begin{abstract}

Planning smooth and energy-efficient motions for wheeled mobile robots is a central task for applications ranging from autonomous driving to service and intralogistic robotics. Over the past decades, a wide variety of motion planners, steer functions and path-improvement techniques have been proposed for such non-holonomic systems. With the objective of comparing this large assortment of state-of-the-art motion-planning techniques, we introduce a novel open-source motion-planning benchmark for wheeled mobile robots, whose scenarios resemble real-world applications (such as navigating warehouses, moving in cluttered cities or parking), and propose metrics for planning efficiency and path quality. Our benchmark is easy to use and extend, and thus allows practitioners and researchers to evaluate new motion-planning algorithms, scenarios and metrics easily. We use our benchmark to highlight the strengths and weaknesses of several common state-of-the-art motion planners and provide recommendations on when they should be used.
\end{abstract}

\section{Introduction}

Motion planning is a central component in the application of mobile robots to various important real-world domains, such as autonomous driving, warehouse logistics, and service robotics \cite{paden2016survey}. Besides finding complex paths in obstacle-rich environments, they need to account for the kinodynamic constraints that a wheeled system enforces on the state space.

In particular, in this work, we focus on global motion planning algorithms that find paths in large, cluttered and complex environments, often by considering only static or semi-static information of the environment and an approximate robot dynamics model.

Over the years, various motion planning algorithms, steer functions, and path improvement (so-called post-smoothing) methods have been introduced, while the interest in autonomous robots navigating complex spaces is ever increasing.
To investigate the current state of the art in motion planning for wheeled mobile robots, in this work, we establish a benchmarking framework that is tailored toward these kinds of kinodynamic systems and their application in real-world scenarios.

\begin{figure}
    \centering
    \includegraphics[width=\columnwidth]{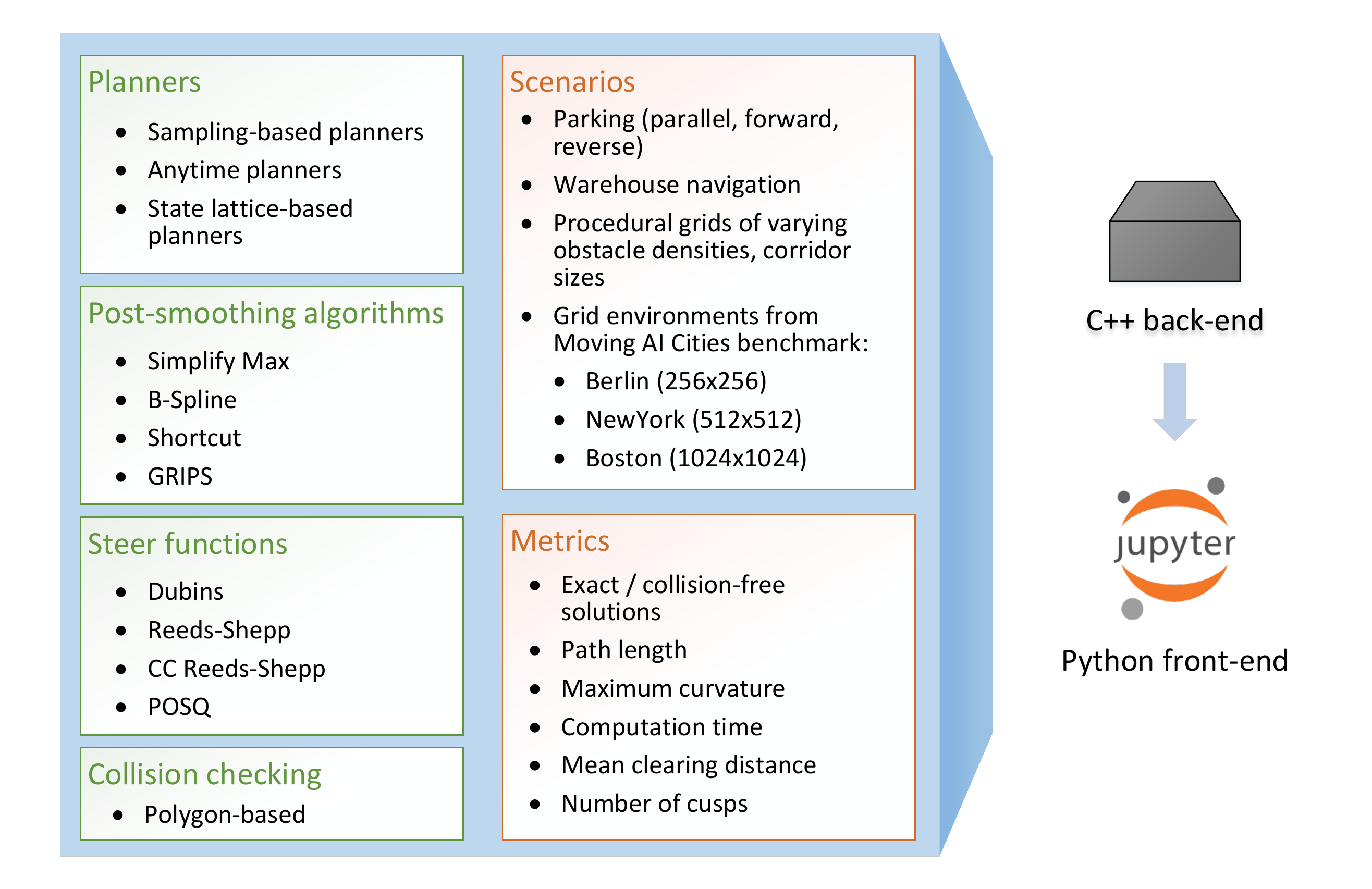}
    \caption{Architecture of the proposed motion-planning benchmarking framework. The components necessary for motion planning are shown in the box on the left (green), and the ingredients used in the evaluation are shown in the box on the right (red). The implementation is split into a C++ back-end for running the resource-intensive motion-planning components, and a Python front-end for providing a flexible interface to the design and evaluation of the benchmarking scenarios through Jupyter notebooks.}
    \label{fig:architecture}
\end{figure}

As shown in \autoref{fig:architecture}, our benchmarking is based on the following ingredients: motion planners, post-smoothing methods, steer functions\footnote{Often in literature denoted also as steering or extend function.} and collision checkers.
The combination of these building blocks is then evaluated in a variety of scenarios (environments with start and goal configurations) along various metrics.

In a typical experiment, the scenario determines the environment and the start and goal configuration. A motion planner is instantiated with a defined steer function to connect two vertices during the search while ensuring kinodynamic feasibility. Throughout the planning phase, a collision checker validates the currently considered solution with regards to the shape of the robot and the obstacles in the environment. After the planner has found a feasible solution, it can optionally be improved through post-smoothing algorithms that modify the path in order to reduce path length and curvature. 

Each of these ingredients are chosen carefully to ensure they match our application constraints. For example, we focus solely on polygon-based collision checking that places an additional burden on planners that make heavy use of the state validation test. In other experiments, we investigate how post-smoothing methods can benefit the efficiency of the planning framework and show that in some cases the fast solutions found by feasible sampling-based motion planners \cite{lavalle1998rapidly,kavraki1996probabilistic} can be smoothed in a way that they outperform their slower, albeit asymptotically optimal \cite{karaman2011sampling}, anytime motion planning counterparts. Drawing from conclusions of these findings, we give recommendations on the combination of the considered motion-planning components for particular application areas.

While much of our benchmarking software largely builds on the Open Motion Planning Library (OMPL)~\cite{sucan2012the-open-motion-planning-library}, we provide interfaces to implementations of planners (such as SBPL planners and Theta$^*$) and steer functions (POSQ and continuous-curvature steering) outside of OMPL. This enables us to get a more comprehensive picture of the current progress in motion planning for wheeled mobile robots, while being more agnostic to particular implementations of the building blocks.

\section{Related Work}
Several benchmarks have been proposed recently for analyzing the performance of different planning algorithms for a large variety of robotic systems \cite{althoff2017commonroad, moll2015benchmarking, cohen2012generic, luo2014empirical, sturtevant2012benchmarks, parasolBenchmarkinWebsite, pathBench}. All of them are generic and do not deeply analyse, as in our case, algorithms' planning performance for the specific case of wheeled mobile robots. Following we detail the most prominent ones. 

The Pathfinding Benchmarks~\cite{sturtevant2012benchmarks} from the Moving AI lab are designed for 2D path finders which consider no kinematic constraints of the robotics system. The benchmark offers a large set of scenarios (start, goal positions with length of optimal 2D path) for each grid map (of different nature i.e. mazes, cities). Contrarily to the latter, our benchmark, which uses some of its maps, considers different nonholonomic systems and different evaluation metrics rather than only path length.

Althoff et al.~\cite{althoff2017commonroad} propose a composable benchmark for motion planning and control on roads, specific for cars autonomously driving on road networks' lanes. Differently our benchmark focuses its attention on static open spaces (indoor and outdoor) where a planner finds a global path for maneuvering an autonomous system (i.e. differential drive robot or car).

Moll et al.~\cite{moll2015benchmarking} introduce a generic benchmarking tool for motion planning algorithms highly coupled with OMPL~\cite{sucan2012the-open-motion-planning-library}. 
This benchmark suite is highly customizable (it is straightforward to integrate novel collision checkers or sampling-based planners) but lacks of specific benchmark scenarios for mobile robotics applications. On the contrary, we propose a benchmark that contains a set of scenarios, problems to solve and metrics specific for mobile robotics settings. Also \cite{parasolBenchmarkinWebsite} collects a set of classical benchmarks (e.g. alpha puzzle, bug-trap) for different systems, but the benchmark offers a relatively small amount of examples for wheeled mobile robots.

Similarly to the benchmark presented by Cohen et al.~\cite{cohen2012generic} and differently from~\cite{moll2015benchmarking}, our approach enables researchers and practitioners to test different classes of motion planners, i.e., sampling-based planners (e.g. RRT$^*$, PRM$^*$~\cite{karaman2011sampling}, RRT~\cite{lavalle2001randomized}), discrete-search approaches (e.g. A$^*$~\cite{hart1968formal}, Theta$^*$~\cite{nash2007theta}), state-lattice based planners (e.g. ARA$^*$~\cite{likhachev2004ara}, ANA$^*$~\cite{van2011anytime}).
Differently from both of them, we provide several definitions of publicly-available steering functions for wheeled mobile robots (e.g. POSQ~\cite{palmieri2014novel}, Continuous Curvature~\cite{fraichard2004reeds}, Reeds-Shepp~\cite{reeds1990optimal}, and Dubins~\cite{dubins1957curves}). 

Luo et al.~\cite{luo2014empirical} introduce a benchmark on asymptotically optimal planners. These are compared on four different environments, with a single pair of predefined start and goal poses. The study considers only straight line connections (no particular kinematics or nonholonomic constraints). In this work, we propose a benchmark with a much larger selection of diverse environments, and consider different nonholonomic constraints.

Several other works~\cite{Calisi2009, weisz2016robobench, rano2006steps, sprunk2016experimental} have presented approaches to benchmark motion planning algorithms of robots moving in dynamic environments. Our work focuses its attention to planning considering a current static description of the environment: a fundamental single planning step performed during robot navigation in dynamic environments.

\section{Approach}
\label{sec:approach}
In this paper, we benchmark global motion planning algorithms commonly used for wheeled mobile robots, and provide general recommendations on the usage of these methods, considering their combination with post-smoothing methods and various steer functions. 
Our benchmark is based on two fundamental pillars: the components involved in motion planning and the evaluation procedures (shown in the box on the left and right, respectively, in \autoref{fig:architecture}). 
In particular, evaluating the performance of a motion planning algorithm requires selecting the appropriate testing environments (e.g., considering different types of map representations) and metrics (related to planning efficiency and quality of the results). We carefully selected these components by considering their scientific impact, and their recognition and popularity in the open source community \cite{sucan2012the-open-motion-planning-library,likhachev2005anytime,fraichard2004reeds}. Our choices are thoroughly presented in Sections~\ref{sec:planning_components}-\ref{sec:evaluation} and will be used to solve the following motion planning problems.

\subsection{Motion Planning Problem}
Let $\setState \subset \setReal^D$ be a manifold defining a configuration space, $\setControl \subset \setReal^M$ the symmetric control space, $\setObs \subset \setState$ the obstacle space and $\setFree = \setState \setminus \setObs$ the free space. A wheeled mobile robot can be described by an ordinary differential equation denoting a driftless control-affine system \cite{laumond1998guidelines}:
\begin{equation}
\label{eq:smoothFunction}
\dot{\state}(t)= \sum_{j=1}^{M}g_j\left(\state(t)\right)\,\control(t)
\end{equation}
where $\state(t) \in \setState $ is the state of the system, $\control(t) \in \setControl$ the control applied to it, for all $t$, and $g_1, \ldots, g_M$ are the system vector fields on $\setState$. 

Let $\gamma$ denote a planning query, defined by its initial state $\start \in \setState$ and goal state $\goal \in \setState$. We define the set of all possible 
solution paths for a given query $\gamma$ as $\Sigma_\gamma$, with $\sigma \in \Sigma_\gamma : [0,1] \to \setFree$ being one of the possible solutions such that $\sigma(0)=\start$ and $\sigma(1)=\goal$. The arc-length of a path $\sigma$ is defined by $l(\sigma) = \int_{0}^{1} ||\dot{\sigma}(t)||_2~dt$. The 
arc-length induces a \emph{sub-Riemannian} distance $\mathrm{dist}$ on $\setState$: $\mathrm{dist}(\state,\mathbf{z})=\inf_\sigma l(\sigma)$, i.e., the length 
of the optimal path connecting $\state$ to $\mathbf{z}$, which due to our assumptions is also symmetric. Let $\sigma^*$ denote the set of all points along a path $\sigma$. The $\mathrm{dist}$-clearance of a path $\sigma$ is defined as 
\begin{equation}
\delta_\mathrm{dist}(\sigma)=\sup \big\{r\in \setReal~|~
\setReach({\state}, r)\subseteq \setFree
~\forall {\state}\in \sigma^*\big\}
\end{equation}
where $\setReach({\state}, r)$ is the cost-limited reachable set for the system in Eq.~\ref{eq:smoothFunction} centered at ${\state}$ within a path length of $r$ (e.g., a sphere for Euclidean systems):
\begin{equation}
\label{eq:reachableset}
\setReach(\state, r) = \big\{\mathbf{z} \in \setState~|~\mathrm{dist}(\state, \mathbf{z}) \leq r\big\}.
\end{equation}
The $\mathrm{dist}$-clearance of a query $\gamma$ is defined as 
\begin{equation}
\delta_\mathrm{dist}(\gamma)=\sup \big\{\delta_\mathrm{dist}(\sigma)~|~\sigma \in 
\Sigma_\gamma\big\}
\end{equation}
and denotes the maximum clearance that a solution path to a query can have.
A planning algorithm solves the following $\hat{\delta}_\mathrm{dist}$-robustly feasible motion planning problem $\mathcal{P}$: given a query $\hat{\gamma}$ with a $\mathrm{dist}$-clearance of $\delta_\mathrm{dist}(\hat{\gamma})>\hat{\delta}_\mathrm{dist}$,  find a control $\control(t) \in \setControl$ with domain $[0,1]$ such 
that the unique trajectory $\traj$ satisfying \autoref{eq:smoothFunction} is fully contained in the free space $\setFree \subseteq \setState$ and connects $\start$ to $\goal$. 
Moreover, in case of (asymptotically) optimal planning, the planner minimizes (as the number of samples goes to infinity) a defined cost function $\costFunction:\Sigma_\gamma\rightarrow\setReal_{\geq 0}$. Hereinafter, we will use the term \emph{steer} function to indicate a function that generates a path in $\setState$ connecting two specified states.

\section{Planning Components}
\label{sec:planning_components}
In this section we detail the ingredients used in our benchmarking framework, i.e., motion planners, post-smoothing methods, collision checkers, and steer functions (see \autoref{fig:architecture}).

\subsection{Motion Planners}
\label{sec:motionplanners}
We compare a variety of planners belonging to four different families, namely \emph{feasible sampling-based motion planners}, \emph{any-angle path planners}, \emph{anytime or asymptotically optimal motion planners} and \emph{state-lattice-based planners}\footnote{For the sake of brevity we leave out detailed explanations of the planning algorithms and direct the reader to the corresponding references.}. We choose the most prominent open-source available planners for each class. 

\subsubsection{Feasible Sampling-based Motion Planners}
\label{sec:feasiblemotionplanners}
To this class of planners belong all the planners that are only probabilistically complete, i.e., the planner will find a path with a probability of one if the number of samples goes to infinity. We adopt the following ones from the OMPL library: RRT~\cite{lavalle2001randomized}, Stable Sparse RRT (SST)~\cite{li2016asymptotically}, EST~\cite{hsu1997path}, SBL~\cite{balakirsky2010single}, PDST~\cite{ladd2004fast}, PRM~\cite{kavraki1996probabilistic}, SPARS~\cite{dobson2013sparse}, SPARS2~\cite{dobson2013improving}. For all of them we use a uniform distribution with goal biasing, we plan in future to extend it also to deterministic sampling approaches \cite{janson2018deterministic, yershova2004deterministic, palmieribrunsRAL2019}. 

\subsubsection{Anytime or Asymptotically Optimal Sampling-based Motion Planners}
\label{sec:anytimeoptimalplanners}
In contrast to the planners from Sec.~\ref{sec:feasiblemotionplanners}, anytime, or optimal, sampling-based motion planners are asymptotically optimal planners (i.e. the probability of finding an optimal solution approaches one as the number of samples increases to infinity). This category includes the following planners from OMPL: RRT${}^*$~\cite{karaman2011sampling}, Informed RRT$^*$~\cite{gammell2014informed}, SORRT$^*$~\cite{gammell2018informed}, BIT$^*$~\cite{gammell2015batch}, RRT$\#$~\cite{arslan2013use}, BFMT$^*$~\cite{starek2015asymptotically}, PRM$^*$~\cite{karaman2011sampling}, and CForest~\cite{otte2013c}.
In this class, we additionally include planners that perform \emph{informed} search (Informed RRT$^*$, SORRT$^*$, BIT$^*$) or use multiple trees in parallel (CForest). We configure these algorithms to sample from a uniform distribution with goal biasing.

\subsubsection{Any-Angle Path Planners}
\label{sec:anyangleplanners}
In contrast to classical grid-based path finding approaches, such as A${}^*$, any-angle planners do not constrain their solutions to grid edges. Due to their advantageous smoothness and planning efficiency, we choose this class of planners, instead of classical path planners on the grid. In particular, in our benchmarking, we adopt the algorithm Theta${}^*$~\cite{nash2007theta} by also considering connections between grid points (thus samples from the configuration space) generated by a steer function. To enable Theta${}^*$ to use steer functions, we leverage the approach presented in~\cite{palmieri2016rrt}. 

\subsubsection{State-Lattice-based Planners}
\label{sec:statelatticeplanners}
In this benchmark, we include state-lattice-based planners that use deterministic sampling. In particular, they approximate the configuration by using a state lattice generated through a forward approach~\cite{pivtoraiko2009differentially}. We make use of the SBPL library~\cite{likhachev2004ara} with the following planners: ARA$^*$~\cite{likhachev2004ara}, AD$^*$~\cite{likhachev2005anytime}, MHA$^*$~\cite{islam2015dynamic}.

\subsection{Steer Functions}
\label{sec:steer-functions}
Throughout this benchmark we consider wheeled mobile robots with nonholonomic constraints.
Connecting two states for this class of systems is known as solving a two-point boundary value problem (2P-BVP) which is typically accomplished by a steer function \cite{laumond1998guidelines}. In the following, we introduce the steer functions used in our benchmark, namely: Dubins, Reeds-Sheep, Continuous Curvature, POSQ and motion primitives. For all of them we use the following kinematic model: $\dot{x} = v\cos{\theta}, \dot{y} = v\sin{\theta}, \dot{\theta}=\omega$, with $x, y$ being the robot Euclidean coordinates measured against a fixed world frame, $\theta$ the robot heading, $v$ its tangential velocity, and $\omega$ its angular velocity.

\subsubsection{Dubins Curves}
Dubins et al.~\cite{dubins1957curves} (DS) assume a car driving with constant speed $v=1$ (i.e. moving forward).  Its optimal paths are a combination of no more than three motion primitives (go straight (S, $\omega=0$), turn left (L, $\omega=1$), turn right (R, $\omega=-1$)). More precisely, Dubins et al. showed that optimal paths are a composition of only the following family of curves: LSR (turn left, go straight, turn right), LSL, RSR, RSL, RLR, and LRL.

\subsubsection{Reeds-Shepp}
Reeds-Shepp curves~\cite{reeds1990optimal} (RS) are an extension of Dubins steering. Besides the Dubins primitives, Reeds-Shepp curves consider a car that can also move backwards with constant speed (thus $v$ can be $-1$ or $+1$). The problem to solve is more complex than Dubins, having now 46 possibilities of composed primitives. 

\subsubsection{Continuous Curvature Steer Functions}
Due to their system definition, Reeds-Shepp and Dubins steering require the system to be stopped each time a new turn is requested. To counteract this issue, Fraichard et al.~\cite{fraichard2004reeds} propose a new class of steer functions for car-like kinematics called \emph{continuous curvature} (CC) steering functions. Differently from Reeds-Shepp and Dubins, continuous-curvature steer functions enforce continuity on the curvature $\kappa$ of the paths (extending the state to also considering the curvature). By considering the curvature, the complexity of finding an optimal path slightly increases when compared to Reeds-Shepp and Dubins. Banzhaf et al.~\cite{banzhaf2017hybrid} further extend this class by allowing the continuous-curvature functions to fall back to the Reeds-Shepp family, thus having curvature discontinuities at switches in the driving direction, a useful property when operating in very cluttered environments (i.e., the car is allowed to turn the steering wheel while not moving). As representative of the continuous-curvature steer functions, we include continuous-curvature Reeds-Shepp steering (referred to as CC Reeds-Shepp) in this benchmark.

\subsubsection{POSQ}
In~\cite{palmieri2014novel}, the authors exponentially solve the 2P-BVP for the kinematic car-like system by extending the discontinuous control approach developed by Astolfi et al. \cite{astolfi1999exponential}. The approach, unlike Dubins or Reeds-Shepp, does not produce optimal paths, but it was nonetheless shown to produce \emph{smooth} paths. Moreover it does not consider constant velocities, thus allowing the robot to move more freely in cluttered environments.

\subsubsection{Motion Primitives}
State lattice planning uses a set of precomputed motion primitives (pairs of $v,\omega$) instead of steer functions. Hence, the approach does not fully solve the 2P-BVP. In this benchmark, we use a forward-propagation approach and use unicycle motion primitives for all the SBPL planners, which are available from the SBPL repository. %

\subsection{Post-smoothing Methods}
Besides the planners, in this benchmark, we take algorithms for path improvement into consideration. We adopt existing post-smoothing methods from the OMPL library, namely the B-Spline, Shortcut and Simplify Max algorithms~\cite{sucan2012the-open-motion-planning-library}. In addition, we compare them against the recently introduced gradient-informed post smoothing (GRIPS) algorithm~\cite{heiden2018gradient}, a hybrid approach that uses short-cutting and locally optimizes vertexes placement.

\subsection{Collision Checking}
\begin{figure}
    \centering
    \includegraphics[width=.5\columnwidth]{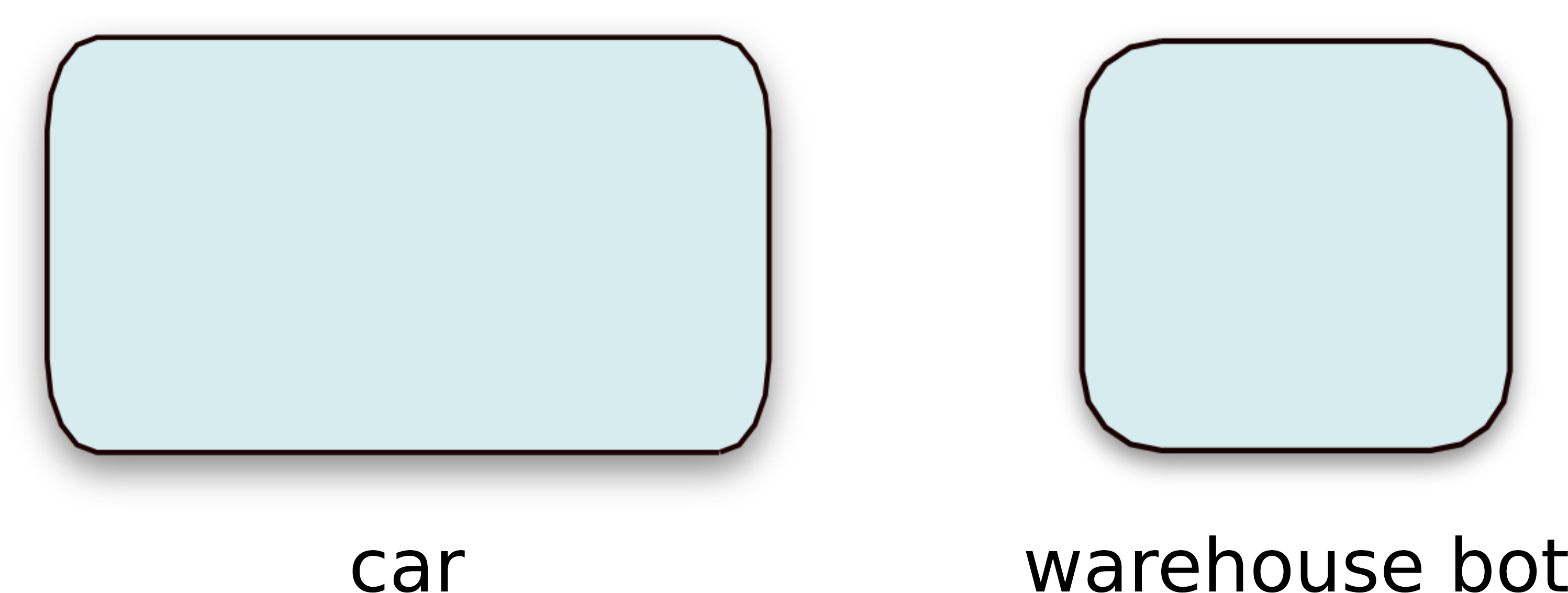}
    \caption{The two different polygon-based collision models used throughout this benchmark.}
    \label{fig:collision_models}
\end{figure}

Throughout all our experiments, we use two-dimensional polygon-based collision models (see \autoref{fig:collision_models}) where the robot is represented by a convex shape. Based on the \emph{Separating Axis Theorem}~\cite{gottschalk1996sat}, we check for intersections between the robot and the obstacle polygons. Compared to testing for point collision, this model is significantly more demanding to evaluate, resulting in larger computation times for state validation checks. However, since we are interested in motion planners that are relevant to mobile robots, such collision models need to be taken into account.

\section{Evaluation}
\label{sec:evaluation}
In this section, we describe the set of experiments, environments and the metrics used to evaluate the planners and post-smoothing methods in terms of planning efficiency and in returned path quality. The list of all experiments, pointing to the related results' sections, is reported in Table~\ref{tab:experiments}. The table collects eleven different types of experiment we run: three of them use the Moving-AI grid environments described in Section~\ref{subsec:movingaidescription}, four the procedurally generated grids detailed in Section~\ref{subsec:procedurallydescription}, and four use a polygonal representation of the environment and robot (see Section~\ref{sec:polygon-envs}). Of the latter, three study the behavior of the planners when the environment complexity change and one properties of post-smoothers and planners' combinations.

\ifextended
\begin{table}[]
    \centering
    \resizebox{\columnwidth}{!}{%
    \newcolumntype{P}[1]{>{\centering\arraybackslash}p{#1}}
    \begin{tabular}{p{2.4cm}P{1.8cm}P{1cm}p{4cm}}
        \toprule
        \bf Experiment / Section & \bf Environment & \bf Collision model & \bf Description \B\\
        \midrule
        \T
        \texttt{cross\_corridor}\newline\autoref{sec:cross-corridor} & $100\times100$ grid (procedural) & car & Evaluation on procedurally generated corridor environments with varying corridor diameters (\autoref{fig:grid_envs} bottom) \\
        \T
        \texttt{cross\_turning}\newline\autoref{sec:cross-turning} & $100\times100$ grid (procedural) & car & Evaluation on procedurally generated grid environments with varying turning radii in Reeds Shepp steering \\
        \T
        \texttt{cross\_density}\newline\autoref{sec:cross-density} & $100\times100$ grid (procedural) & car & Evaluation on procedurally generated grid environments with varying obstacle densities (\autoref{fig:grid_envs} top) \\
        \T
        \texttt{sam\_vs\_any}\newline\autoref{sec:sam-vs-any} & $150\times150$ grid (procedural) & car & Comparison of anytime planners vs. a combination of sampling-based planners and post-smoothing methods \\
        \T
        \texttt{Berlin\_0\_256}\newline\autoref{sec:moving-ai} & $256\times256$ grid (MovingAI) & car & Evaluation of the 50 hardest scenarios from the Berlin\_0\_256 MovingAI benchmark \\
        \T
        \texttt{NewYork\_1\_512}\newline\autoref{sec:moving-ai} & $512\times512$ grid (MovingAI) & car & Evaluation of the 50 hardest scenarios from the NewYork\_1\_512 MovingAI benchmark \\
        \T
        \texttt{Boston\_1\_1024}\newline\autoref{sec:moving-ai} & $1024\times1024$ grid (MovingAI) & car & Evaluation of the 50 hardest scenarios from the Boston\_1\_1024 MovingAI benchmark \\
        \T
        \texttt{parking\_1}\newline\autoref{sec:parking-results} & polygon & car & Evaluation on the polygon-based environment parking\_1 (\autoref{fig:polygon_mazes}) \\
        \T
        \texttt{parking\_2}\newline\autoref{sec:parking-results} & polygon & car & Evaluation on the polygon-based environment parking\_2 (\autoref{fig:polygon_mazes}) \\
        \T
        \texttt{parking\_3}\newline\autoref{sec:parking-results} & polygon & car & Evaluation on the polygon-based environment parking\_3 (\autoref{fig:polygon_mazes}) \\
        \texttt{warehouse}\newline\autoref{sec:warehouse-results} & polygon & warehouse bot & Evaluation on the polygon-based environment warehouse (\autoref{fig:polygon_mazes}) \\
        \bottomrule
    \end{tabular}
    }
    \caption{Overview of experiments conducted in this benchmark.}
    \label{tab:experiments}
\end{table}
\fi

\subsection{Environments}
\label{sec:environments}
In the following, we describe the two types of environments we consider throughout our benchmarking, as well as the how the \emph{scenarios} are defined, i.e. the start and goal configurations for each environment. We consider the two main classes of environmental representation used nowadays for motion planning: grids and the polygon-based ones. Section~\ref{subsec:gridenvironments} details a set of experiments based on grid representations of the obstacles, a typical approach used in robotics navigation, in particular when planning in large environments (i.e. cities, airports, train stations or large office-like environments). Polygon-based environments, described in Section~\ref{sec:polygon-envs}, are often adopted when planning in tight and small environments (i.e. parking and warehouse like environments), where the planning system should more carefully and precisely consider obstacles' geometry.

\subsubsection{Grid-based Environments}
\label{subsec:gridenvironments}
\begin{figure}
    \centering
    \includegraphics[width=\columnwidth, trim=0 1.5cm 0 1.5cm]{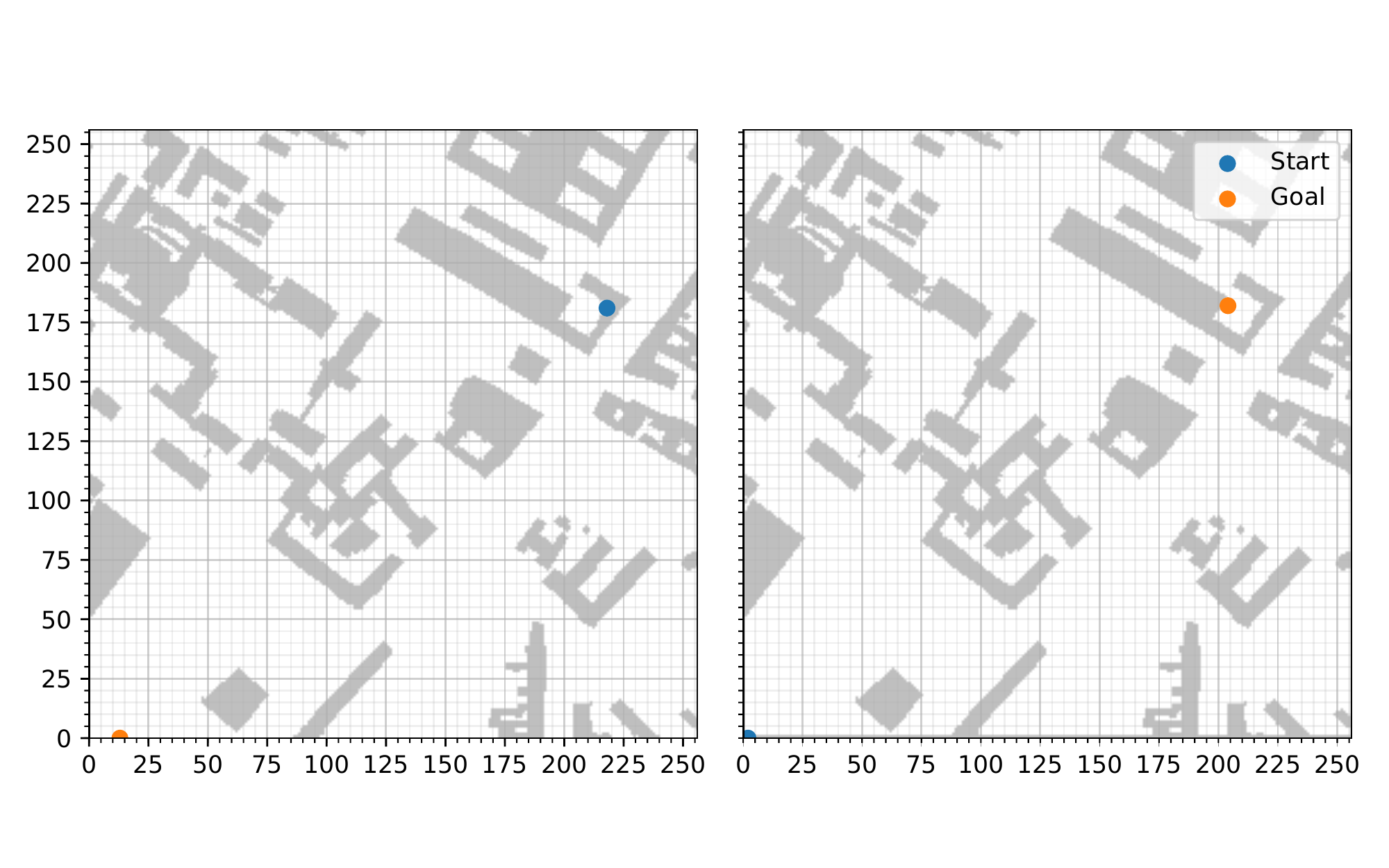}
    \caption{Many of the challenging Moving AI Cities scenarios define start and goal locations that are too close to obstacles to be solvable by a polygon-based collision model of the robot. Shown here are two scenarios from the \emph{Berlin\_0\_256} map with highlighted start and goal positions.}
    \label{fig:moving_ai_too_narrow}
\end{figure}

We design two sets of environments, a sub-selection of the grids form the Moving AI benchmark \cite{sturtevant2012benchmarks} and a set of grids procedurally generated by varying corridors' size or obstacles density.
\paragraph{Moving AI environments}
\label{subsec:movingaidescription}
The Cities Dataset from Moving AI Lab's Pathfinding Benchmarks~\cite{sturtevant2012benchmarks} contains occupancy grid maps of various cities at varying resolutions, ranging between $256\times256$ and $1024\times1024$ grid cells. The Moving AI benchmarks are equipped with scenarios for each environment, i.e., pairs of start and goal positions, sorted by difficulty (in terms of the length of the shortest path from the start to the goal). In our benchmark, for each environment considered, we select the last 50 scenarios that correspond to start and goal locations which are the most apart from each other, see an example in \autoref{fig:moving_ai_too_narrow}.

\begin{figure}
    \centering
    \includegraphics[width=\columnwidth]{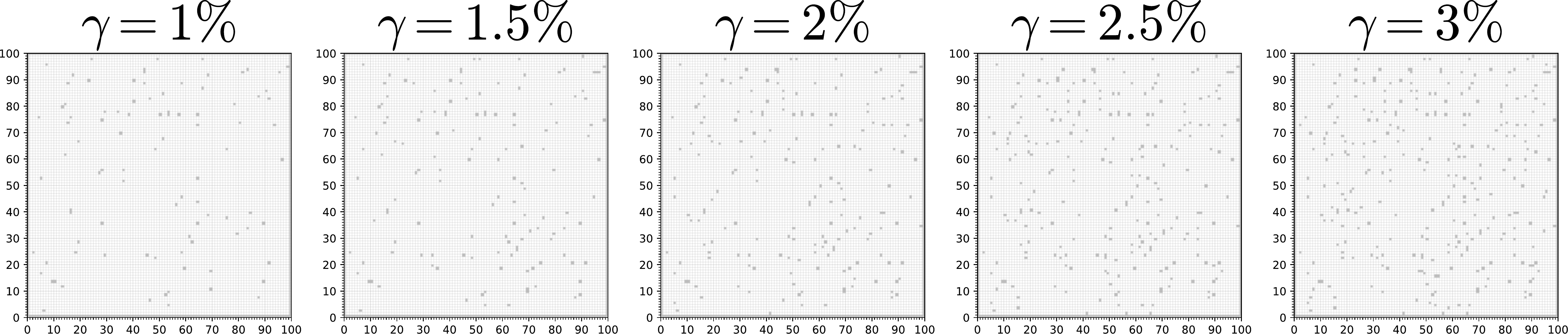}\\\vspace{.8em}
    \includegraphics[width=\columnwidth]{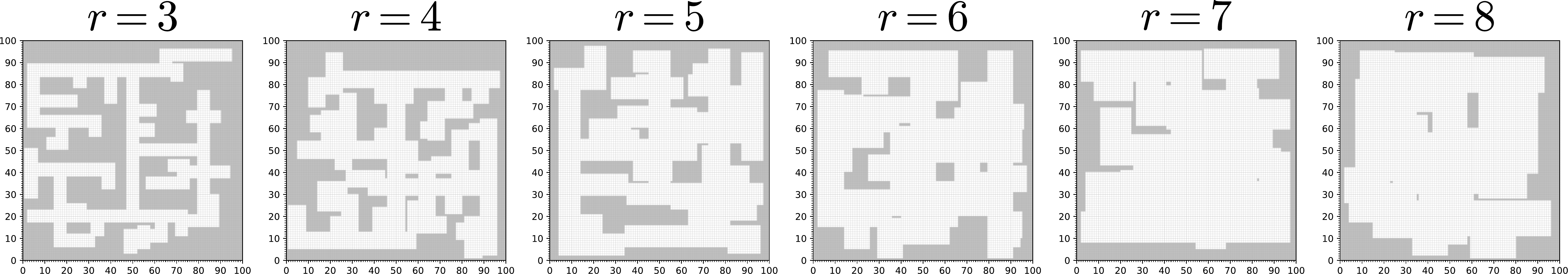}
    \caption{Examples of the procedurally generated grid environments. Top: varying obstacle ratios. Bottom: varying corridor sizes.}
    \label{fig:grid_envs}
\end{figure}

\paragraph{Procedurally-generated environments}
\label{subsec:procedurallydescription}
Besides the Moving AI environments, we generate grid mazes procedurally to investigate how the planners behave under specific conditions that influence the available free space. As shown in the bottom row of \autoref{fig:grid_envs}, the grid worlds we generate resemble typical indoor scenes with complex networks of rectangular spaces, such as rooms and corridors. We generate these environments by starting with a completely occupied grid and apply a few iterations from a modified RRT exploration that only connects the nearest tree node to the randomly sampled point via either horizontal or vertical lines of a certain width. This allows us to generate environments, with different corridor sizes (see bottom row in \autoref{fig:grid_envs}). Following from the RRT tree that generates the free space in our procedural grid environments, we select the furthest two points in the tree as the start and goal positions for each scenario.
Additionally, to compare the planners on more generic environments, we implemented environments where cells are randomly sampled from a uniform distribution and set to being occupied. This random process is repeated until a desired ratio of occupied versus free cells has been reached, as shown in the top row of \autoref{fig:grid_envs}.

\subsubsection{Polygon-based Environments}
\label{sec:polygon-envs}
Furthermore our benchmark includes environments where the obstacles are represented by convex shapes and the robot itself is analogously represented by a polygon. Scenarios of this kind come close to real-world, two-dimensional navigation scenarios where the collision checker has to take into account the geometry of the robot and its environment to evaluate the validity of state. For example, in the case of a robot represented by an elongated rectangle, the orientation angle can greatly influence whether a narrow pass in the environment can be traversed, whereas in the point-based collision model such considerations need not be made.

\begin{figure}
    \centering
    \includegraphics[width=\columnwidth]{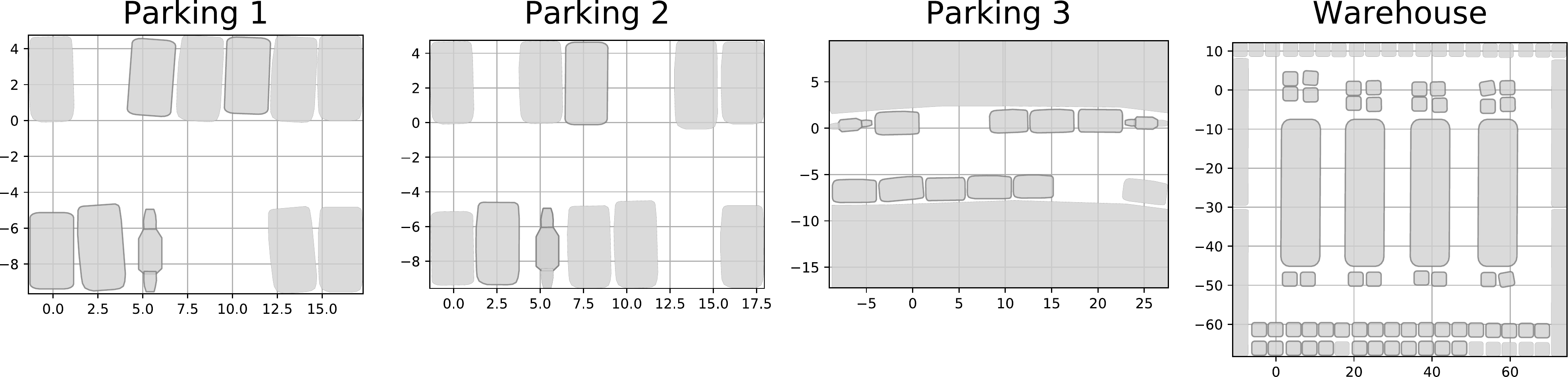}
    \caption{The four polygon-based environments where obstacles are represented by convex shapes.}
    \label{fig:polygon_mazes}
\end{figure}

\begin{figure*}
    \centering
    \includegraphics[width=.8\textwidth]{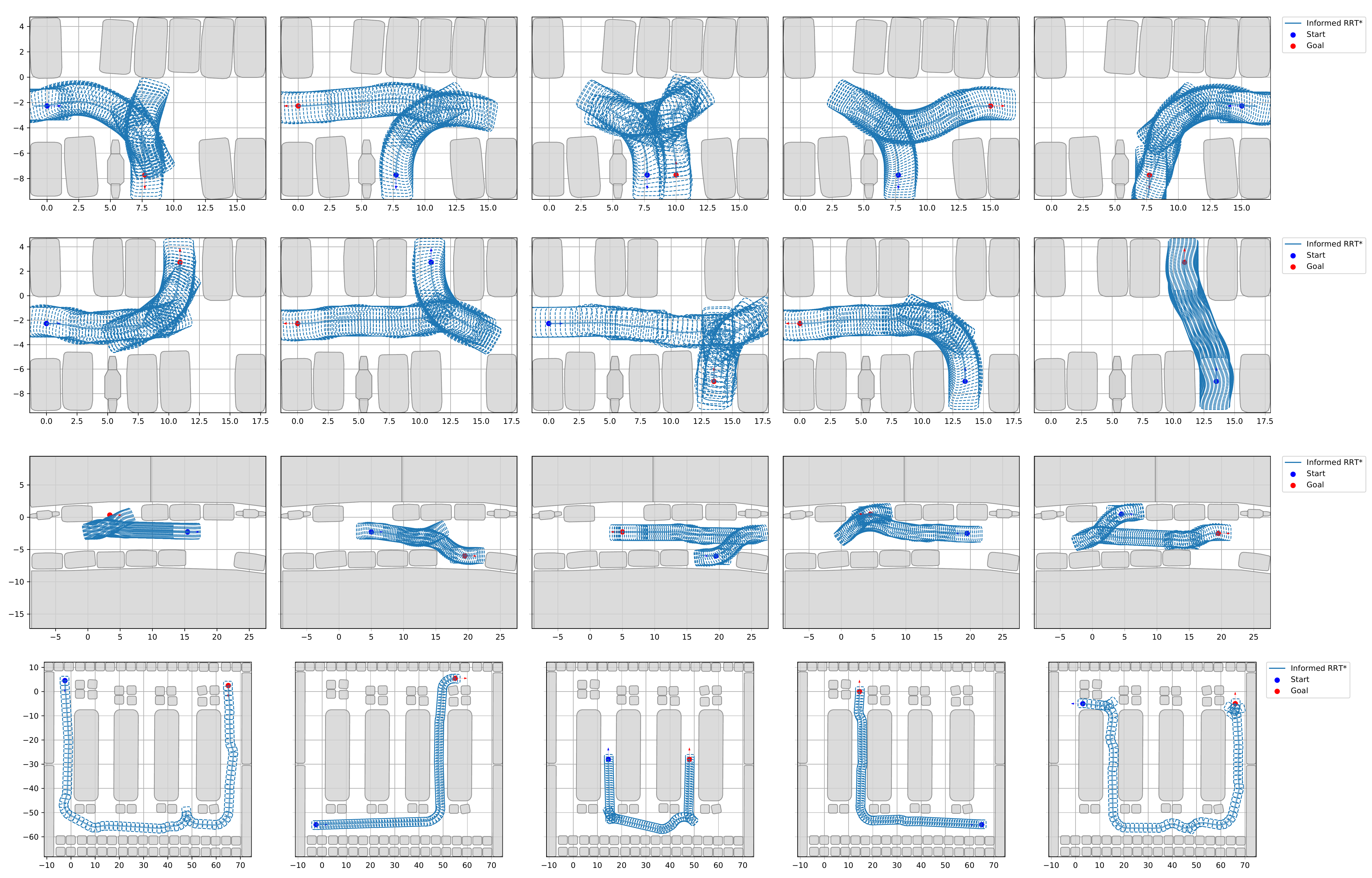}
    \caption{Exemplary results for the polygon-based environments \emph{parking1}, \emph{parking2}, \emph{parking3}, and \emph{warehouse} (from top to bottom) with all five different start/goal configurations. Each subplot shows the computed trajectories from the Informed RRT${}^*$ planner using the CC Reeds-Shepp steer function.}
    \label{fig:parking_traj}
\end{figure*}

We show the four types of polygon-based environments we designed in \autoref{fig:polygon_mazes} and with example paths in \autoref{fig:parking_traj}. We choose five start-and-goal configurations and validate them by ensuring that the planner BFMT\footnote{Through preliminary experiments, we found BFMT to be among the most reliable planning algorithms that gave high-quality solutions in short time on the polygon-based environments.} finds exact solutions using the Reeds Shepp steer function (\autoref{fig:polygon_mazes}). In the first three cases, we consider the scenario of an autonomous car-like vehicle that needs to park itself among a set of surrounding parked cars and other obstacles. We consider the three common cases of parking: (1) parking forward, (2) parking backward into a parking lot, and (3) parallel-park in a street of parked cars. In the last type of polygon-based environments (4), a complex warehouse-like environment is simulated where the robot has to navigate between shelves of various sizes and irregular orientations.

\begin{figure*}
    \centering
    \includegraphics[width=\textwidth]{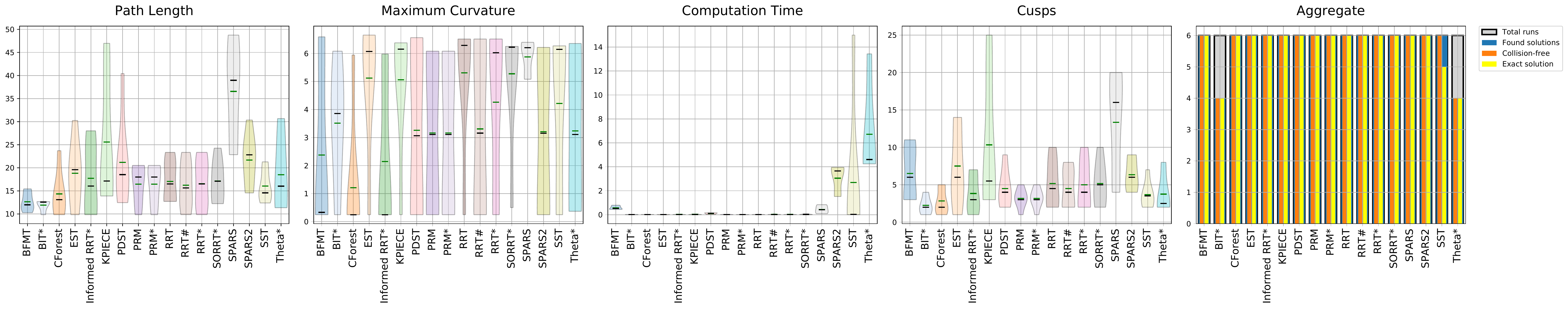}\\
    \includegraphics[width=\textwidth]{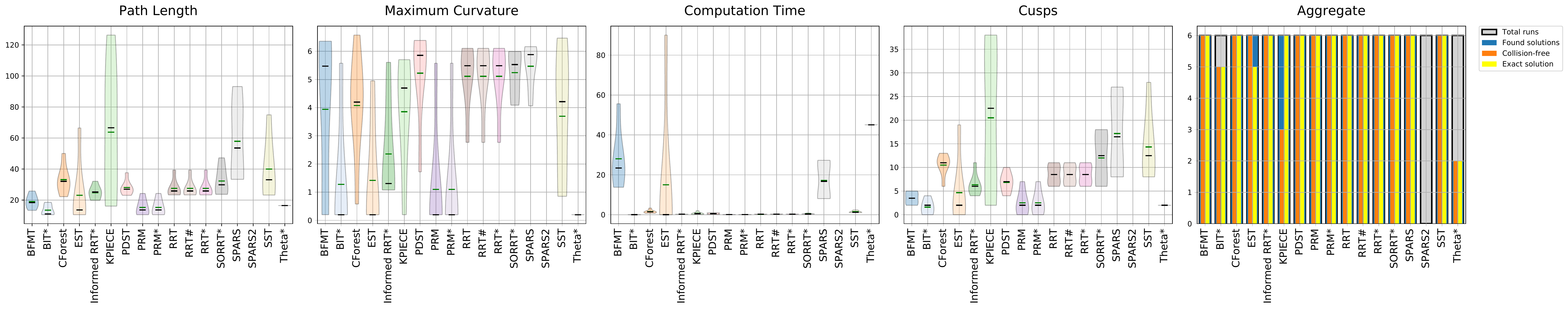}\\
    \includegraphics[width=\textwidth]{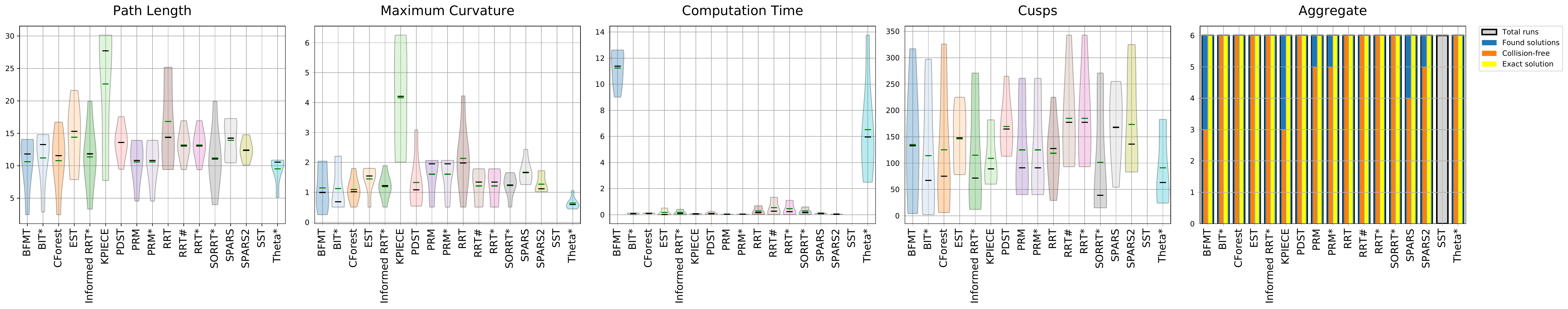}\\
    \includegraphics[width=\textwidth]{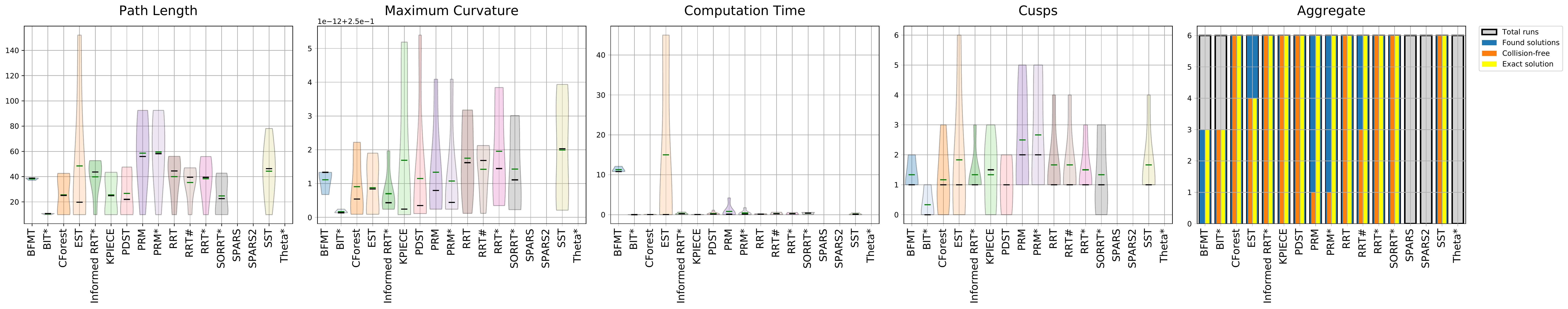}
    \caption{Statistics for the \emph{parking1} scenarios. First row: Reeds Shepp steering, second row: CC Reeds Shepp steering, third row: POSQ steering, fourth row: Dubins steering.}
    \label{fig:parking1_stats}
\end{figure*}

\subsection{Metrics}
\label{sec:metrics}
We compare the planners based on a selection of metrics relevant to wheeled mobile robotics applications, such as autonomous driving, service and intralogistic robotics. In particular, we evaluate the planners in terms of quality of the returned solutions and in planning efficiency by considering the following metrics:
\begin{itemize}
     \item \emph{Success statistics} that measure the ratio of found, collision-free, and exact
     \footnote{A trajectory is \emph{exact} if it connects the start and goal nodes.} solutions.
     \item \emph{Path length} of the obtained solution in the workspace $\mathcal{W}$. All the asymptotically optimal planners are configured to minimize path length, thus we measure how well the planners performs based on their main objective.
     \item \emph{Curvature} ($\kappa$) and \emph{Maximum curvature} (${\kappa_{\max}}$): as a way to measure the induced comfort and smoothness of the obtained paths. Keeping the maximum curvature at a low level corresponds to smoother maneuvers, therefore less control effort and energy to steer the robot.
    \item \emph{Computation time} to find the first solution.
     \item \emph{Mean clearing distance} ($\overline{\delta}_\mathrm{dist}(\gamma)$): with lower values indicating that the solutions are closer to the obstacles.
      \item \emph{Number of cusps} following~\cite{banzhaf2017hybrid}: maneuvering in difficult environments may require the robots to stop and turn the wheels in the opposite direction, thus yielding a cusp in the trajectory. Having more cusps correspond to less smooth and more difficult to drive paths.
     \end{itemize}
 
\section{Benchmark Implementation}
We develop our benchmarking system in C++ and provide a high-level front-end in Python\footnote{Our code will be made open-source at \codeurl.}. The experiments are implemented in Jupyter notebooks that leverage our Python front-end and enable the user to monitor through rich progress reports and plotting capabilities the status of the execution. We are collecting the experimental results and derived plots on our website at \websiteurl{} where the complete data can be analyzed. 

We run the benchmark on a server featuring \SI{256}{\giga\byte} RAM, two Intel Xeon Gold 6154 @ 3.00GHz CPUs offering 72 threads in total, running on Ubuntu 18.04 (kernel version 4.15.0). Each experiment is run using 20 parallel processes that correspond to different environment seeds, in the case of the procedurally generated environments. Each process runs a sequence of planners and post-smoothing methods on its predefined environment. We limit the parallelism to 20 out of 72 available CPU cores due to the fact that planners such as CForest spawn multiple threads on its own to find a solution. By further randomizing the order in which each of our benchmark processes executes the planners, we can keep the number of parallel threads in check (e.g. avoid running 20 parallel CForest instances). Each process is automatically cancelled if twice of its time limit has been exceeded (time out), or if its memory consumption has exceeded \SI{18}{\giga\byte}.

\section{Results}
\label{sec:results}

This section summarizes the results obtained in our experiments, while focusing on the main findings. The complete statistical analysis will be published on our website.

\subsection{Moving AI Scenarios}
\label{sec:moving-ai}
This section reports the results obtained of the grids selected from the Moving AI benchmark, 
\ifextended
    see Tables~\ref{tab:moving_ai_berlin}, \ref{tab:moving_ai_newyork} and \ref{tab:moving_ai_boston}.
\fi
The solution column contains two numbers separated by a `/': the second number indicates the number of solutions found (highlighted by the orange bar in the background), the first number indices how many of these solutions are collision-free. Each planner is run on a total of 51 scenarios.
The following columns indicate the planning statistics in the format mean $\pm$ standard deviation across the metrics (planning time, path length, maximum curvature and average curvature along the paths). The last column shows the total number of cusps in all solutions combined. We group these statistics by the steer functions, for which we selected different time limits, as shown in the tables next to the group labels. These time limits have been determined empirically to ensure that many solutions could be found. The SBPL planners are treated separately since they did not use any of the provided steer functions but their particular unicycle motion primitive.

\paragraph{Path Length and Smoothness}
Results are detailed in \autoref{tab:moving_ai_berlin}. In terms of path length and smoothness, anytime path planners achieve better performance within the given maximum planning time for all the steer functions. Feasible planners generate often longer and less smooth paths (higher curvature and number of cusps).
Specifically for the scenario \emph{Berlin\_0\_256}, BFMT achieves the shortest path lengths within the fastest time with average curvature, except with POSQ steering where it has poor runtime and path length. KPIECE throughout all experiments finishes among the fastest but consistently has the longest paths and among the worst maximum curvature. 
For the Dubins curves, it was considerably more difficult for the planners to find feasible solutions -- sampling-based planners, such as EST, SST, PDST and KPIECE were the most successful in finding exact, collision-free paths.
\ifextended
    In the \emph{NewYork\_1\_512} scenarios, RRT$^*$, RRT$^\#$, SORRT$^*$, Informed RRT$^*$ and CForest achieve the shortest solutions with the lowest curvature. While CForest generally finds short solutions with low computation times, few of them are collision-free. EST finds the most and shortest solutions with POSQ, although with a significant number of cusps resulting in relatively high curvature.
    \emph{Boston\_1\_1024} is the largest of the grid-based environments and requires significantly longer computation times for the majority of planners to return an exact and collision-free path. In most cases, only the feasible sampling-based motion planners, such as EST, SST and RRT manage to find valid solutions, whereas CForest, Informed RRT$^*$ and SORRT$^*$ do not find any collision-free paths with Reeds-Shepp steering within a time limit of \SI{7.5}{\minute}. BFMT finds the most collision-free solutions with Reeds-Shepp and CC Reeds-Shepp with among the shortest path length and lowest curvature (among planners which also find valid paths). With POSQ and Dubins as steer functions, however, it fails to find any (POSQ) or more than one (Dubins) solutions. 
\fi

\begin{figure*}
    \centering
    \includegraphics[height=4cm]{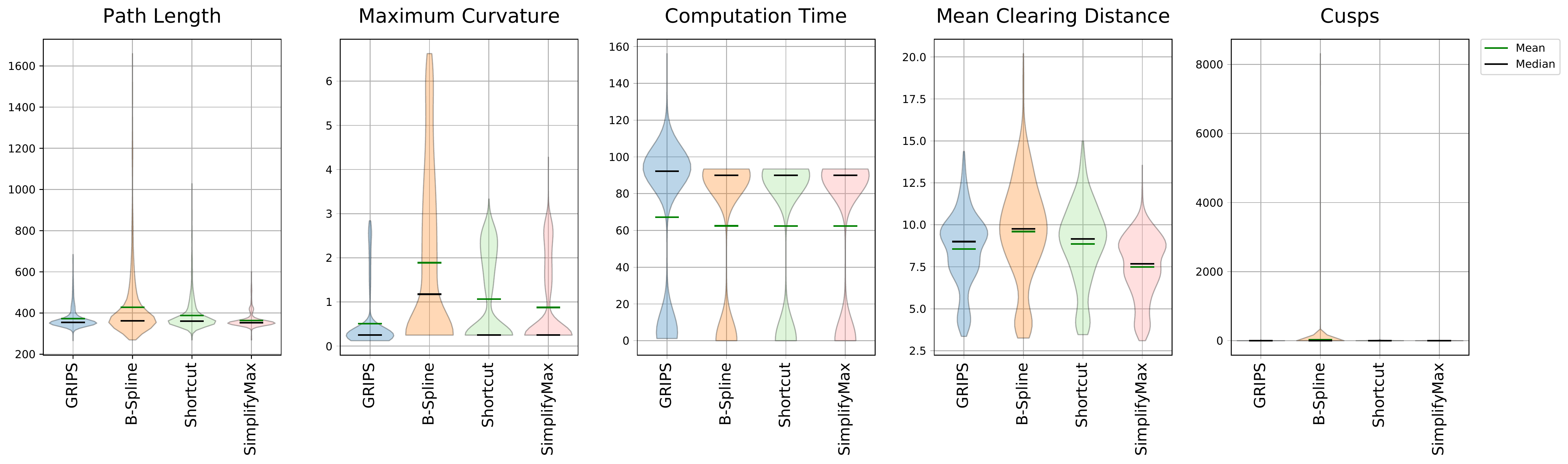}\\
    \includegraphics[height=4cm]{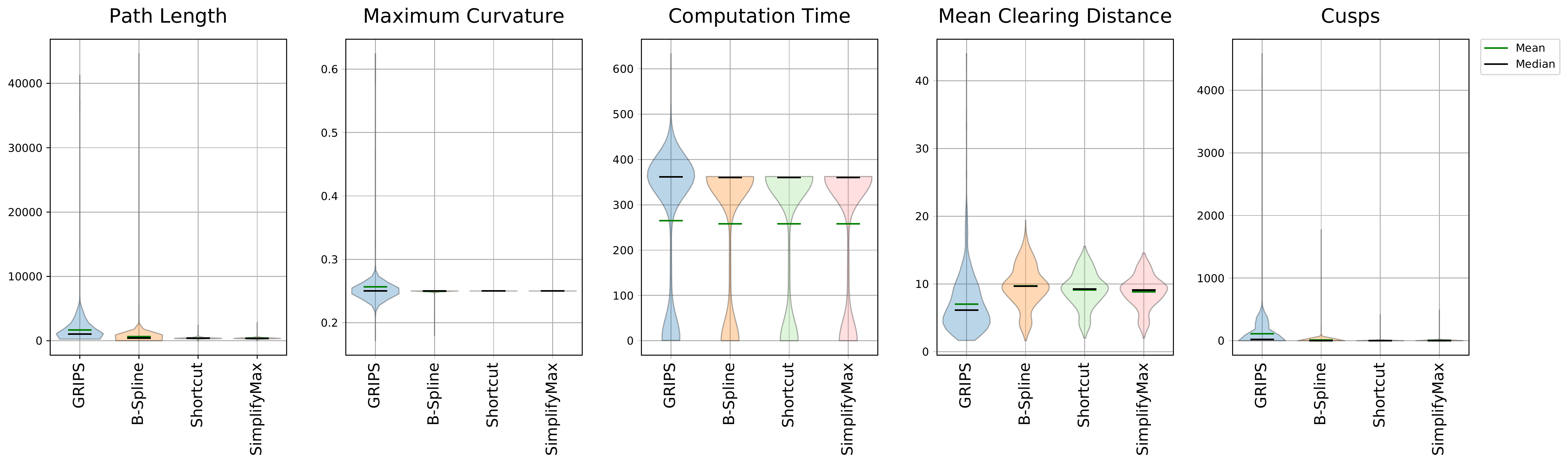}\\
    \includegraphics[height=4cm]{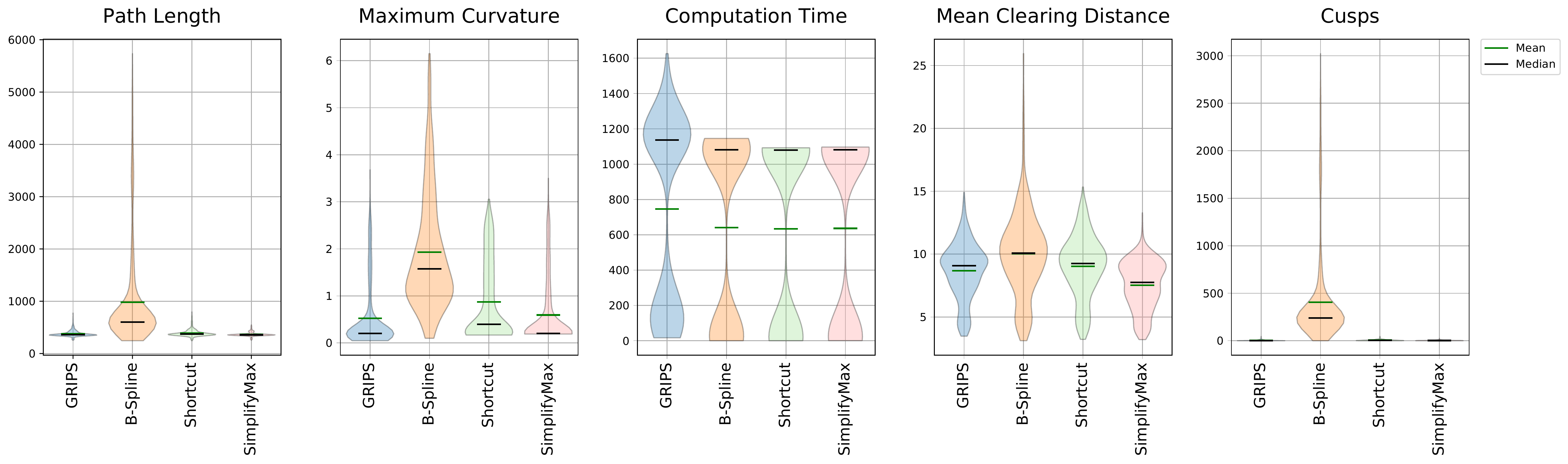}\\
    \includegraphics[height=4cm]{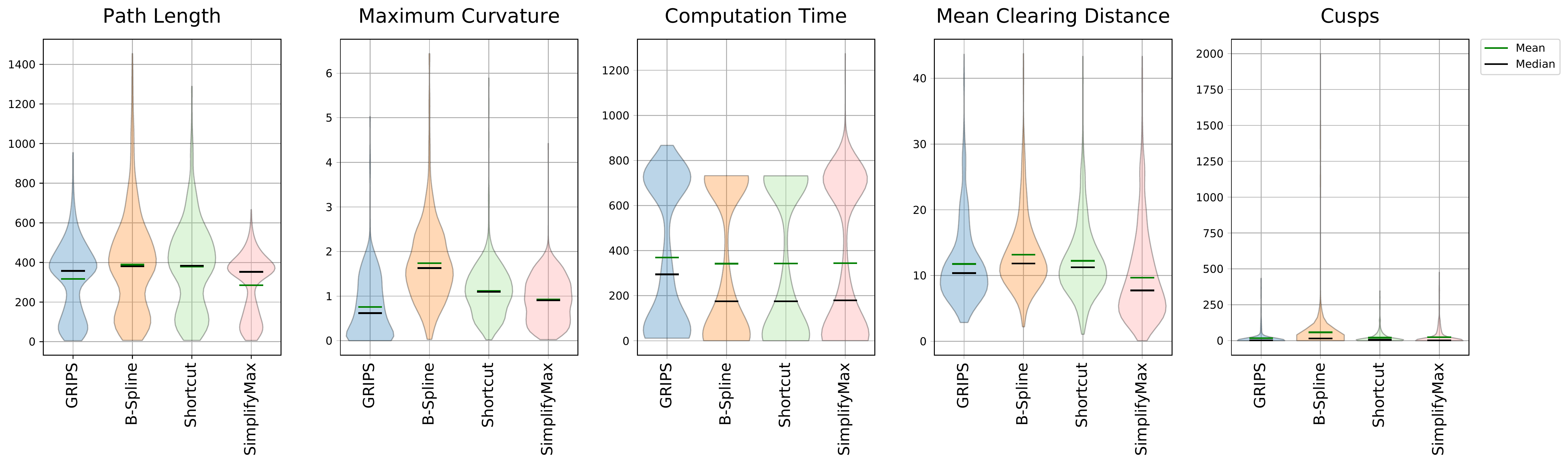}
    \caption{Planning statistics for the post-smoothing algorithms GRIPS, B-Spline, Shortcut, SimplifyMax (left to right per subplot) using different steer functions from the \emph{Berlin\_0\_256} scenario from the Moving AI benchmark. These are the 50 most difficult start-goal configurations from the benchmark.
    First row: Reeds Shepp steering, second row: Dubins steering, third row: CC Reeds Shepp steering, fourth row: POSQ steering.}
    \label{fig:smooth_berlin}
\end{figure*}

\paragraph{Post-Smoothing Results}
In \autoref{fig:smooth_berlin} we summarize the post-smoothing results across all planners in the \emph{Berlin\_0\_256} scenarios, which is representative for the other Moving AI benchmark environments. GRIPS often outperforms the other methods in maximum curvature while achieving similar path length as SimplifyMax. In computation times, B-Spline, Shortcut and SimplifyMax perform similarly, except with POSQ steering where the latter is significantly slower with a median computation time almost twice as high as the other methods. SimplifyMax yields solutions which often have very small clearing distance. B-Spline solutions have considerably more cusps than the results obtained with the other methods.

\paragraph{Theta$^*$ and SBPL Issues}
On the larger-scale environments considered throughout this benchmark (particularly the Moving AI scenarios), we noticed that our current implementation of Theta$^*$ makes heavy use of the collision checker that significantly deteriorates its computation time. As can be seen in \autoref{tab:moving_ai_berlin}, only in the case of a \SI{6}{\minute} time limit for a fast-to-evaluate steer function, such as Dubins, does this algorithm find a competitive number of collision-free, exact solutions. In other cases, our implementation does not yield a solution before the time limit is up.
Similarly, the planners AD$^*$, ARA$^*$ and MHA$^*$ from SBPL were often unable to find feasible solutions within the time limit.
\ifextended
    On the \emph{Boston\_1\_1024} scenario, MHA$^*$ found only a single solution within \SI{60}{\minute}, while non of the other SBPL planners returned any feasible path. We therefore excluded these results from \autoref{tab:moving_ai_boston}.
\fi

\subsection{Polygon-based Environments}
\label{sec:polygon_env_results}

The following scenarios are particularly tailored toward autonomous driving. Instead of navigating grid world, the environments use arbitrary convex shapes to represent obstacles.

\subsubsection{Parking scenarios}
\label{sec:parking-results}

\paragraph{Path Length and Smoothness}
Similarly to the grid-based environments, in these scenarios, anytime planners achieve better performance in terms of path length and smoothness than feasible planners, although at the price of being slower.
In the scenarios for the first parking environments, we notice that RRT, Informed RRT$^*$, RRT$^*$ and SORRT$^*$ always find solutions, across all tested steer functions, as shown in \autoref{fig:parking1_stats}. SST, Theta$^*$, SPARS and SPARS2, however, do often not find any solutions. Particularly SPARS2 is the only planner that cannot find any solutions for CC Reeds Shepp steering, SST is the only algorithm that is unable to solve any scenarios with POSQ steering. The Dubins steer function appears to be particularly challenging, as SPARS, SPARS2 and Theta$^*$ cannot find any paths, while various other planners, such as PRM, PRM$^*$, BFMT and BIT$^*$ only solve a small fraction of the scenarios exactly.
\ifextended
    We observe similar behavior on the second parking environment (\autoref{fig:parking2_stats}). The parallel parking environment (\emph{parking3}) proves more challenging (\autoref{fig:parking3_stats}) for most planners which leads to considerable less collision-free and exact solutions, particularly under the kinodynamic constraints of Dubins steering.
\fi

\subsubsection{Warehouse scenarios}
\label{sec:warehouse-results}
\begin{figure*}
    \centering
    \includegraphics[width=\textwidth]{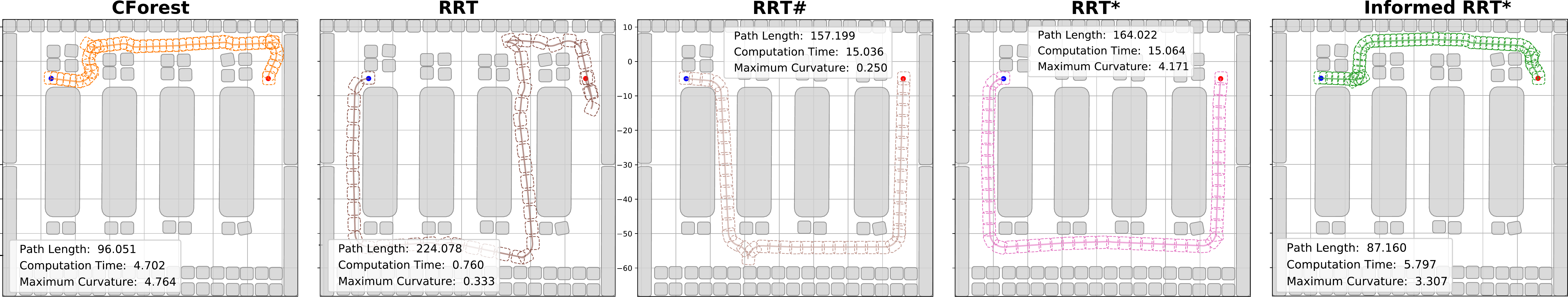}
    \caption{Example trajectories for the different planners in one of the five \emph{warehouse} scenarios with Reeds-Shepp steering.}
    \label{fig:warehouse_trajectories}
\end{figure*}

We visualize example solutions obtained from all planners on the fourth scenario from the warehouse environment with Reeds Shepp steering in \autoref{fig:warehouse_trajectories}. BFMT, CForest, Informed RRT$^*$ and SORRT$^*$ find the shortest solutions which all lie in the same homotopy class.

\begin{figure*}
    \centering
    \includegraphics[width=\textwidth]{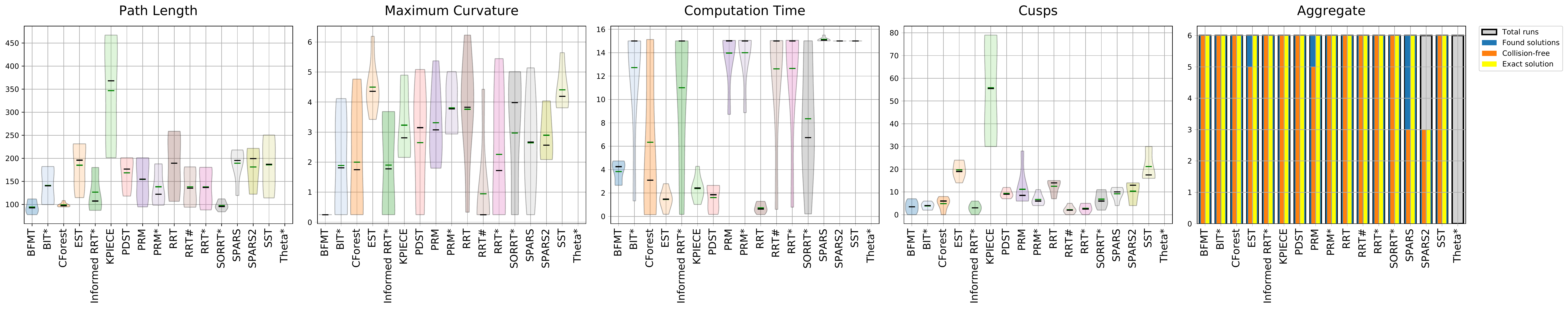}\\
    \includegraphics[width=\textwidth]{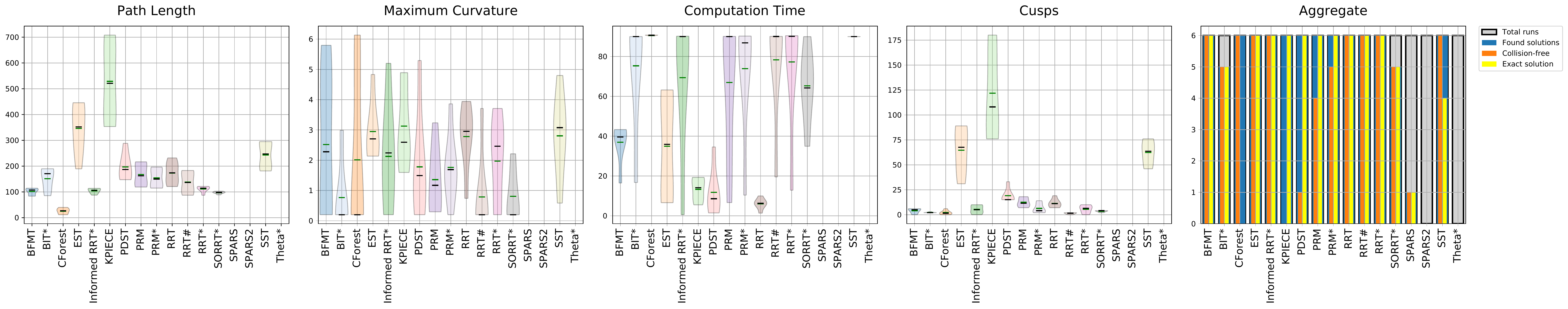}\\
    \includegraphics[width=\textwidth]{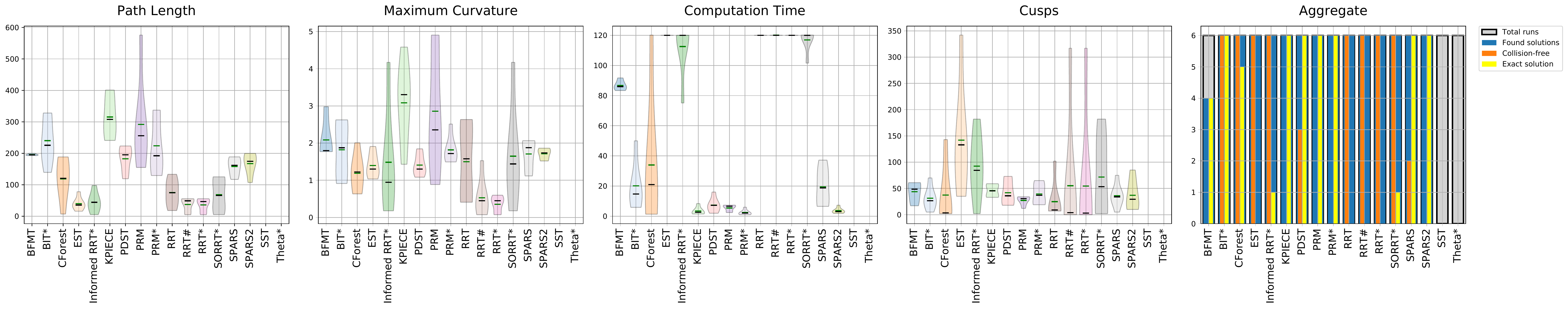}\\
    \includegraphics[width=\textwidth]{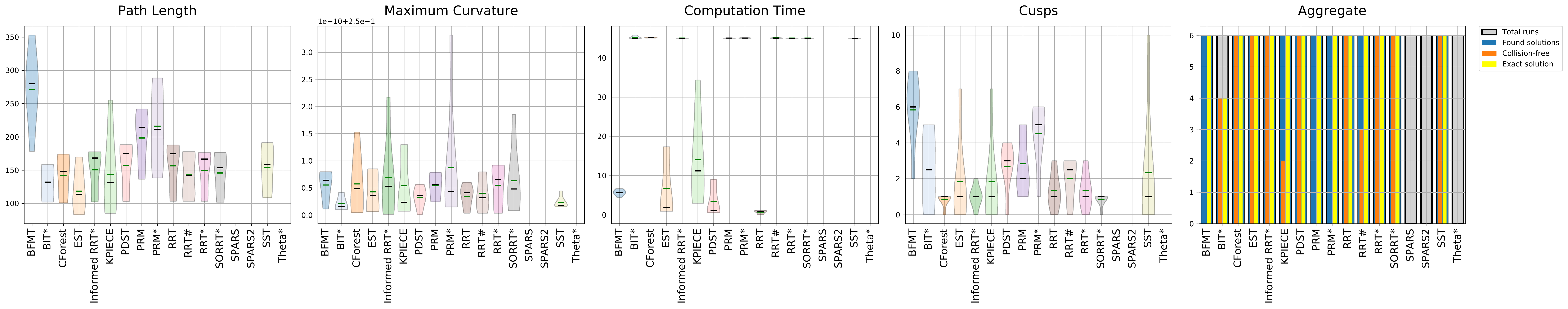}
    \caption{Statistics for the \emph{warehouse} scenarios. First row: Reeds Shepp steering, second row: CC Reeds Shepp steering, third row: POSQ steering, fourth row: Dubins steering.}
    \label{fig:warehouse_stats}
\end{figure*}

Compared to most parking scenarios, the warehouse environment typically requires longer computation times for the planners (especially anytime planners) to find solutions. It offers considerably more opportunity for the planners to find solutions of varying homotopy classes (cf. \autoref{fig:warehouse_trajectories}), resulting in a larger variance of path length. CForest, Informed RRT${}^*$, and SORRT${}^*$ consistently find among the shortest paths, although Informed RRT${}^*$ has among the longest computation times (cf. \autoref{fig:warehouse_stats}).

\subsection{Procedurally-generated grid environments}
\label{sec:procedural}

As described in \autoref{sec:environments}, we procedurally generate environments to have full control over the shape of the free space within the planners need to find solutions. This allows us to precisely analyze how varying features of the environments influence the planning results.

\subsubsection{Varying corridor sizes}
\label{sec:cross-corridor}

\begin{figure*}
    \centering
    \includegraphics[width=.9\textwidth]{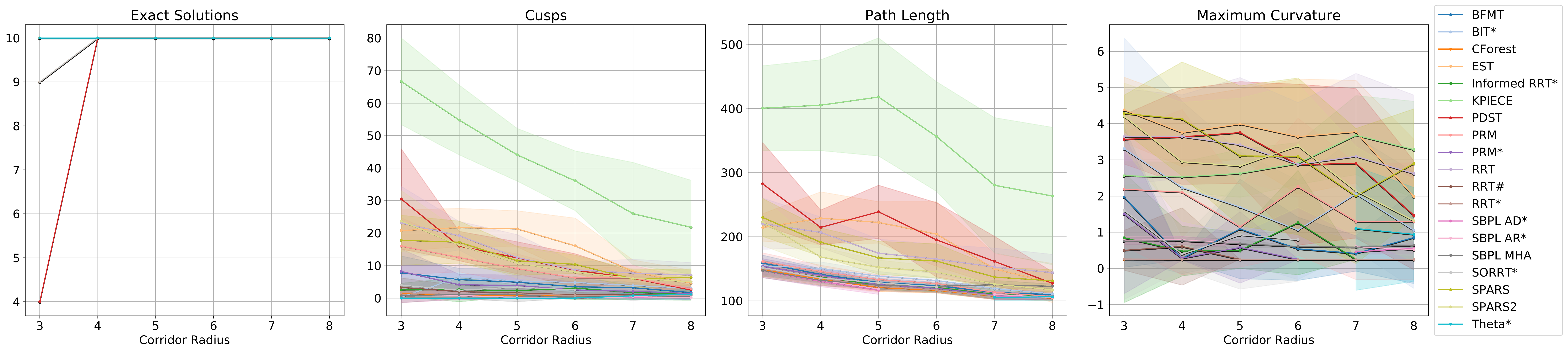}
    \caption{Various planning statistics for the Reeds Shepp steer function in the procedurally grid environments with varying corridor sizes.}
    \label{fig:cross_corridor_radii}
\end{figure*}

As shown on the abscissa in \autoref{fig:cross_corridor_radii}, the corridor sizes are expressed in the number of grid cells. We sample five $100\times100$ grid environments for each corridor radius (\autoref{fig:grid_envs} bottom row), sampled from the same starting seed over radii between three and eight grid cells. As we increase the corridor size, the path lengths of all planners decrease, as well as the number of cusps. The curvature metric remains mostly unaffected, except for PDST, PRM and SPARS2 where it considerable decreases. Theta$^*$, the SBPL planners, Informed RRT$^*$ and RRT$\#$ constantly have a low number of cusps and achieve very low path lengths across all conditions. KPIECE performs the worst in number of cusps and path length. EST, KPIECE, SPARS, SPARS2 and PRM have poor curvature, but PDST improves by a factor of two toward the maximum corridor size. While PDST and RRT initially find four and nine out of ten possible solutions, only at a corridor radius of four cells do all planners find exact solutions in every case.

\subsubsection{Varying turning radii}
\label{sec:cross-turning}

\begin{figure*}
    \centering
    \includegraphics[width=.9\textwidth]{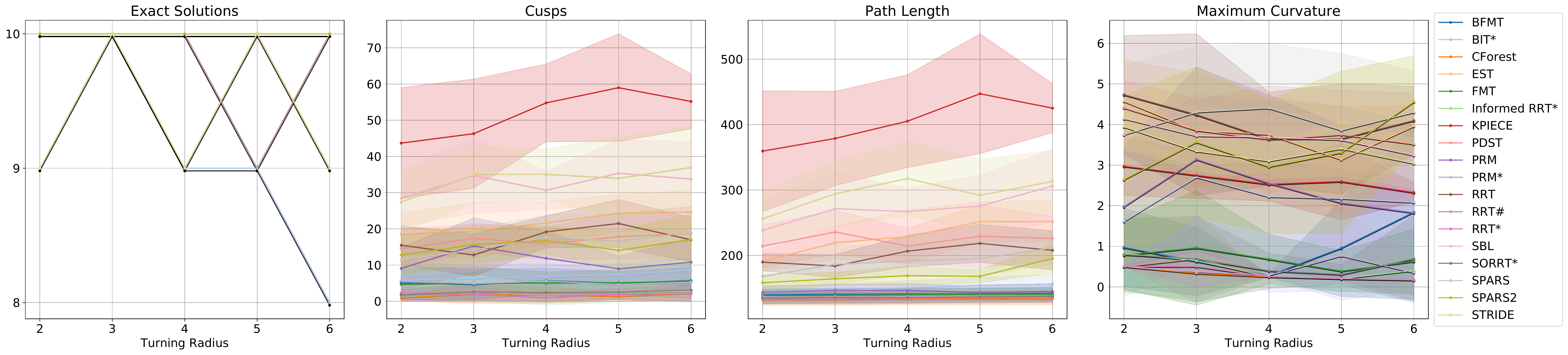}
    \caption{Various planning statistics for different turning radii (in meters) of the Reeds Shepp steer function in the procedurally grid environments (size: $100\times100$).}
    \label{fig:cross_turn_radii}
\end{figure*}

We vary the turning radius used by the Reeds Shepp steer function and evaluate the planners on a $100\times100$ indoor-like grid environment with a corridor radius of five grid cells (cf. \autoref{fig:grid_envs} bottom row). The change in turning radius has a surprisingly little effect on the path quality, see \autoref{fig:cross_turn_radii}. Slight developments can be observed where the path lengths tend to increase as the turning radius becomes larger. Especially PRM has a pronounced inclination in the number of cusps. The curvature is generally not tending in any direction significantly. The number of exact solutions is at zero for Theta$^*$, PRM constantly finds two out of ten solutions, while the other planners find all of the solutions exactly.

\subsubsection{Varying obstacle densities}
\label{sec:cross-density}

\begin{figure*}
    \centering
    \includegraphics[width=.9\textwidth]{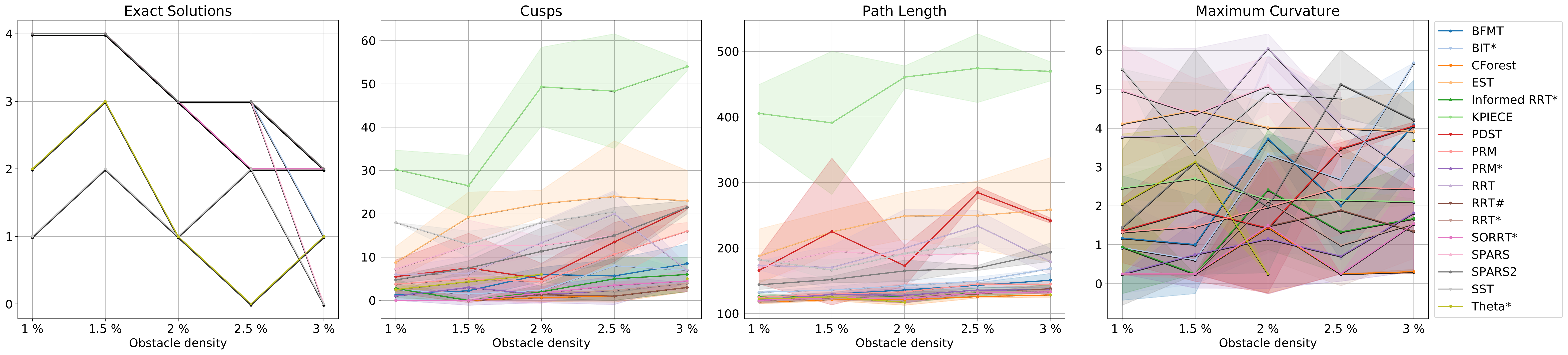}
    \caption{Planning statistics of the Reeds Shepp steer function in the procedurally generated grid environments (size: $100\times100$) with varying occupancy ratios.}
    \label{fig:cross_density}
\end{figure*}

As described in \autoref{sec:environments}, in this experiment, we randomly set cells of a $100\times100$ grid environment to be occupied until a selected density, i.e. ratio between occupied and free cells, has been achieved (see \autoref{fig:grid_envs} top row). Through various experiments, we determined the ranges between $1\%$ and $3\%$ to yield meaningful results. We successively increase the obstacle density in steps of $0.5\%$, and yet the influence on the quality of the found solutions is significant. From \autoref{fig:cross_density}, we can see that none of the planners are able to find exact solutions in all ten cases, most start at four solutions which drops to one and zero as the maximum obstacle density is approached. Meanwhile, the number of cusps increases dramatically, especially for KPIECE, PDST; and even BFMT, SPARS and PRM$^*$ have a relatively strong increase. The path lengths are not as much affected, although increasing in many cases, such as EST, SPARS2, SPARS, KPIECE. The curvature is increasing for many planners, such as PDST, PRM, PRM$^*$.

\subsection{Planning and Post-Smoothing}
\label{sec:sam-vs-any}

Based on our experiments with varying time limits over a range of time limits between zero and 30 seconds, we are investigating how post-smoothing methods can benefit the motion planning pipeline. In combination with sampling-based planners, which quickly find feasible solutions, can these improvement techniques yield results that are qualitatively competitive with the solutions obtained by anytime planners within shorter computation times?

To answer this question, we run a set of sampling-based planners (EST, RRT, SBL, STRIDE) with all post-smoothing methods considered in this benchmark, and compare it against anytime planners run at time limits ranging between five and 60 seconds.

\begin{figure*}
    \centering
    \includegraphics[height=5cm]{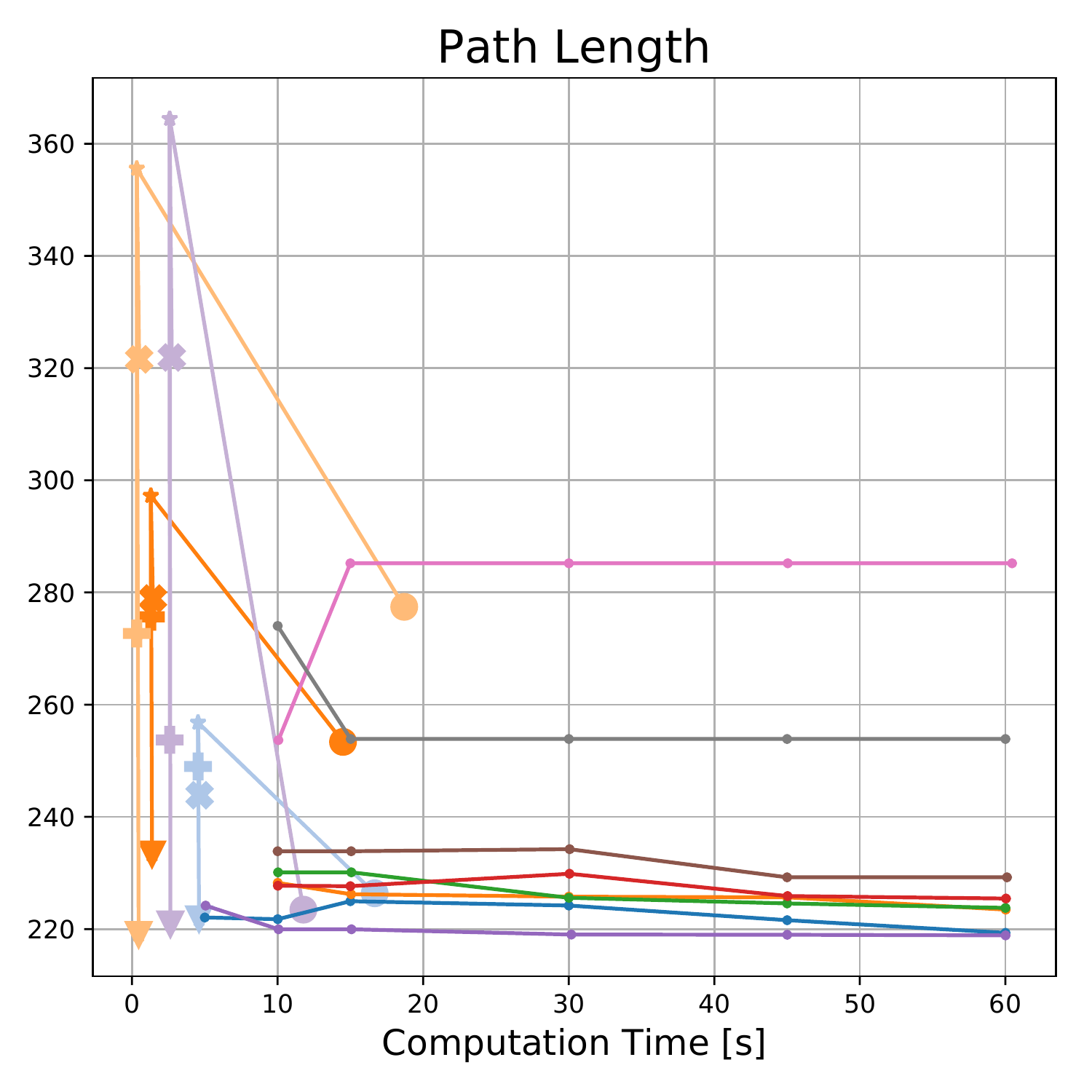}
    \includegraphics[height=5cm]{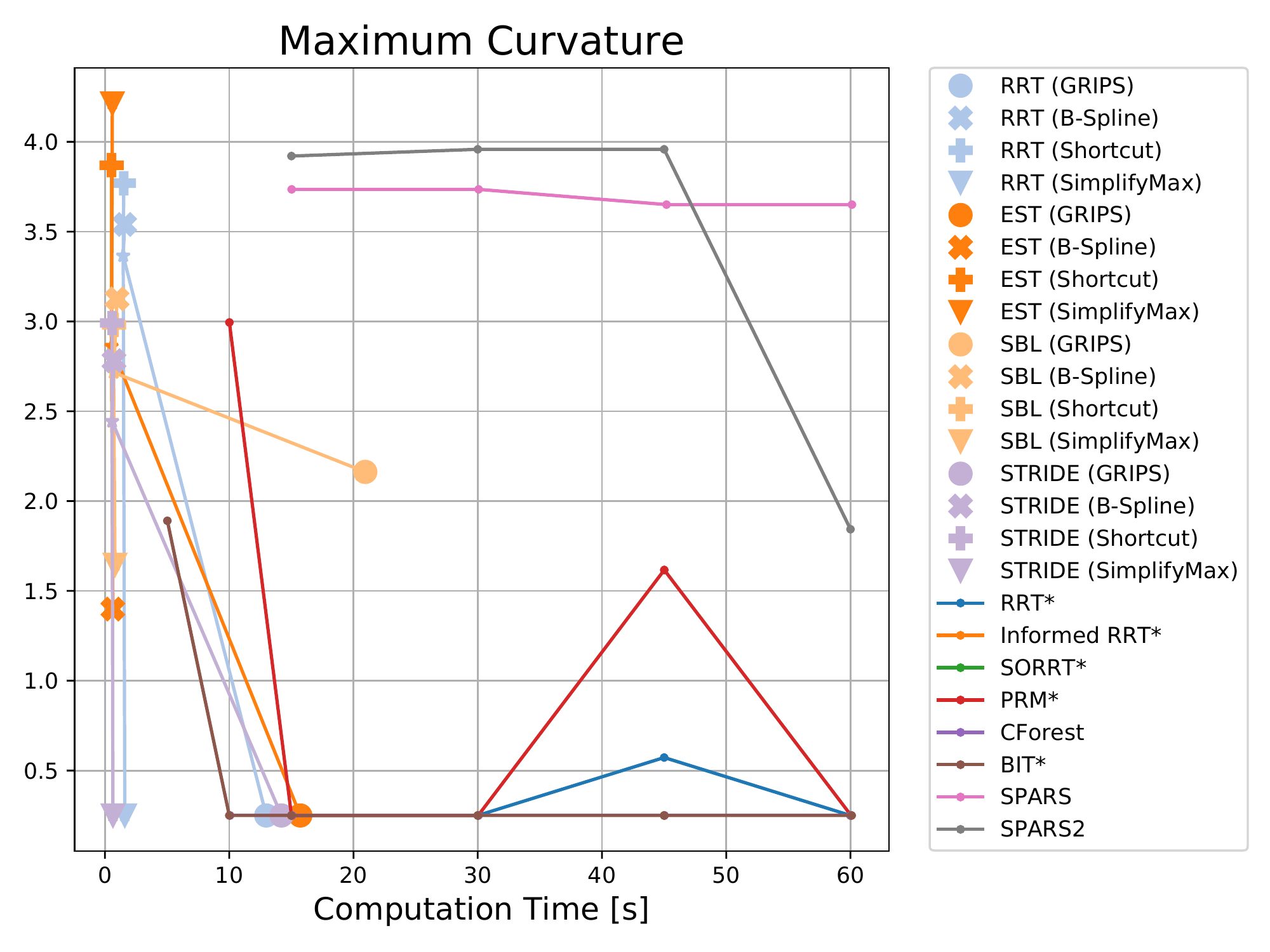}
    \caption{Comparison of sampling-based planners in combination with post-smoothing methods and anytime planners evaluated over maximum time limits 5, 10, 15, 30, 45, 60 seconds on a $150\times150$ grid environment.
    The initial solution found by the sampling-based planners is indicated by a \ding{72} symbol, the post-smoothers GRIPS (\ding{108}), B-Spline (\ding{54}), Shortcut (\ding{58}) and SimplifyMax (\ding{116}) are marked according to the legend. The anytime planners are shown as solid lines with $\cdot$ markers.
    \textit{Left:} path length of the respective solutions. \textit{Right:} maximum curvature.
    }
    \label{fig:sam_vs_any}
\end{figure*}

As shown in \autoref{fig:sam_vs_any}, we observe that the algorithms GRIPS and Simplify Max yield significant improvements in path length and maximum curvature. They both reduce the path length typically by a factor of two and similarly smooth the path in a way that the maximum curvature drops by close to a factor of two. In most cases, Simplify Max is considerably faster than GRIPS to obtain these results. The B-spline algorithm does not always improve the path quality, which may be explained by the problem that B-splines do not translate well to curves that can be followed by Reeds Shepp steering, leading to slight turns that increase the curvature.

\begin{figure*}
    \centering
    \includegraphics[height=4.5cm]{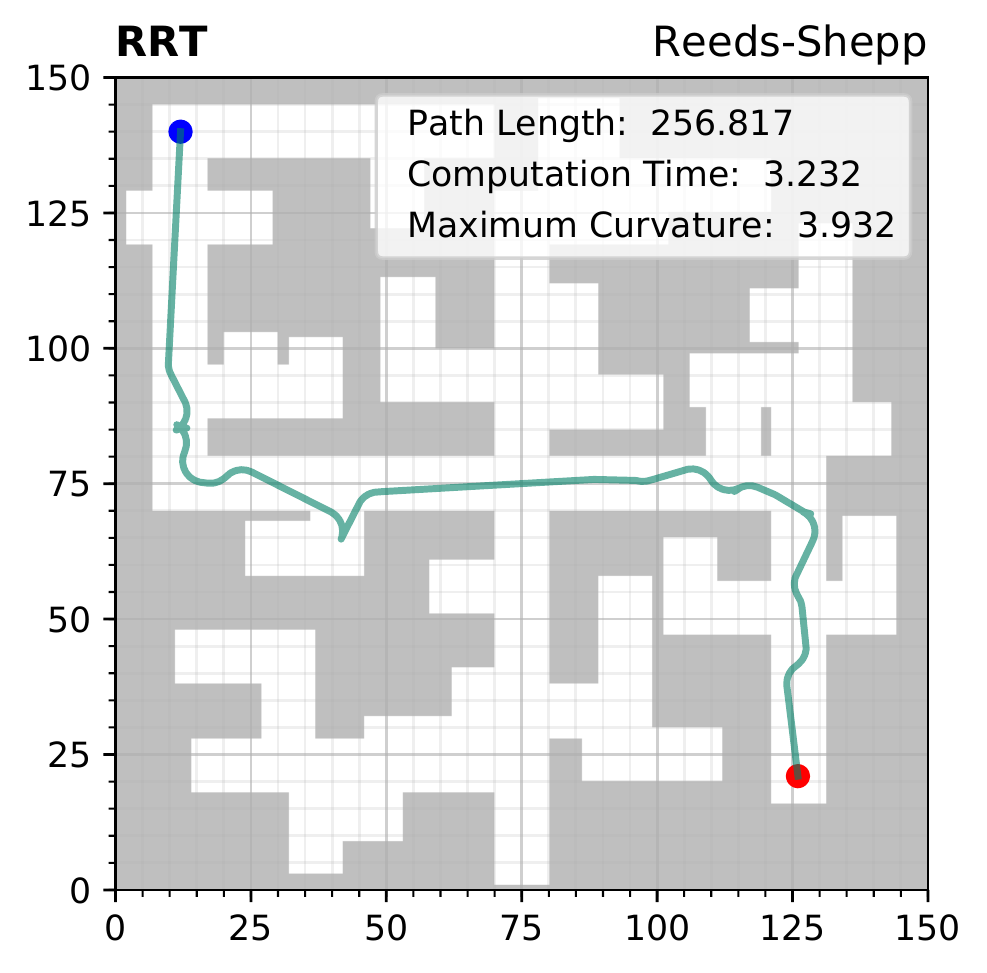}
    \includegraphics[height=4.5cm]{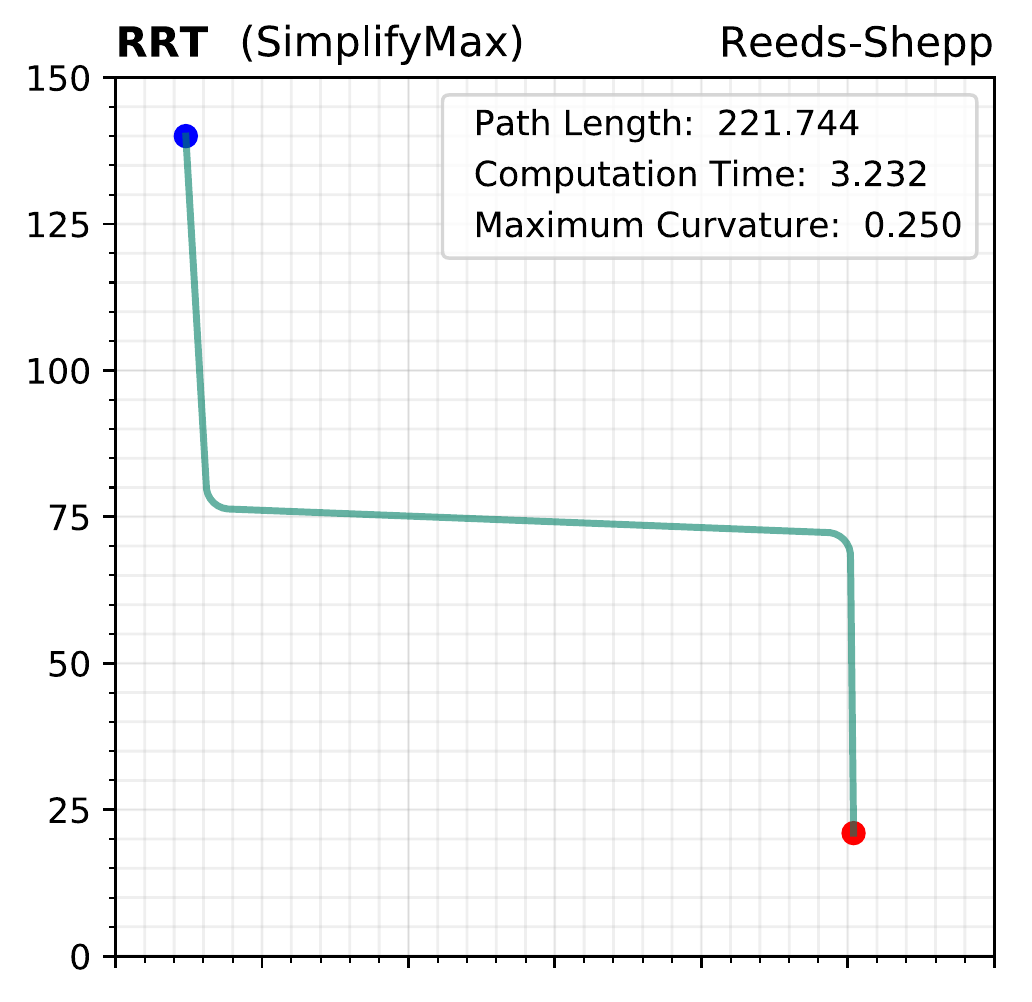}
    \includegraphics[height=4.5cm]{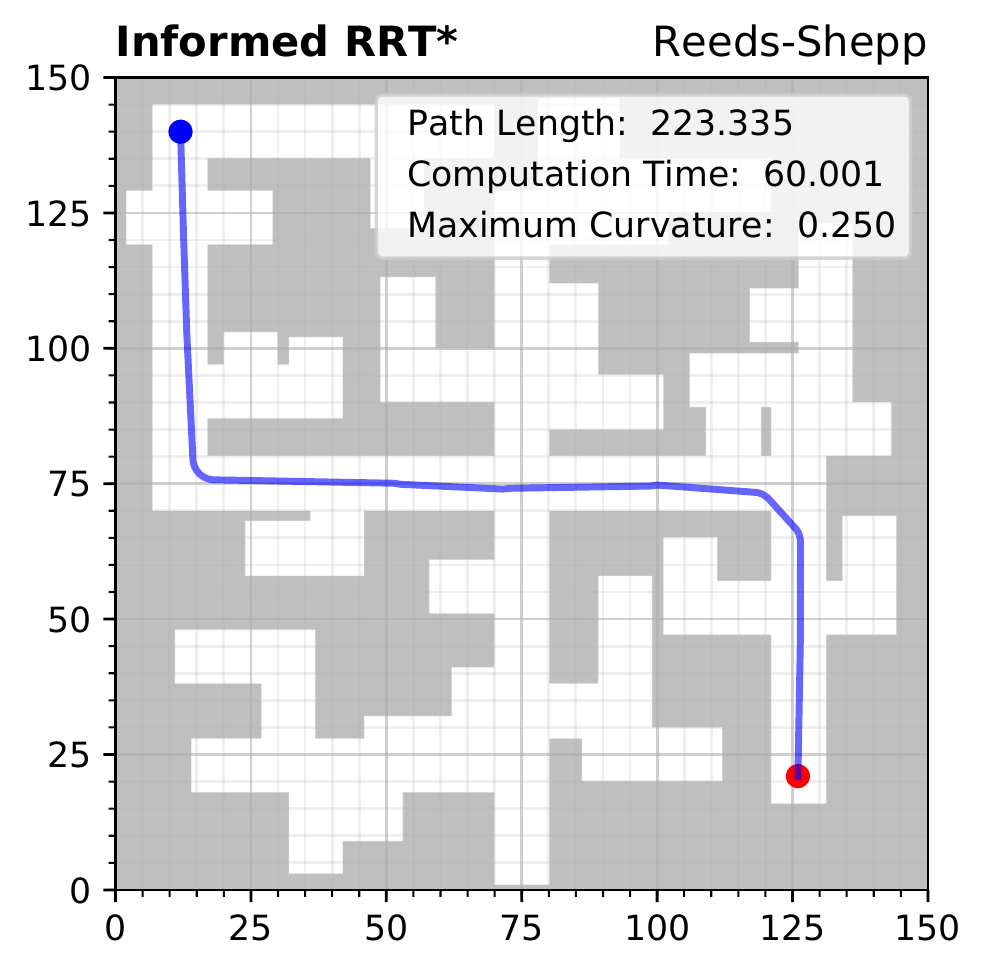}
    \caption{Trajectories resulting from the comparison of sampling-based planners in combination with post-smoothing methods against anytime planners. The solution on the left is obtained from the sampling-based planner RRT after \SI{3.232}{\second}. Using the SimplifyMax algorithm, this solution is smoothed (center), within a total time (including RRT planning) of \SI{3.232}{\second}. On the right, the solution from Informed RRT${}^*$ is shown, which is computed after \SI{60.001}{\second}.
    }
    \label{fig:sam_vs_any_traj}
\end{figure*}

Overall, there exist several couplings between sampling-based planners and post-smoothers that outperform anytime planners in speed and solution quality. For example, within three seconds RRT combined with Simplify Max smoothing achieves a maximum curvature at the same level as an anytime planner such as Informed RRT$^*$ after 60 seconds, while yielding a shorter path length (\autoref{fig:sam_vs_any_traj}).

\begin{table*}[t!]
    \centering\footnotesize
    \resizebox{\textwidth}{!}{%
        \begin{tabular}{p{2cm}>{\centering}m{1.2\maxlen}*{5}{S[table-format=3.2(8)]}}
\toprule
\bf Planner & \mcc{Solutions} & \mcc{Time [s]} & \mcc{Path~Length} & \mcc{Curvature} & \mcc{Clearance} & \mcc{Cusps} \\\midrule
\rowlabel{\textbf{Scenario: \texttt{Berlin\_0\_256}} (SBPL, \SI{12}{\minute} time limit)}
\\
SBPL AD${}^*$                            & %
	{\hspace{-1.5cm}\databartwo{0.18}{0.18}\makebox[0pt][c]{\hspace{1cm}9 / 9}} &
	{\databar{1.00}}	120.05 \pm 0.00 &
	{\databar{0.91}}	361.92 \pm 6.23 &
	{\databar{0.56}}	1.36 \pm 1.74 &
	{\databar{0.81}}	10.40 \pm 1.88 &
	{\databar{0.92}}	34 %
\\
SBPL ARA${}^*$                            & %
	{\hspace{-1.5cm}\databartwo{0.08}{0.08}\makebox[0pt][c]{\hspace{1cm}4 / 4}} &
	{\databar{1.00}}	120.05 \pm 0.01 &
	{\databar{0.91}}	360.43 \pm 3.94 &
	{\databar{0.31}}	0.77 \pm 0.07 &
	{\databar{0.86}}	10.99 \pm 1.35 &
	{\databar{0.30}}	11 %
\\
SBPL MHA$^*$                                & %
	{\hspace{-1.5cm}\databartwo{0.20}{0.20}\makebox[0pt][c]{\hspace{1cm}10 / 10}} &
	{\databar{0.04}}	4.52 \pm 4.53 &
	{\databar{1.00}}	397.16 \pm 5.58 &
	{\databar{1.00}}	2.45 \pm 2.62 &
	{\databar{1.00}}	12.81 \pm 2.90 &
	{\databar{1.00}}	37 %
\\
\rowlabel{\textbf{Scenario: \texttt{Berlin\_0\_256}} (Reeds-Shepp steering, \SI{1.5}{\minute} time limit)}
\\
BFMT                                     & %
	{\hspace{-1.5cm}\databartwo{1.00}{0.98}\makebox[0pt][c]{\hspace{1cm}50 / 51}} &
	{\databar{0.02}}	1.39 \pm 3.89 &
	{\databar{0.34}}	369.51 \pm 8.72 &
	{\databar{0.55}}	1.13 \pm 0.96 &
	{\databar{0.62}}	8.82 \pm 2.87 &
	{\databar{0.11}}	204 %
\\
BIT${}^*$                                & %
	{\hspace{-1.5cm}\databartwo{0.98}{0.25}\makebox[0pt][c]{\hspace{1cm}13 / 50}} &
	{\databar{1.00}}	90.05 \pm 0.09 &
	{\databar{0.34}}	362.10 \pm 6.34 &
	{\databar{0.61}}	1.27 \pm 1.01 &
	{\databar{0.56}}	7.94 \pm 2.11 &
	{\databar{0.09}}	159 %
\\
CForest                                  & %
	{\hspace{-1.5cm}\databartwo{1.00}{0.10}\makebox[0pt][c]{\hspace{1cm}5 / 51}} &
	{\databar{1.00}}	90.04 \pm 0.07 &
	{\databar{0.32}}	347.10 \pm 5.54 &
	{\databar{0.27}}	0.56 \pm 0.73 &
	{\databar{0.49}}	7.05 \pm 1.83 &
	{\databar{0.04}}	70 %
\\
EST                                      & %
	{\hspace{-1.5cm}\databartwo{1.00}{0.90}\makebox[0pt][c]{\hspace{1cm}46 / 51}} &
	{\databar{0.02}}	2.19 \pm 12.47 &
	{\databar{0.53}}	566.35 \pm 116.92 &
	{\databar{0.82}}	1.70 \pm 0.43 &
	{\databar{0.78}}	11.16 \pm 1.57 &
	{\databar{0.43}}	769 %
\\
Informed RRT${}^*$                       & %
	{\hspace{-1.5cm}\databartwo{1.00}{0.25}\makebox[0pt][c]{\hspace{1cm}13 / 51}} &
	{\databar{1.00}}	90.01 \pm 0.01 &
	{\databar{0.33}}	350.08 \pm 5.89 &
	{\databar{0.17}}	0.35 \pm 0.33 &
	{\databar{0.54}}	7.68 \pm 2.02 &
	{\databar{0.04}}	67 %
\\
KPIECE                                   & %
	{\hspace{-1.5cm}\databartwo{1.00}{0.88}\makebox[0pt][c]{\hspace{1cm}45 / 51}} &
	{\databar{0.01}}	1.03 \pm 6.95 &
	{\databar{1.00}}	1077.02 \pm 344.55 &
	{\databar{0.57}}	1.17 \pm 0.51 &
	{\databar{0.78}}	11.22 \pm 1.38 &
	{\databar{1.00}}	1791 %
\\
PDST                                     & %
	{\hspace{-1.5cm}\databartwo{1.00}{0.84}\makebox[0pt][c]{\hspace{1cm}43 / 51}} &
	{\databar{0.04}}	3.40 \pm 13.55 &
	{\databar{0.54}}	580.61 \pm 143.46 &
	{\databar{0.68}}	1.42 \pm 0.61 &
	{\databar{0.79}}	11.32 \pm 1.74 &
	{\databar{0.31}}	555 %
\\
PRM                                      & %
	{\hspace{-1.5cm}\databartwo{1.00}{0.47}\makebox[0pt][c]{\hspace{1cm}24 / 51}} &
	{\databar{1.00}}	90.08 \pm 0.07 &
	{\databar{0.34}}	364.16 \pm 10.69 &
	{\databar{0.59}}	1.21 \pm 1.12 &
	{\databar{0.62}}	8.80 \pm 2.40 &
	{\databar{0.18}}	329 %
\\
PRM${}^*$                                & %
	{\hspace{-1.5cm}\databartwo{1.00}{0.65}\makebox[0pt][c]{\hspace{1cm}33 / 51}} &
	{\databar{1.00}}	90.07 \pm 0.06 &
	{\databar{0.33}}	357.09 \pm 8.46 &
	{\databar{0.32}}	0.67 \pm 0.83 &
	{\databar{0.63}}	8.99 \pm 2.51 &
	{\databar{0.09}}	156 %
\\
RRT                                      & %
	{\hspace{-1.5cm}\databartwo{1.00}{0.94}\makebox[0pt][c]{\hspace{1cm}48 / 51}} &
	{\databar{0.05}}	4.08 \pm 17.43 &
	{\databar{0.44}}	475.59 \pm 70.71 &
	{\databar{0.93}}	1.93 \pm 0.53 &
	{\databar{0.77}}	11.00 \pm 1.44 &
	{\databar{0.36}}	636 %
\\
RRT\#                                    & %
	{\hspace{-1.5cm}\databartwo{1.00}{0.51}\makebox[0pt][c]{\hspace{1cm}26 / 51}} &
	{\databar{1.00}}	90.09 \pm 0.47 &
	{\databar{0.32}}	346.96 \pm 14.78 &
	{\databar{0.28}}	0.58 \pm 0.75 &
	{\databar{0.51}}	7.36 \pm 1.97 &
	{\databar{0.05}}	90 %
\\
RRT${}^*$                                & %
	{\hspace{-1.5cm}\databartwo{1.00}{0.35}\makebox[0pt][c]{\hspace{1cm}18 / 51}} &
	{\databar{1.00}}	90.01 \pm 0.01 &
	{\databar{0.32}}	347.58 \pm 12.50 &
	{\databar{0.21}}	0.44 \pm 0.55 &
	{\databar{0.51}}	7.31 \pm 1.94 &
	{\databar{0.04}}	80 %
\\
SORRT${}^*$                              & %
	{\hspace{-1.5cm}\databartwo{1.00}{0.41}\makebox[0pt][c]{\hspace{1cm}21 / 51}} &
	{\databar{1.00}}	90.01 \pm 0.01 &
	{\databar{0.32}}	350.00 \pm 5.64 &
	{\databar{0.21}}	0.44 \pm 0.59 &
	{\databar{0.52}}	7.44 \pm 1.89 &
	{\databar{0.04}}	75 %
\\
SPARS                                    & %
	{\hspace{-1.5cm}\databartwo{1.00}{0.90}\makebox[0pt][c]{\hspace{1cm}46 / 51}} &
	{\databar{1.00}}	90.33 \pm 0.43 &
	{\databar{0.44}}	471.09 \pm 100.80 &
	{\databar{0.81}}	1.68 \pm 0.71 &
	{\databar{0.77}}	11.05 \pm 0.95 &
	{\databar{0.33}}	586 %
\\
SPARS2                                   & %
	{\hspace{-1.5cm}\databartwo{0.98}{0.86}\makebox[0pt][c]{\hspace{1cm}44 / 50}} &
	{\databar{1.00}}	90.01 \pm 0.01 &
	{\databar{0.38}}	410.60 \pm 41.50 &
	{\databar{1.00}}	2.07 \pm 0.56 &
	{\databar{0.72}}	10.26 \pm 2.46 &
	{\databar{0.27}}	490 %
\\
SST                                      & %
	{\hspace{-1.5cm}\databartwo{1.00}{0.94}\makebox[0pt][c]{\hspace{1cm}48 / 51}} &
	{\databar{1.00}}	90.01 \pm 0.01 &
	{\databar{0.47}}	506.46 \pm 77.41 &
	{\databar{0.98}}	2.03 \pm 0.52 &
	{\databar{0.66}}	9.43 \pm 1.61 &
	{\databar{0.75}}	1348 %
\\
Theta${}^*$                              & %
	0 &
	N / A&
	N / A&
	N / A&
	N / A&
	N / A%
\\
\rowlabel{\textbf{Scenario: \texttt{Berlin\_0\_256}} (CC Reeds-Shepp steering, \SI{18}{\minute} time limit)}
\\
BFMT                                     & %
	{\hspace{-1.5cm}\databartwo{0.98}{0.96}\makebox[0pt][c]{\hspace{1cm}49 / 50}} &
	{\databar{0.03}}	35.67 \pm 4.26 &
	{\databar{0.36}}	366.93 \pm 12.15 &
	{\databar{0.40}}	0.59 \pm 0.77 &
	{\databar{0.75}}	8.78 \pm 2.14 &
	{\databar{0.02}}	126 %
\\
BIT${}^*$                                & %
	{\hspace{-1.5cm}\databartwo{0.55}{0.51}\makebox[0pt][c]{\hspace{1cm}26 / 28}} &
	{\databar{1.00}}	1080.46 \pm 0.85 &
	{\databar{0.36}}	372.78 \pm 9.39 &
	{\databar{0.52}}	0.75 \pm 0.70 &
	{\databar{0.67}}	7.83 \pm 2.83 &
	{\databar{0.03}}	101 %
\\
CForest                                  & %
	{\hspace{-1.5cm}\databartwo{1.00}{0.00}\makebox[0pt][c]{\hspace{1cm}0 / 51}} &
	{\databar{0.01}}	5.66 \pm 20.36 &
	{\databar{0.33}}	334.50 \pm 46.35 &
	{\databar{0.51}}	0.74 \pm 0.86 &
	{\databar{0.68}}	7.95 \pm 2.60 &
	{\databar{0.02}}	122 %
\\
EST                                      & %
	{\hspace{-1.5cm}\databartwo{1.00}{0.96}\makebox[0pt][c]{\hspace{1cm}49 / 51}} &
	{\databar{0.04}}	48.41 \pm 148.31 &
	{\databar{0.66}}	670.16 \pm 114.72 &
	{\databar{0.97}}	1.41 \pm 0.32 &
	{\databar{0.99}}	11.46 \pm 1.89 &
	{\databar{0.26}}	1793 %
\\
Informed RRT${}^*$                       & %
	{\hspace{-1.5cm}\databartwo{1.00}{0.76}\makebox[0pt][c]{\hspace{1cm}39 / 51}} &
	{\databar{1.00}}	1080.19 \pm 0.18 &
	{\databar{0.35}}	353.80 \pm 13.32 &
	{\databar{0.27}}	0.39 \pm 0.56 &
	{\databar{0.65}}	7.61 \pm 2.00 &
	{\databar{0.01}}	73 %
\\
KPIECE                                   & %
	{\hspace{-1.5cm}\databartwo{1.00}{0.51}\makebox[0pt][c]{\hspace{1cm}26 / 51}} &
	{\databar{0.07}}	77.82 \pm 235.60 &
	{\databar{1.00}}	1022.48 \pm 242.38 &
	{\databar{0.72}}	1.04 \pm 0.35 &
	{\databar{0.97}}	11.23 \pm 1.25 &
	{\databar{0.46}}	3221 %
\\
PDST                                     & %
	{\hspace{-1.5cm}\databartwo{1.00}{0.78}\makebox[0pt][c]{\hspace{1cm}40 / 51}} &
	{\databar{0.12}}	132.05 \pm 244.65 &
	{\databar{0.51}}	523.04 \pm 129.90 &
	{\databar{0.82}}	1.18 \pm 0.61 &
	{\databar{1.00}}	11.63 \pm 1.69 &
	{\databar{0.08}}	540 %
\\
PRM                                      & %
	{\hspace{-1.5cm}\databartwo{1.00}{0.75}\makebox[0pt][c]{\hspace{1cm}38 / 51}} &
	{\databar{0.98}}	1062.67 \pm 124.51 &
	{\databar{0.38}}	389.49 \pm 32.72 &
	{\databar{0.58}}	0.84 \pm 0.80 &
	{\databar{0.78}}	9.06 \pm 1.89 &
	{\databar{0.06}}	411 %
\\
PRM${}^*$                                & %
	{\hspace{-1.5cm}\databartwo{1.00}{0.75}\makebox[0pt][c]{\hspace{1cm}38 / 51}} &
	{\databar{1.00}}	1080.34 \pm 0.19 &
	{\databar{0.37}}	379.15 \pm 28.99 &
	{\databar{0.59}}	0.86 \pm 0.89 &
	{\databar{0.78}}	9.12 \pm 1.94 &
	{\databar{0.04}}	260 %
\\
RRT                                      & %
	{\hspace{-1.5cm}\databartwo{1.00}{0.96}\makebox[0pt][c]{\hspace{1cm}49 / 51}} &
	{\databar{0.03}}	35.39 \pm 153.67 &
	{\databar{0.49}}	500.27 \pm 75.97 &
	{\databar{0.87}}	1.26 \pm 0.66 &
	{\databar{0.98}}	11.42 \pm 1.59 &
	{\databar{0.08}}	527 %
\\
RRT\#                                    & %
	{\hspace{-1.5cm}\databartwo{1.00}{1.00}\makebox[0pt][c]{\hspace{1cm}51 / 51}} &
	{\databar{1.00}}	1080.54 \pm 0.62 &
	{\databar{0.34}}	351.14 \pm 19.94 &
	{\databar{0.20}}	0.29 \pm 0.39 &
	{\databar{0.69}}	7.97 \pm 1.96 &
	{\databar{0.01}}	69 %
\\
RRT${}^*$                                & %
	{\hspace{-1.5cm}\databartwo{1.00}{0.92}\makebox[0pt][c]{\hspace{1cm}47 / 51}} &
	{\databar{1.00}}	1080.15 \pm 0.08 &
	{\databar{0.34}}	348.93 \pm 17.16 &
	{\databar{0.25}}	0.36 \pm 0.48 &
	{\databar{0.66}}	7.71 \pm 1.95 &
	{\databar{0.01}}	55 %
\\
SORRT${}^*$                              & %
	{\hspace{-1.5cm}\databartwo{0.29}{0.18}\makebox[0pt][c]{\hspace{1cm}9 / 15}} &
	{\databar{1.00}}	1080.23 \pm 0.28 &
	{\databar{0.35}}	352.77 \pm 21.92 &
	{\databar{0.14}}	0.20 \pm 0.00 &
	{\databar{0.53}}	6.19 \pm 2.62 &
	{\databar{0.02}}	48 %
\\
SPARS                                    & %
	{\hspace{-1.5cm}\databartwo{0.96}{0.94}\makebox[0pt][c]{\hspace{1cm}48 / 49}} &
	{\databar{1.00}}	1083.04 \pm 2.92 &
	{\databar{0.46}}	468.54 \pm 75.90 &
	{\databar{0.96}}	1.39 \pm 0.73 &
	{\databar{0.97}}	11.28 \pm 1.30 &
	{\databar{0.06}}	429 %
\\
SPARS2                                   & %
	0 &
	N / A&
	N / A&
	N / A&
	N / A&
	N / A%
\\
SST                                      & %
	{\hspace{-1.5cm}\databartwo{1.00}{0.96}\makebox[0pt][c]{\hspace{1cm}49 / 51}} &
	{\databar{1.00}}	1080.03 \pm 0.02 &
	{\databar{0.59}}	605.81 \pm 79.93 &
	{\databar{1.00}}	1.45 \pm 1.10 &
	{\databar{0.84}}	9.81 \pm 1.57 &
	{\databar{1.00}}	6950 %
\\
Theta${}^*$                              & %
	0 &
	N / A&
	N / A&
	N / A&
	N / A&
	N / A%
\\
\rowlabel{\textbf{Scenario: \texttt{Berlin\_0\_256}} (POSQ steering, \SI{12}{\minute} time limit)}
\\
BFMT                                     & %
	{\hspace{-1.5cm}\databartwo{0.20}{0.08}\makebox[0pt][c]{\hspace{1cm}4 / 10}} &
	{\databar{0.81}}	582.35 \pm 69.16 &
	{\databar{0.82}}	1010.48 \pm 287.89 &
	{\databar{0.68}}	0.98 \pm 0.33 &
	{\databar{0.80}}	13.03 \pm 1.39 &
	{\databar{0.08}}	137 %
\\
BIT${}^*$                                & %
	{\hspace{-1.5cm}\databartwo{0.80}{0.69}\makebox[0pt][c]{\hspace{1cm}35 / 41}} &
	{\databar{0.20}}	141.23 \pm 96.90 &
	{\databar{0.64}}	790.39 \pm 368.17 &
	{\databar{0.68}}	0.98 \pm 0.36 &
	{\databar{0.79}}	12.78 \pm 2.23 &
	{\databar{0.24}}	396 %
\\
CForest                                  & %
	{\hspace{-1.5cm}\databartwo{1.00}{0.55}\makebox[0pt][c]{\hspace{1cm}28 / 51}} &
	{\databar{0.21}}	149.11 \pm 189.51 &
	{\databar{0.28}}	341.97 \pm 144.55 &
	{\databar{0.64}}	0.92 \pm 0.41 &
	{\databar{0.85}}	13.75 \pm 5.26 &
	{\databar{0.10}}	171 %
\\
EST                                      & %
	{\hspace{-1.5cm}\databartwo{1.00}{0.98}\makebox[0pt][c]{\hspace{1cm}50 / 51}} &
	{\databar{1.00}}	720.03 \pm 0.03 &
	{\databar{0.11}}	138.47 \pm 57.90 &
	{\databar{0.99}}	1.42 \pm 0.37 &
	{\databar{0.98}}	15.91 \pm 6.65 &
	{\databar{0.94}}	1563 %
\\
Informed RRT${}^*$                       & %
	{\hspace{-1.5cm}\databartwo{1.00}{0.96}\makebox[0pt][c]{\hspace{1cm}49 / 51}} &
	{\databar{0.92}}	663.75 \pm 193.44 &
	{\databar{0.14}}	177.58 \pm 111.16 &
	{\databar{0.69}}	0.99 \pm 0.46 &
	{\databar{0.90}}	14.64 \pm 5.63 &
	{\databar{0.38}}	636 %
\\
KPIECE                                   & %
	{\hspace{-1.5cm}\databartwo{1.00}{0.41}\makebox[0pt][c]{\hspace{1cm}21 / 51}} &
	{\databar{0.03}}	24.20 \pm 100.24 &
	{\databar{1.00}}	1236.73 \pm 349.43 &
	{\databar{0.69}}	1.00 \pm 0.32 &
	{\databar{0.66}}	10.72 \pm 1.57 &
	{\databar{1.00}}	1662 %
\\
PDST                                     & %
	{\hspace{-1.5cm}\databartwo{1.00}{0.14}\makebox[0pt][c]{\hspace{1cm}7 / 51}} &
	{\databar{0.09}}	65.20 \pm 126.95 &
	{\databar{0.47}}	579.09 \pm 137.92 &
	{\databar{1.00}}	1.44 \pm 0.48 &
	{\databar{0.67}}	10.86 \pm 2.25 &
	{\databar{0.89}}	1487 %
\\
PRM                                      & %
	{\hspace{-1.5cm}\databartwo{1.00}{0.08}\makebox[0pt][c]{\hspace{1cm}4 / 51}} &
	{\databar{0.12}}	88.85 \pm 223.91 &
	{\databar{0.52}}	643.63 \pm 237.00 &
	{\databar{0.87}}	1.25 \pm 0.67 &
	{\databar{0.49}}	7.94 \pm 1.84 &
	{\databar{0.69}}	1150 %
\\
PRM${}^*$                                & %
	{\hspace{-1.5cm}\databartwo{1.00}{0.08}\makebox[0pt][c]{\hspace{1cm}4 / 51}} &
	{\databar{0.09}}	66.34 \pm 190.56 &
	{\databar{0.49}}	607.59 \pm 179.27 &
	{\databar{0.82}}	1.18 \pm 0.61 &
	{\databar{0.50}}	8.09 \pm 1.87 &
	{\databar{0.49}}	817 %
\\
RRT                                      & %
	{\hspace{-1.5cm}\databartwo{0.98}{0.96}\makebox[0pt][c]{\hspace{1cm}49 / 50}} &
	{\databar{0.96}}	688.95 \pm 141.08 &
	{\databar{0.40}}	496.80 \pm 184.33 &
	{\databar{0.75}}	1.09 \pm 0.76 &
	{\databar{0.84}}	13.67 \pm 3.12 &
	{\databar{0.23}}	378 %
\\
RRT\#                                    & %
	{\hspace{-1.5cm}\databartwo{1.00}{0.86}\makebox[0pt][c]{\hspace{1cm}44 / 51}} &
	{\databar{0.96}}	692.14 \pm 138.87 &
	{\databar{0.12}}	145.06 \pm 97.17 &
	{\databar{0.69}}	1.00 \pm 0.47 &
	{\databar{1.00}}	16.19 \pm 6.63 &
	{\databar{0.65}}	1087 %
\\
RRT${}^*$                                & %
	{\hspace{-1.5cm}\databartwo{1.00}{0.90}\makebox[0pt][c]{\hspace{1cm}46 / 51}} &
	{\databar{0.96}}	692.09 \pm 138.98 &
	{\databar{0.12}}	152.48 \pm 105.35 &
	{\databar{0.70}}	1.01 \pm 0.43 &
	{\databar{0.96}}	15.57 \pm 6.15 &
	{\databar{0.75}}	1240 %
\\
SORRT${}^*$                              & %
	{\hspace{-1.5cm}\databartwo{0.98}{0.88}\makebox[0pt][c]{\hspace{1cm}45 / 50}} &
	{\databar{0.93}}	669.23 \pm 176.53 &
	{\databar{0.13}}	162.12 \pm 109.61 &
	{\databar{0.74}}	1.06 \pm 0.48 &
	{\databar{1.00}}	16.22 \pm 7.13 &
	{\databar{0.35}}	563 %
\\
SPARS                                    & %
	{\hspace{-1.5cm}\databartwo{0.92}{0.00}\makebox[0pt][c]{\hspace{1cm}0 / 47}} &
	{\databar{0.23}}	165.72 \pm 171.92 &
	{\databar{0.54}}	662.60 \pm 168.32 &
	{\databar{0.68}}	0.98 \pm 0.32 &
	{\databar{0.62}}	10.01 \pm 1.44 &
	{\databar{0.20}}	317 %
\\
SPARS2                                   & %
	{\hspace{-1.5cm}\databartwo{1.00}{0.14}\makebox[0pt][c]{\hspace{1cm}7 / 51}} &
	{\databar{0.04}}	28.64 \pm 61.18 &
	{\databar{0.44}}	549.87 \pm 126.71 &
	{\databar{0.83}}	1.19 \pm 0.43 &
	{\databar{0.64}}	10.41 \pm 1.82 &
	{\databar{0.31}}	517 %
\\
SST                                      & %
	0 &
	N / A&
	N / A&
	N / A&
	N / A&
	N / A%
\\
Theta${}^*$                              & %
	{\hspace{-1.5cm}\databartwo{0.18}{0.18}\makebox[0pt][c]{\hspace{1cm}9 / 9}} &
	{\databar{0.70}}	504.14 \pm 140.32 &
	{\databar{0.29}}	364.74 \pm 10.22 &
	{\databar{0.55}}	0.79 \pm 0.19 &
	{\databar{0.19}}	3.08 \pm 0.53 &
	{\databar{0.02}}	36 %
\\
\rowlabel{\textbf{Scenario: \texttt{Berlin\_0\_256}} (Dubins steering, \SI{6}{\minute} time limit)}
\\
BFMT                                     & %
	{\hspace{-1.5cm}\databartwo{0.76}{0.06}\makebox[0pt][c]{\hspace{1cm}3 / 39}} &
	{\databar{0.11}}	39.60 \pm 69.14 &
	{\databar{0.43}}	517.98 \pm 66.21 &
	{\databar{0.09}}	0.25 \pm 0.00 &
	{\databar{0.75}}	9.64 \pm 1.73 &
	{\databar{0.24}}	343 %
\\
BIT${}^*$                                & %
	{\hspace{-1.5cm}\databartwo{0.71}{0.41}\makebox[0pt][c]{\hspace{1cm}21 / 36}} &
	{\databar{1.00}}	360.06 \pm 0.08 &
	{\databar{0.30}}	363.68 \pm 16.81 &
	{\databar{0.09}}	0.25 \pm 0.00 &
	{\databar{0.62}}	7.95 \pm 2.15 &
	{\databar{0.01}}	18 %
\\
CForest                                  & %
	{\hspace{-1.5cm}\databartwo{1.00}{0.12}\makebox[0pt][c]{\hspace{1cm}6 / 51}} &
	{\databar{1.00}}	360.07 \pm 0.03 &
	{\databar{0.29}}	352.27 \pm 20.64 &
	{\databar{0.09}}	0.25 \pm 0.02 &
	{\databar{0.58}}	7.52 \pm 2.18 &
	{\databar{0.03}}	43 %
\\
EST                                      & %
	{\hspace{-1.5cm}\databartwo{1.00}{0.76}\makebox[0pt][c]{\hspace{1cm}39 / 51}} &
	{\databar{0.17}}	59.46 \pm 130.88 &
	{\databar{0.56}}	675.37 \pm 167.61 &
	{\databar{0.09}}	0.25 \pm 0.00 &
	{\databar{0.95}}	12.22 \pm 1.54 &
	{\databar{0.27}}	372 %
\\
Informed RRT${}^*$                       & %
	{\hspace{-1.5cm}\databartwo{0.98}{0.37}\makebox[0pt][c]{\hspace{1cm}19 / 50}} &
	{\databar{1.00}}	360.04 \pm 0.03 &
	{\databar{0.29}}	346.49 \pm 50.74 &
	{\databar{0.09}}	0.25 \pm 0.01 &
	{\databar{0.61}}	7.82 \pm 2.19 &
	{\databar{0.04}}	52 %
\\
KPIECE                                   & %
	{\hspace{-1.5cm}\databartwo{1.00}{0.73}\makebox[0pt][c]{\hspace{1cm}37 / 51}} &
	{\databar{0.12}}	44.94 \pm 115.76 &
	{\databar{1.00}}	1210.09 \pm 393.20 &
	{\databar{0.09}}	0.25 \pm 0.02 &
	{\databar{1.00}}	12.87 \pm 1.41 &
	{\databar{0.63}}	885 %
\\
PDST                                     & %
	{\hspace{-1.5cm}\databartwo{0.96}{0.61}\makebox[0pt][c]{\hspace{1cm}31 / 49}} &
	{\databar{0.13}}	48.08 \pm 119.29 &
	{\databar{0.43}}	524.89 \pm 114.19 &
	{\databar{0.09}}	0.25 \pm 0.00 &
	{\databar{0.90}}	11.59 \pm 2.12 &
	{\databar{0.09}}	124 %
\\
PRM                                      & %
	{\hspace{-1.5cm}\databartwo{1.00}{0.00}\makebox[0pt][c]{\hspace{1cm}0 / 51}} &
	{\databar{1.00}}	360.05 \pm 0.03 &
	{\databar{0.48}}	580.32 \pm 93.38 &
	{\databar{0.09}}	0.25 \pm 0.00 &
	{\databar{0.64}}	8.20 \pm 2.58 &
	{\databar{0.40}}	554 %
\\
PRM${}^*$                                & %
	{\hspace{-1.5cm}\databartwo{1.00}{0.00}\makebox[0pt][c]{\hspace{1cm}0 / 51}} &
	{\databar{1.00}}	360.07 \pm 0.05 &
	{\databar{0.45}}	545.61 \pm 59.86 &
	{\databar{0.09}}	0.25 \pm 0.00 &
	{\databar{0.57}}	7.31 \pm 1.97 &
	{\databar{0.38}}	532 %
\\
RRT                                      & %
	{\hspace{-1.5cm}\databartwo{1.00}{0.73}\makebox[0pt][c]{\hspace{1cm}37 / 51}} &
	{\databar{0.16}}	59.14 \pm 126.73 &
	{\databar{0.43}}	518.19 \pm 155.43 &
	{\databar{0.09}}	0.25 \pm 0.01 &
	{\databar{0.95}}	12.21 \pm 2.31 &
	{\databar{0.09}}	126 %
\\
RRT\#                                    & %
	{\hspace{-1.5cm}\databartwo{1.00}{0.04}\makebox[0pt][c]{\hspace{1cm}2 / 51}} &
	{\databar{1.00}}	360.03 \pm 0.02 &
	{\databar{0.28}}	338.41 \pm 71.86 &
	{\databar{0.09}}	0.25 \pm 0.01 &
	{\databar{0.58}}	7.42 \pm 2.09 &
	{\databar{0.04}}	50 %
\\
RRT${}^*$                                & %
	{\hspace{-1.5cm}\databartwo{1.00}{0.45}\makebox[0pt][c]{\hspace{1cm}23 / 51}} &
	{\databar{1.00}}	360.04 \pm 0.03 &
	{\databar{0.28}}	337.69 \pm 70.68 &
	{\databar{0.09}}	0.25 \pm 0.00 &
	{\databar{0.58}}	7.50 \pm 2.15 &
	{\databar{0.03}}	43 %
\\
SORRT${}^*$                              & %
	{\hspace{-1.5cm}\databartwo{1.00}{0.35}\makebox[0pt][c]{\hspace{1cm}18 / 51}} &
	{\databar{1.00}}	360.04 \pm 0.04 &
	{\databar{0.29}}	347.88 \pm 52.29 &
	{\databar{0.09}}	0.25 \pm 0.01 &
	{\databar{0.61}}	7.83 \pm 2.24 &
	{\databar{0.04}}	59 %
\\
SPARS                                    & %
	{\hspace{-1.5cm}\databartwo{0.57}{0.00}\makebox[0pt][c]{\hspace{1cm}0 / 29}} &
	{\databar{1.00}}	360.80 \pm 0.77 &
	{\databar{0.47}}	569.63 \pm 83.17 &
	{\databar{0.09}}	0.25 \pm 0.00 &
	{\databar{0.73}}	9.43 \pm 1.60 &
	{\databar{0.07}}	104 %
\\
SPARS2                                   & %
	{\hspace{-1.5cm}\databartwo{0.59}{0.00}\makebox[0pt][c]{\hspace{1cm}0 / 30}} &
	{\databar{1.00}}	360.01 \pm 0.01 &
	{\databar{0.42}}	504.95 \pm 80.48 &
	{\databar{0.09}}	0.25 \pm 0.00 &
	{\databar{0.73}}	9.39 \pm 1.51 &
	{\databar{0.11}}	155 %
\\
SST                                      & %
	{\hspace{-1.5cm}\databartwo{0.96}{0.76}\makebox[0pt][c]{\hspace{1cm}39 / 49}} &
	{\databar{1.00}}	360.02 \pm 0.02 &
	{\databar{0.49}}	593.49 \pm 637.50 &
	{\databar{0.10}}	0.27 \pm 0.11 &
	{\databar{0.83}}	10.75 \pm 3.38 &
	{\databar{1.00}}	1375 %
\\
Theta${}^*$                              & %
	{\hspace{-1.5cm}\databartwo{0.27}{0.27}\makebox[0pt][c]{\hspace{1cm}14 / 14}} &
	{\databar{0.49}}	177.62 \pm 101.84 &
	{\databar{0.38}}	454.41 \pm 43.19 &
	{\databar{0.09}}	0.25 \pm 0.00 &
	{\databar{0.60}}	7.66 \pm 1.55 &
	{\databar{0.05}}	70 %
\\
\bottomrule
\end{tabular}
    }
    \caption{Planning statistics using different steer functions from the \texttt{Berlin\_0\_256} scenario from the Moving AI benchmark.}
    \label{tab:moving_ai_berlin}
\end{table*}

\section{General Observations}
\label{sec:overall}
Based on the results detailed in \autoref{sec:results}, in this section we provide a general analysis across the experiments and give specific recommendations.

\subsection{Planning Time}
Feasible planners are much faster and reliable in finding a single solution. RRT consistently ranked  among the fastest of the planners we evaluated. While anytime, i.e. asymptotically optimal, planners require more time to find solutions, these are of higher quality than the paths found by feasible planners. The complexity of the steer function also severely impacts the performance of the planners. Dubins curves, for example, are computationally challenging systems for planning in very cluttered environments. An added burden on the runtime complexity stems from the polygon-based collision model, that, in contrast to typical point-based collision checkers, further penalizes algorithms that are not implemented in a way to make as few state validity checks as possible, such as our non-optimized Theta${}^*$ implementation. Collision checking often consumed most of the allotted planning time such that this planner, in many cases, did not find any solution.

\subsection{Quality of Anytime Solutions}
Overall the results confirm what we know from the theory: on average, anytime planners obtain better solutions in terms of path length, number of cusps and maximum curvature. Informed anytime approaches (e.g., Informed-RRT*, SORRT*, BIT*) can achieve sometimes shorter paths throughout all tested steer functions. However, this is not always the case. These approaches are still impacted by larger complexity in the environment, and do not perform faster in highly constrained environments.

\subsection{Variability of the Results}
The main concern regarding sampling-based planners (feasible and anytime ones) is the high variance of the obtained results, which may lead also occasionally to low performance.  In particular, we believe that the stochasticity of the sampling phase is a major drawback that should be addressed from the community. Deterministic sampling \cite{janson2018deterministic,yershova2004deterministic,palmieriRAL2019} is an approach that mitigates this issue. State-lattice planners are an example of deterministic techniques, which, at the price of the solution quality, offer deterministic performance.

\subsection{Post-smoothing Synergies}
\label{sec:overall-smoothing}
Post-smoothing combined with feasible planners is a good strategy in terms of planning efficiency and final path quality (sub-optimal and may not completely fulfill kinodynamic requirements). The results show that there exist several couplings of feasible sampling-based planners and post-smoothers that outperform anytime planners both in computation time and solution quality.

\subsection{Environment Complexity}
Our benchmarking confirms that the environments significantly influence the performance of the planners.  Environments, such as the polygon-based warehouse scenarios, revealed vastly different solutions between the planners (see \autoref{fig:warehouse_trajectories}).

As pointed out in Section \ref{sec:procedural}, the planning performance is further impacted by different environment characteristics, such as narrow corridors and spaces cluttered with small obstacles. Plain state-of-the-art approaches that do not implement additional sampling heuristics, such as goal biasing, often fail to return solutions in very difficult environments where the corridors are small or the obstacle density is high. 

\subsection{Influence of the Steer Function}
Regarding the steer functions, we have observed two main phenomena which confirm previous theoretical claims~\cite{laumond1998guidelines}. Computationally complex steer functions, such as CC Reeds-Shepp, severly impact the planning efficiency of all the algorithms. Solving planning queries for systems with complex nonholonomic constraints in very cluttered environments also requires more planning time, i.e. particularly for systems which are not small-time locally controllable, such as Dubins curves. On the larger-scale experiments (e.g. \autoref{sec:moving-ai}) we observed a significant variance in the planning time allotment necessary for the planners to find solutions with different steer functions, ranging from \SI{1.5}{\minute} (Reeds-Shepp) to more than \SI{18}{\minute} (CC Reeds-Shepp).

\section{Conclusion}
Following the need for more reproducible evaluations of commonly used AI algorithms, and with the goal of comparing a large set of state-of-the-art motion planning techniques, the presented paper establishes a benchmark for motion planners that focuses on problems with nonholonomic systems, in particular wheeled mobile robots. From our experiments, we draw guidelines and highlight use-cases that are close to real-world scenarios of autonomous navigation systems. We are planning to open-source the data and implementation of the benchmarking framework, including the tooling to reproduce all presented results.

\section*{ACKNOWLEDGMENTS}

The authors thank Ziang Liu for his contributions to the software repository and testing of various algorithms.

\bibliographystyle{IEEEtran}
\footnotesize{
\bibliography{literature}
}

\ifextended
\thispagestyle{empty}
\onecolumn
\section*{Appendix}

\begin{table*}[htp]
    \centering\footnotesize
    \resizebox{.9\textwidth}{!}{%
        \begin{tabular}{p{2cm}>{\centering}m{1.2\maxlen}*{5}{S[table-format=3.2(8)]}}
\toprule
\bf Planner & \mcc{Solutions} & \mcc{Time [s]} & \mcc{Path~Length} & \mcc{Curvature} & \mcc{Clearance} & \mcc{Cusps} \\\midrule
\rowlabel{\textbf{Scenario: \texttt{NewYork\_1\_512}} (SBPL, \SI{20}{\minute} time limit)}
\\
SBPL AD${}^*$                            & %
	0 &
	N / A&
	N / A&
	N / A&
	N / A&
	N / A%
\\
SBPL ARA${}^*$                            & %
	0 &
	N / A&
	N / A&
	N / A&
	N / A&
	N / A%
\\
SBPL MHA$^*$                                 & %
	{\hspace{-1.5cm}\databartwo{0.24}{0.24}\makebox[0pt][c]{\hspace{1cm}12 / 12}} &
	{\databar{0.02}}	18.10 \pm 19.29 &
	{\databar{0.34}}	780.59 \pm 14.66 &
	{\databar{0.52}}	0.77 \pm 0.05 &
	{\databar{0.45}}	8.59 \pm 0.74 &
	{\databar{0.09}}	59 %
\\
\rowlabel{\textbf{Scenario: \texttt{NewYork\_1\_512}} (Reeds-Shepp steering, \SI{2.5}{\minute} time limit)}
\\
BFMT                                     & %
	{\hspace{-1.5cm}\databartwo{1.00}{1.00}\makebox[0pt][c]{\hspace{1cm}51 / 51}} &
	{\databar{0.01}}	0.94 \pm 0.48 &
	{\databar{0.32}}	717.88 \pm 13.39 &
	{\databar{0.68}}	0.83 \pm 0.45 &
	{\databar{0.84}}	11.81 \pm 1.22 &
	{\databar{0.11}}	275 %
\\
BIT${}^*$                                & %
	{\hspace{-1.5cm}\databartwo{1.00}{0.22}\makebox[0pt][c]{\hspace{1cm}11 / 51}} &
	{\databar{0.99}}	148.34 \pm 12.48 &
	{\databar{0.32}}	711.51 \pm 12.96 &
	{\databar{0.47}}	0.58 \pm 0.52 &
	{\databar{0.76}}	10.67 \pm 1.72 &
	{\databar{0.04}}	100 %
\\
CForest                                  & %
	{\hspace{-1.5cm}\databartwo{1.00}{0.02}\makebox[0pt][c]{\hspace{1cm}1 / 51}} &
	{\databar{0.99}}	148.26 \pm 12.48 &
	{\databar{0.31}}	685.82 \pm 6.93 &
	{\databar{0.26}}	0.32 \pm 0.26 &
	{\databar{0.68}}	9.53 \pm 1.56 &
	{\databar{0.02}}	40 %
\\
EST                                      & %
	{\hspace{-1.5cm}\databartwo{1.00}{0.88}\makebox[0pt][c]{\hspace{1cm}45 / 51}} &
	{\databar{0.01}}	0.86 \pm 1.15 &
	{\databar{0.51}}	1125.24 \pm 222.17 &
	{\databar{0.72}}	0.88 \pm 0.24 &
	{\databar{0.93}}	13.02 \pm 2.10 &
	{\databar{0.35}}	839 %
\\
Informed RRT${}^*$                       & %
	{\hspace{-1.5cm}\databartwo{1.00}{0.06}\makebox[0pt][c]{\hspace{1cm}3 / 51}} &
	{\databar{0.99}}	148.25 \pm 12.48 &
	{\databar{0.31}}	691.68 \pm 7.08 &
	{\databar{0.39}}	0.47 \pm 0.44 &
	{\databar{0.70}}	9.84 \pm 1.54 &
	{\databar{0.03}}	72 %
\\
KPIECE                                   & %
	{\hspace{-1.5cm}\databartwo{1.00}{0.84}\makebox[0pt][c]{\hspace{1cm}43 / 51}} &
	{\databar{0.00}}	0.67 \pm 1.60 &
	{\databar{1.00}}	2215.37 \pm 706.02 &
	{\databar{0.46}}	0.57 \pm 0.21 &
	{\databar{0.94}}	13.12 \pm 2.17 &
	{\databar{0.83}}	2016 %
\\
PDST                                     & %
	{\hspace{-1.5cm}\databartwo{1.00}{0.71}\makebox[0pt][c]{\hspace{1cm}36 / 51}} &
	{\databar{0.10}}	14.83 \pm 37.00 &
	{\databar{0.44}}	982.94 \pm 275.41 &
	{\databar{0.69}}	0.84 \pm 0.35 &
	{\databar{0.88}}	12.36 \pm 1.71 &
	{\databar{0.23}}	560 %
\\
PRM                                      & %
	{\hspace{-1.5cm}\databartwo{1.00}{0.65}\makebox[0pt][c]{\hspace{1cm}33 / 51}} &
	{\databar{0.99}}	148.33 \pm 12.49 &
	{\databar{0.33}}	722.12 \pm 10.27 &
	{\databar{0.77}}	0.95 \pm 0.47 &
	{\databar{0.82}}	11.56 \pm 1.73 &
	{\databar{0.16}}	393 %
\\
PRM${}^*$                                & %
	{\hspace{-1.5cm}\databartwo{1.00}{0.69}\makebox[0pt][c]{\hspace{1cm}35 / 51}} &
	{\databar{0.99}}	148.34 \pm 12.49 &
	{\databar{0.32}}	706.14 \pm 9.38 &
	{\databar{0.54}}	0.66 \pm 0.47 &
	{\databar{0.81}}	11.34 \pm 1.65 &
	{\databar{0.06}}	154 %
\\
RRT                                      & %
	{\hspace{-1.5cm}\databartwo{1.00}{0.92}\makebox[0pt][c]{\hspace{1cm}47 / 51}} &
	{\databar{0.00}}	0.59 \pm 1.20 &
	{\databar{0.40}}	890.31 \pm 144.49 &
	{\databar{0.73}}	0.90 \pm 0.34 &
	{\databar{0.88}}	12.29 \pm 1.87 &
	{\databar{0.26}}	621 %
\\
RRT\#                                    & %
	{\hspace{-1.5cm}\databartwo{1.00}{0.47}\makebox[0pt][c]{\hspace{1cm}24 / 51}} &
	{\databar{0.99}}	148.31 \pm 12.49 &
	{\databar{0.31}}	690.45 \pm 6.77 &
	{\databar{0.35}}	0.44 \pm 0.39 &
	{\databar{0.76}}	10.68 \pm 1.44 &
	{\databar{0.03}}	67 %
\\
RRT${}^*$                                & %
	{\hspace{-1.5cm}\databartwo{1.00}{0.33}\makebox[0pt][c]{\hspace{1cm}17 / 51}} &
	{\databar{0.99}}	148.25 \pm 12.48 &
	{\databar{0.31}}	689.28 \pm 6.10 &
	{\databar{0.38}}	0.47 \pm 0.41 &
	{\databar{0.75}}	10.52 \pm 1.46 &
	{\databar{0.03}}	73 %
\\
SORRT${}^*$                              & %
	{\hspace{-1.5cm}\databartwo{1.00}{0.04}\makebox[0pt][c]{\hspace{1cm}2 / 51}} &
	{\databar{0.99}}	148.24 \pm 12.48 &
	{\databar{0.31}}	690.31 \pm 6.34 &
	{\databar{0.31}}	0.37 \pm 0.35 &
	{\databar{0.71}}	9.93 \pm 1.76 &
	{\databar{0.02}}	46 %
\\
SPARS                                    & %
	{\hspace{-1.5cm}\databartwo{1.00}{0.92}\makebox[0pt][c]{\hspace{1cm}47 / 51}} &
	{\databar{0.99}}	148.94 \pm 12.60 &
	{\databar{0.45}}	989.52 \pm 212.95 &
	{\databar{0.73}}	0.90 \pm 0.31 &
	{\databar{1.00}}	14.03 \pm 3.49 &
	{\databar{0.24}}	573 %
\\
SPARS2                                   & %
	{\hspace{-1.5cm}\databartwo{1.00}{0.65}\makebox[0pt][c]{\hspace{1cm}33 / 51}} &
	{\databar{0.99}}	148.24 \pm 12.48 &
	{\databar{0.34}}	756.76 \pm 23.56 &
	{\databar{1.00}}	1.23 \pm 0.48 &
	{\databar{0.83}}	11.66 \pm 1.28 &
	{\databar{0.21}}	511 %
\\
SST                                      & %
	{\hspace{-1.5cm}\databartwo{1.00}{0.82}\makebox[0pt][c]{\hspace{1cm}42 / 51}} &
	{\databar{0.99}}	148.24 \pm 12.47 &
	{\databar{0.44}}	966.53 \pm 83.01 &
	{\databar{0.96}}	1.17 \pm 0.19 &
	{\databar{0.68}}	9.52 \pm 1.26 &
	{\databar{1.00}}	2424 %
\\
\rowlabel{\textbf{Scenario: \texttt{NewYork\_1\_512}} (CC Reeds-Shepp steering, \SI{15}{\minute} time limit)}
\\
\\
BFMT                                     & %
	{\hspace{-1.5cm}\databartwo{1.00}{0.98}\makebox[0pt][c]{\hspace{1cm}50 / 51}} &
	{\databar{0.04}}	32.40 \pm 6.97 &
	{\databar{0.35}}	717.22 \pm 15.37 &
	{\databar{0.47}}	0.41 \pm 0.41 &
	{\databar{0.89}}	12.04 \pm 1.82 &
	{\databar{0.04}}	135 %
\\
BIT${}^*$                                & %
	{\hspace{-1.5cm}\databartwo{0.51}{0.20}\makebox[0pt][c]{\hspace{1cm}10 / 26}} &
	{\databar{0.91}}	817.27 \pm 194.97 &
	{\databar{0.36}}	744.30 \pm 20.77 &
	{\databar{0.51}}	0.45 \pm 0.39 &
	{\databar{0.73}}	9.87 \pm 1.66 &
	{\databar{0.06}}	98 %
\\
CForest                                  & %
	{\hspace{-1.5cm}\databartwo{1.00}{0.00}\makebox[0pt][c]{\hspace{1cm}0 / 51}} &
	{\databar{0.01}}	7.05 \pm 20.65 &
	{\databar{0.35}}	710.01 \pm 86.08 &
	{\databar{0.57}}	0.50 \pm 0.46 &
	{\databar{0.70}}	9.48 \pm 2.57 &
	{\databar{0.04}}	127 %
\\
EST                                      & %
	{\hspace{-1.5cm}\databartwo{0.94}{0.92}\makebox[0pt][c]{\hspace{1cm}47 / 48}} &
	{\databar{0.07}}	65.22 \pm 46.91 &
	{\databar{0.56}}	1143.59 \pm 213.58 &
	{\databar{0.87}}	0.77 \pm 0.29 &
	{\databar{0.87}}	11.76 \pm 2.08 &
	{\databar{0.42}}	1360 %
\\
Informed RRT${}^*$                       & %
	{\hspace{-1.5cm}\databartwo{0.90}{0.37}\makebox[0pt][c]{\hspace{1cm}19 / 46}} &
	{\databar{0.91}}	818.04 \pm 193.96 &
	{\databar{0.34}}	698.21 \pm 19.82 &
	{\databar{0.42}}	0.37 \pm 0.40 &
	{\databar{0.81}}	10.96 \pm 1.66 &
	{\databar{0.03}}	93 %
\\
KPIECE                                   & %
	{\hspace{-1.5cm}\databartwo{0.94}{0.80}\makebox[0pt][c]{\hspace{1cm}41 / 48}} &
	{\databar{0.11}}	102.45 \pm 176.41 &
	{\databar{1.00}}	2044.73 \pm 551.96 &
	{\databar{0.73}}	0.65 \pm 0.38 &
	{\databar{0.91}}	12.27 \pm 2.70 &
	{\databar{1.00}}	3204 %
\\
PDST                                     & %
	{\hspace{-1.5cm}\databartwo{0.94}{0.73}\makebox[0pt][c]{\hspace{1cm}37 / 48}} &
	{\databar{0.24}}	213.28 \pm 325.13 &
	{\databar{0.46}}	944.86 \pm 209.86 &
	{\databar{0.87}}	0.77 \pm 0.50 &
	{\databar{0.90}}	12.23 \pm 1.57 &
	{\databar{0.18}}	578 %
\\
PRM                                      & %
	{\hspace{-1.5cm}\databartwo{0.82}{0.76}\makebox[0pt][c]{\hspace{1cm}39 / 42}} &
	{\databar{0.88}}	789.64 \pm 232.76 &
	{\databar{0.37}}	761.86 \pm 30.03 &
	{\databar{0.94}}	0.82 \pm 0.51 &
	{\databar{0.89}}	11.99 \pm 2.00 &
	{\databar{0.11}}	307 %
\\
PRM${}^*$                                & %
	{\hspace{-1.5cm}\databartwo{0.84}{0.69}\makebox[0pt][c]{\hspace{1cm}35 / 43}} &
	{\databar{0.89}}	804.51 \pm 202.57 &
	{\databar{0.36}}	742.29 \pm 22.96 &
	{\databar{0.66}}	0.58 \pm 0.51 &
	{\databar{0.87}}	11.81 \pm 1.96 &
	{\databar{0.08}}	239 %
\\
RRT                                      & %
	{\hspace{-1.5cm}\databartwo{0.94}{0.80}\makebox[0pt][c]{\hspace{1cm}41 / 48}} &
	{\databar{0.02}}	19.81 \pm 39.98 &
	{\databar{0.45}}	914.38 \pm 119.77 &
	{\databar{0.76}}	0.67 \pm 0.38 &
	{\databar{0.87}}	11.72 \pm 1.50 &
	{\databar{0.15}}	491 %
\\
RRT\#                                    & %
	{\hspace{-1.5cm}\databartwo{0.75}{0.73}\makebox[0pt][c]{\hspace{1cm}37 / 38}} &
	{\databar{0.89}}	801.79 \pm 209.42 &
	{\databar{0.34}}	703.77 \pm 15.21 &
	{\databar{0.41}}	0.36 \pm 0.39 &
	{\databar{0.86}}	11.70 \pm 1.41 &
	{\databar{0.02}}	56 %
\\
RRT${}^*$                                & %
	{\hspace{-1.5cm}\databartwo{0.86}{0.84}\makebox[0pt][c]{\hspace{1cm}43 / 44}} &
	{\databar{0.90}}	814.42 \pm 197.41 &
	{\databar{0.34}}	701.44 \pm 13.19 &
	{\databar{0.27}}	0.24 \pm 0.23 &
	{\databar{0.86}}	11.64 \pm 1.31 &
	{\databar{0.02}}	50 %
\\
SORRT${}^*$                              & %
	{\hspace{-1.5cm}\databartwo{0.14}{0.08}\makebox[0pt][c]{\hspace{1cm}4 / 7}} &
	{\databar{0.91}}	823.17 \pm 188.54 &
	{\databar{0.32}}	653.14 \pm 25.20 &
	{\databar{0.38}}	0.33 \pm 0.34 &
	{\databar{0.77}}	10.45 \pm 1.67 &
	{\databar{0.03}}	12 %
\\
SPARS                                    & %
	{\hspace{-1.5cm}\databartwo{0.71}{0.71}\makebox[0pt][c]{\hspace{1cm}36 / 36}} &
	{\databar{0.97}}	872.82 \pm 124.32 &
	{\databar{0.48}}	972.45 \pm 234.22 &
	{\databar{0.68}}	0.60 \pm 0.35 &
	{\databar{1.00}}	13.53 \pm 0.87 &
	{\databar{0.12}}	335 %
\\
SPARS2                                   & %
	0 &
	N / A&
	N / A&
	N / A&
	N / A&
	N / A%
\\
SST                                      & %
	{\hspace{-1.5cm}\databartwo{0.78}{0.76}\makebox[0pt][c]{\hspace{1cm}39 / 40}} &
	{\databar{0.90}}	805.53 \pm 205.18 &
	{\databar{0.51}}	1046.15 \pm 152.20 &
	{\databar{1.00}}	0.88 \pm 0.52 &
	{\databar{0.69}}	9.39 \pm 1.14 &
	{\databar{0.87}}	2323 %
\\
\rowlabel{\textbf{Scenario: \texttt{NewYork\_1\_512}} (POSQ steering, \SI{20}{\minute} time limit)}
\\
BFMT                                     & %
	0 &
	N / A&
	N / A&
	N / A&
	N / A&
	N / A%
\\
BIT${}^*$                                & %
	{\hspace{-1.5cm}\databartwo{0.04}{0.04}\makebox[0pt][c]{\hspace{1cm}2 / 2}} &
	{\databar{0.12}}	142.16 \pm 135.70 &
	{\databar{0.36}}	823.76 \pm 100.96 &
	{\databar{0.41}}	0.61 \pm 0.15 &
	{\databar{0.88}}	16.68 \pm 2.78 &
	{\databar{0.01}}	10 %
\\
CForest                                  & %
	{\hspace{-1.5cm}\databartwo{0.84}{0.57}\makebox[0pt][c]{\hspace{1cm}29 / 43}} &
	{\databar{0.23}}	272.59 \pm 333.81 &
	{\databar{0.21}}	472.46 \pm 332.60 &
	{\databar{0.59}}	0.87 \pm 0.54 &
	{\databar{0.94}}	17.79 \pm 6.51 &
	{\databar{0.08}}	143 %
\\
EST                                      & %
	{\hspace{-1.5cm}\databartwo{0.92}{0.92}\makebox[0pt][c]{\hspace{1cm}47 / 47}} &
	{\databar{0.95}}	1138.81 \pm 200.91 &
	{\databar{0.09}}	204.71 \pm 153.09 &
	{\databar{0.84}}	1.23 \pm 0.48 &
	{\databar{0.93}}	17.64 \pm 8.57 &
	{\databar{0.48}}	908 %
\\
Informed RRT${}^*$                       & %
	{\hspace{-1.5cm}\databartwo{0.90}{0.86}\makebox[0pt][c]{\hspace{1cm}44 / 46}} &
	{\databar{0.91}}	1088.31 \pm 297.43 &
	{\databar{0.11}}	243.21 \pm 214.34 &
	{\databar{0.56}}	0.82 \pm 0.52 &
	{\databar{0.84}}	15.95 \pm 7.24 &
	{\databar{0.51}}	937 %
\\
KPIECE                                   & %
	{\hspace{-1.5cm}\databartwo{1.00}{0.18}\makebox[0pt][c]{\hspace{1cm}9 / 51}} &
	{\databar{0.02}}	29.08 \pm 36.21 &
	{\databar{1.00}}	2282.64 \pm 855.03 &
	{\databar{0.46}}	0.68 \pm 0.30 &
	{\databar{0.64}}	12.04 \pm 2.25 &
	{\databar{1.00}}	2033 %
\\
PDST                                     & %
	{\hspace{-1.5cm}\databartwo{0.96}{0.12}\makebox[0pt][c]{\hspace{1cm}6 / 49}} &
	{\databar{0.28}}	337.38 \pm 454.74 &
	{\databar{0.46}}	1055.27 \pm 420.35 &
	{\databar{0.98}}	1.44 \pm 0.76 &
	{\databar{0.59}}	11.23 \pm 3.66 &
	{\databar{0.76}}	1488 %
\\
PRM                                      & %
	{\hspace{-1.5cm}\databartwo{0.75}{0.20}\makebox[0pt][c]{\hspace{1cm}10 / 38}} &
	{\databar{0.76}}	911.31 \pm 404.37 &
	{\databar{0.50}}	1137.52 \pm 1130.91 &
	{\databar{1.00}}	1.47 \pm 0.68 &
	{\databar{0.43}}	8.05 \pm 2.56 &
	{\databar{0.79}}	1454 %
\\
PRM${}^*$                                & %
	{\hspace{-1.5cm}\databartwo{0.78}{0.10}\makebox[0pt][c]{\hspace{1cm}5 / 40}} &
	{\databar{0.55}}	658.31 \pm 491.15 &
	{\databar{0.44}}	1012.05 \pm 781.99 &
	{\databar{0.73}}	1.08 \pm 0.55 &
	{\databar{0.46}}	8.68 \pm 2.45 &
	{\databar{0.70}}	1259 %
\\
RRT                                      & %
	{\hspace{-1.5cm}\databartwo{0.88}{0.84}\makebox[0pt][c]{\hspace{1cm}43 / 45}} &
	{\databar{0.86}}	1032.53 \pm 343.26 &
	{\databar{0.27}}	617.91 \pm 419.35 &
	{\databar{0.61}}	0.89 \pm 0.40 &
	{\databar{1.00}}	18.89 \pm 6.17 &
	{\databar{0.22}}	392 %
\\
RRT\#                                    & %
	{\hspace{-1.5cm}\databartwo{0.92}{0.69}\makebox[0pt][c]{\hspace{1cm}35 / 47}} &
	{\databar{0.87}}	1044.79 \pm 328.51 &
	{\databar{0.14}}	320.29 \pm 264.92 &
	{\databar{0.58}}	0.85 \pm 0.50 &
	{\databar{0.87}}	16.52 \pm 8.05 &
	{\databar{0.44}}	831 %
\\
RRT${}^*$                                & %
	{\hspace{-1.5cm}\databartwo{0.88}{0.82}\makebox[0pt][c]{\hspace{1cm}42 / 45}} &
	{\databar{0.86}}	1035.95 \pm 336.34 &
	{\databar{0.14}}	317.83 \pm 265.95 &
	{\databar{0.58}}	0.85 \pm 0.50 &
	{\databar{0.85}}	16.04 \pm 5.93 &
	{\databar{0.58}}	1046 %
\\
SORRT${}^*$                              & %
	{\hspace{-1.5cm}\databartwo{0.92}{0.90}\makebox[0pt][c]{\hspace{1cm}46 / 47}} &
	{\databar{0.91}}	1090.37 \pm 295.73 &
	{\databar{0.10}}	228.19 \pm 220.97 &
	{\databar{0.60}}	0.88 \pm 0.51 &
	{\databar{0.83}}	15.61 \pm 6.61 &
	{\databar{0.60}}	1116 %
\\
SPARS                                    & %
	{\hspace{-1.5cm}\databartwo{0.47}{0.00}\makebox[0pt][c]{\hspace{1cm}0 / 24}} &
	{\databar{0.19}}	228.47 \pm 169.73 &
	{\databar{0.40}}	924.12 \pm 136.62 &
	{\databar{0.55}}	0.81 \pm 0.18 &
	{\databar{0.66}}	12.50 \pm 0.92 &
	{\databar{0.11}}	224 %
\\
SPARS2                                   & %
	{\hspace{-1.5cm}\databartwo{0.63}{0.06}\makebox[0pt][c]{\hspace{1cm}3 / 32}} &
	{\databar{0.24}}	286.44 \pm 339.74 &
	{\databar{0.52}}	1177.41 \pm 481.27 &
	{\databar{0.62}}	0.91 \pm 0.64 &
	{\databar{0.59}}	11.10 \pm 2.07 &
	{\databar{0.18}}	353 %
\\
SST                                      & %
	0 &
	N / A&
	N / A&
	N / A&
	N / A&
	N / A%
\\
\rowlabel{\textbf{Scenario: \texttt{NewYork\_1\_512}} (Dubins steering, \SI{10}{\minute} time limit)}
\\
BFMT                                     & %
	{\hspace{-1.5cm}\databartwo{0.88}{0.04}\makebox[0pt][c]{\hspace{1cm}2 / 45}} &
	{\databar{0.06}}	38.07 \pm 103.58 &
	{\databar{0.39}}	831.61 \pm 53.59 &
	{\databar{0.33}}	0.25 \pm 0.01 &
	{\databar{0.74}}	11.18 \pm 1.74 &
	{\databar{0.19}}	343 %
\\
BIT${}^*$                                & %
	{\hspace{-1.5cm}\databartwo{0.61}{0.22}\makebox[0pt][c]{\hspace{1cm}11 / 31}} &
	{\databar{1.00}}	600.20 \pm 0.32 &
	{\databar{0.34}}	722.41 \pm 25.88 &
	{\databar{0.33}}	0.25 \pm 0.00 &
	{\databar{0.72}}	10.86 \pm 1.92 &
	{\databar{0.02}}	39 %
\\
CForest                                  & %
	{\hspace{-1.5cm}\databartwo{0.98}{0.06}\makebox[0pt][c]{\hspace{1cm}3 / 50}} &
	{\databar{0.97}}	583.29 \pm 82.31 &
	{\databar{0.33}}	696.20 \pm 15.79 &
	{\databar{0.33}}	0.25 \pm 0.01 &
	{\databar{0.65}}	9.84 \pm 1.53 &
	{\databar{0.02}}	33 %
\\
EST                                      & %
	{\hspace{-1.5cm}\databartwo{1.00}{0.57}\makebox[0pt][c]{\hspace{1cm}29 / 51}} &
	{\databar{0.13}}	77.11 \pm 181.12 &
	{\databar{0.58}}	1228.45 \pm 284.31 &
	{\databar{0.33}}	0.25 \pm 0.01 &
	{\databar{0.86}}	12.96 \pm 2.20 &
	{\databar{0.21}}	383 %
\\
Informed RRT${}^*$                       & %
	{\hspace{-1.5cm}\databartwo{1.00}{0.49}\makebox[0pt][c]{\hspace{1cm}25 / 51}} &
	{\databar{0.97}}	583.58 \pm 81.52 &
	{\databar{0.33}}	703.05 \pm 23.47 &
	{\databar{0.32}}	0.25 \pm 0.03 &
	{\databar{0.73}}	11.01 \pm 1.91 &
	{\databar{0.03}}	64 %
\\
KPIECE                                   & %
	{\hspace{-1.5cm}\databartwo{1.00}{0.55}\makebox[0pt][c]{\hspace{1cm}28 / 51}} &
	{\databar{0.07}}	39.39 \pm 141.51 &
	{\databar{1.00}}	2114.74 \pm 791.26 &
	{\databar{0.33}}	0.25 \pm 0.01 &
	{\databar{1.00}}	15.16 \pm 3.54 &
	{\databar{1.00}}	1837 %
\\
PDST                                     & %
	{\hspace{-1.5cm}\databartwo{0.96}{0.65}\makebox[0pt][c]{\hspace{1cm}33 / 49}} &
	{\databar{0.06}}	38.43 \pm 121.62 &
	{\databar{0.43}}	901.96 \pm 214.32 &
	{\databar{0.32}}	0.25 \pm 0.02 &
	{\databar{0.87}}	13.26 \pm 2.72 &
	{\databar{0.08}}	148 %
\\
PRM                                      & %
	{\hspace{-1.5cm}\databartwo{1.00}{0.00}\makebox[0pt][c]{\hspace{1cm}0 / 51}} &
	{\databar{0.97}}	583.58 \pm 81.53 &
	{\databar{0.41}}	864.47 \pm 61.27 &
	{\databar{0.33}}	0.25 \pm 0.01 &
	{\databar{0.70}}	10.57 \pm 1.76 &
	{\databar{0.22}}	408 %
\\
PRM${}^*$                                & %
	{\hspace{-1.5cm}\databartwo{0.98}{0.00}\makebox[0pt][c]{\hspace{1cm}0 / 50}} &
	{\databar{0.97}}	583.27 \pm 82.31 &
	{\databar{0.39}}	830.02 \pm 61.61 &
	{\databar{0.33}}	0.25 \pm 0.00 &
	{\databar{0.70}}	10.53 \pm 1.47 &
	{\databar{0.19}}	342 %
\\
RRT                                      & %
	{\hspace{-1.5cm}\databartwo{1.00}{0.80}\makebox[0pt][c]{\hspace{1cm}41 / 51}} &
	{\databar{0.06}}	36.65 \pm 140.97 &
	{\databar{0.46}}	968.59 \pm 216.20 &
	{\databar{0.33}}	0.25 \pm 0.01 &
	{\databar{0.86}}	13.03 \pm 2.16 &
	{\databar{0.10}}	190 %
\\
RRT\#                                    & %
	{\hspace{-1.5cm}\databartwo{1.00}{0.00}\makebox[0pt][c]{\hspace{1cm}0 / 51}} &
	{\databar{0.97}}	583.58 \pm 81.53 &
	{\databar{0.33}}	694.88 \pm 17.03 &
	{\databar{0.32}}	0.24 \pm 0.03 &
	{\databar{0.66}}	10.00 \pm 1.67 &
	{\databar{0.01}}	27 %
\\
RRT${}^*$                                & %
	{\hspace{-1.5cm}\databartwo{0.98}{0.24}\makebox[0pt][c]{\hspace{1cm}12 / 50}} &
	{\databar{0.97}}	583.24 \pm 82.30 &
	{\databar{0.33}}	697.30 \pm 19.37 &
	{\databar{0.32}}	0.25 \pm 0.02 &
	{\databar{0.71}}	10.76 \pm 1.56 &
	{\databar{0.01}}	24 %
\\
SORRT${}^*$                              & %
	{\hspace{-1.5cm}\databartwo{1.00}{0.37}\makebox[0pt][c]{\hspace{1cm}19 / 51}} &
	{\databar{0.97}}	583.58 \pm 81.53 &
	{\databar{0.33}}	703.81 \pm 24.36 &
	{\databar{0.32}}	0.25 \pm 0.02 &
	{\databar{0.74}}	11.18 \pm 1.90 &
	{\databar{0.02}}	44 %
\\
SPARS                                    & %
	{\hspace{-1.5cm}\databartwo{0.76}{0.00}\makebox[0pt][c]{\hspace{1cm}0 / 39}} &
	{\databar{0.99}}	591.86 \pm 66.57 &
	{\databar{0.53}}	1115.08 \pm 192.55 &
	{\databar{0.33}}	0.25 \pm 0.00 &
	{\databar{0.78}}	11.78 \pm 1.43 &
	{\databar{0.16}}	292 %
\\
SPARS2                                   & %
	{\hspace{-1.5cm}\databartwo{0.69}{0.00}\makebox[0pt][c]{\hspace{1cm}0 / 35}} &
	{\databar{0.98}}	588.01 \pm 69.97 &
	{\databar{0.43}}	913.56 \pm 117.60 &
	{\databar{0.32}}	0.25 \pm 0.02 &
	{\databar{0.69}}	10.48 \pm 1.70 &
	{\databar{0.13}}	247 %
\\
SST                                      & %
	{\hspace{-1.5cm}\databartwo{0.98}{0.80}\makebox[0pt][c]{\hspace{1cm}41 / 50}} &
	{\databar{0.97}}	583.23 \pm 82.31 &
	{\databar{0.49}}	1028.63 \pm 344.21 &
	{\databar{0.33}}	0.25 \pm 0.00 &
	{\databar{0.74}}	11.17 \pm 2.09 &
	{\databar{0.78}}	1438 %
\\
\bottomrule
\end{tabular}
    }
    \caption{Planning statistics using different steer functions from the \texttt{NewYork\_1\_512} scenario from the Moving AI benchmark.}
    \label{tab:moving_ai_newyork}
\end{table*}

\begin{table*}[htp]
    \centering
    \resizebox{.9\textwidth}{!}{%
        \begin{tabular}{p{2cm}>{\centering}m{1.2\maxlen}*{5}{S[table-format=3.2(8)]}}
\toprule
\bf Planner & \mcc{Solutions} & \mcc{Time [s]} & \mcc{Path~Length} & \mcc{Curvature} & \mcc{Clearance} & \mcc{Cusps} \\\midrule
\rowlabel{\textbf{Scenario: \texttt{Boston\_1\_1024}} (Reeds-Shepp steering, \SI{7.5}{\minute} time limit)}
\\
BFMT                                     & %
	{\hspace{-1.5cm}\databartwo{1.00}{0.94}\makebox[0pt][c]{\hspace{1cm}48 / 51}} &
	{\databar{0.00}}	2.20 \pm 8.52 &
	{\databar{0.38}}	1521.12 \pm 31.06 &
	{\databar{0.75}}	0.59 \pm 0.16 &
	{\databar{0.74}}	24.95 \pm 3.04 &
	{\databar{0.14}}	430 %
\\
BIT${}^*$                                & %
	{\hspace{-1.5cm}\databartwo{0.98}{0.24}\makebox[0pt][c]{\hspace{1cm}12 / 50}} &
	{\databar{1.00}}	450.09 \pm 0.17 &
	{\databar{0.36}}	1470.90 \pm 11.15 &
	{\databar{0.42}}	0.33 \pm 0.18 &
	{\databar{0.80}}	27.17 \pm 5.71 &
	{\databar{0.03}}	102 %
\\
CForest                                  & %
	{\hspace{-1.5cm}\databartwo{1.00}{0.00}\makebox[0pt][c]{\hspace{1cm}0 / 51}} &
	{\databar{1.00}}	450.03 \pm 0.02 &
	{\databar{0.35}}	1430.09 \pm 10.05 &
	{\databar{0.27}}	0.21 \pm 0.12 &
	{\databar{0.63}}	21.42 \pm 1.74 &
	{\databar{0.02}}	63 %
\\
EST                                      & %
	{\hspace{-1.5cm}\databartwo{1.00}{0.90}\makebox[0pt][c]{\hspace{1cm}46 / 51}} &
	{\databar{0.03}}	12.08 \pm 62.55 &
	{\databar{0.58}}	2361.56 \pm 348.57 &
	{\databar{0.54}}	0.43 \pm 0.11 &
	{\databar{0.95}}	32.24 \pm 4.87 &
	{\databar{0.23}}	689 %
\\
Informed RRT${}^*$                       & %
	{\hspace{-1.5cm}\databartwo{1.00}{0.00}\makebox[0pt][c]{\hspace{1cm}0 / 51}} &
	{\databar{1.00}}	450.01 \pm 0.01 &
	{\databar{0.35}}	1436.52 \pm 12.42 &
	{\databar{0.37}}	0.29 \pm 0.17 &
	{\databar{0.64}}	21.72 \pm 1.79 &
	{\databar{0.02}}	69 %
\\
KPIECE                                   & %
	{\hspace{-1.5cm}\databartwo{1.00}{0.86}\makebox[0pt][c]{\hspace{1cm}44 / 51}} &
	{\databar{0.00}}	0.73 \pm 2.13 &
	{\databar{1.00}}	4054.10 \pm 1042.07 &
	{\databar{0.36}}	0.28 \pm 0.10 &
	{\databar{0.96}}	32.44 \pm 3.93 &
	{\databar{0.47}}	1432 %
\\
PDST                                     & %
	{\hspace{-1.5cm}\databartwo{1.00}{0.63}\makebox[0pt][c]{\hspace{1cm}32 / 51}} &
	{\databar{0.14}}	64.53 \pm 137.23 &
	{\databar{0.58}}	2363.34 \pm 516.57 &
	{\databar{0.65}}	0.51 \pm 0.27 &
	{\databar{0.99}}	33.61 \pm 4.32 &
	{\databar{0.21}}	626 %
\\
PRM                                      & %
	{\hspace{-1.5cm}\databartwo{0.98}{0.55}\makebox[0pt][c]{\hspace{1cm}28 / 50}} &
	{\databar{1.00}}	450.22 \pm 0.19 &
	{\databar{0.37}}	1492.05 \pm 12.06 &
	{\databar{0.90}}	0.71 \pm 0.49 &
	{\databar{0.77}}	26.05 \pm 5.13 &
	{\databar{0.18}}	541 %
\\
PRM${}^*$                                & %
	{\hspace{-1.5cm}\databartwo{0.98}{0.59}\makebox[0pt][c]{\hspace{1cm}30 / 50}} &
	{\databar{1.00}}	450.17 \pm 0.13 &
	{\databar{0.36}}	1460.72 \pm 10.00 &
	{\databar{0.53}}	0.42 \pm 0.20 &
	{\databar{0.67}}	22.79 \pm 2.60 &
	{\databar{0.09}}	258 %
\\
RRT                                      & %
	{\hspace{-1.5cm}\databartwo{1.00}{0.90}\makebox[0pt][c]{\hspace{1cm}46 / 51}} &
	{\databar{0.00}}	0.64 \pm 1.91 &
	{\databar{0.46}}	1859.12 \pm 181.52 &
	{\databar{0.82}}	0.64 \pm 0.24 &
	{\databar{0.91}}	30.92 \pm 2.49 &
	{\databar{0.15}}	446 %
\\
RRT\#                                    & %
	{\hspace{-1.5cm}\databartwo{1.00}{0.18}\makebox[0pt][c]{\hspace{1cm}9 / 51}} &
	{\databar{1.00}}	450.37 \pm 1.03 &
	{\databar{0.35}}	1438.70 \pm 9.92 &
	{\databar{0.28}}	0.22 \pm 0.11 &
	{\databar{0.65}}	21.99 \pm 1.77 &
	{\databar{0.02}}	58 %
\\
RRT${}^*$                                & %
	{\hspace{-1.5cm}\databartwo{1.00}{0.04}\makebox[0pt][c]{\hspace{1cm}2 / 51}} &
	{\databar{1.00}}	450.01 \pm 0.01 &
	{\databar{0.35}}	1436.19 \pm 9.65 &
	{\databar{0.30}}	0.24 \pm 0.15 &
	{\databar{0.65}}	21.90 \pm 1.94 &
	{\databar{0.02}}	62 %
\\
SORRT${}^*$                              & %
	{\hspace{-1.5cm}\databartwo{1.00}{0.00}\makebox[0pt][c]{\hspace{1cm}0 / 51}} &
	{\databar{1.00}}	450.01 \pm 0.01 &
	{\databar{0.35}}	1435.67 \pm 8.22 &
	{\databar{0.35}}	0.28 \pm 0.15 &
	{\databar{0.64}}	21.68 \pm 1.80 &
	{\databar{0.02}}	53 %
\\
SPARS                                    & %
	{\hspace{-1.5cm}\databartwo{0.98}{0.69}\makebox[0pt][c]{\hspace{1cm}35 / 50}} &
	{\databar{1.00}}	451.52 \pm 2.06 &
	{\databar{0.48}}	1933.09 \pm 103.78 &
	{\databar{0.58}}	0.46 \pm 0.10 &
	{\databar{0.95}}	32.21 \pm 2.13 &
	{\databar{0.15}}	460 %
\\
SPARS2                                   & %
	{\hspace{-1.5cm}\databartwo{1.00}{0.67}\makebox[0pt][c]{\hspace{1cm}34 / 51}} &
	{\databar{1.00}}	450.01 \pm 0.01 &
	{\databar{0.38}}	1559.96 \pm 37.65 &
	{\databar{0.80}}	0.63 \pm 0.29 &
	{\databar{1.00}}	33.89 \pm 2.50 &
	{\databar{0.17}}	525 %
\\
SST                                      & %
	{\hspace{-1.5cm}\databartwo{1.00}{0.90}\makebox[0pt][c]{\hspace{1cm}46 / 51}} &
	{\databar{1.00}}	450.01 \pm 0.01 &
	{\databar{0.46}}	1855.10 \pm 110.14 &
	{\databar{1.00}}	0.79 \pm 0.19 &
	{\databar{0.75}}	25.39 \pm 3.97 &
	{\databar{1.00}}	3031 %
\\
\rowlabel{\textbf{Scenario: \texttt{Boston\_1\_1024}} (CC Reeds Shepp steering, \SI{45}{\minute} time limit)}
\\
BFMT                                     & %
	{\hspace{-1.5cm}\databartwo{0.92}{0.92}\makebox[0pt][c]{\hspace{1cm}47 / 47}} &
	{\databar{0.02}}	66.92 \pm 16.26 &
	{\databar{0.43}}	1526.63 \pm 22.58 &
	{\databar{0.57}}	0.41 \pm 0.21 &
	{\databar{0.80}}	28.06 \pm 5.28 &
	{\databar{0.11}}	264 %
\\
BIT${}^*$                                & %
	{\hspace{-1.5cm}\databartwo{0.51}{0.27}\makebox[0pt][c]{\hspace{1cm}14 / 26}} &
	{\databar{0.96}}	2604.03 \pm 484.81 &
	{\databar{0.42}}	1497.13 \pm 15.97 &
	{\databar{0.57}}	0.41 \pm 0.23 &
	{\databar{0.73}}	25.66 \pm 5.81 &
	{\databar{0.08}}	110 %
\\
CForest                                  & %
	{\hspace{-1.5cm}\databartwo{0.98}{0.00}\makebox[0pt][c]{\hspace{1cm}0 / 50}} &
	{\databar{0.00}}	1.50 \pm 1.73 &
	{\databar{0.40}}	1406.58 \pm 204.73 &
	{\databar{0.43}}	0.31 \pm 0.33 &
	{\databar{0.63}}	22.16 \pm 3.50 &
	{\databar{0.03}}	86 %
\\
EST                                      & %
	{\hspace{-1.5cm}\databartwo{0.98}{0.90}\makebox[0pt][c]{\hspace{1cm}46 / 50}} &
	{\databar{0.14}}	367.45 \pm 664.86 &
	{\databar{0.66}}	2344.28 \pm 306.80 &
	{\databar{0.60}}	0.44 \pm 0.10 &
	{\databar{0.93}}	32.59 \pm 4.32 &
	{\databar{0.54}}	1454 %
\\
Informed RRT${}^*$                       & %
	{\hspace{-1.5cm}\databartwo{0.84}{0.24}\makebox[0pt][c]{\hspace{1cm}12 / 43}} &
	{\databar{0.96}}	2583.17 \pm 530.75 &
	{\databar{0.39}}	1385.12 \pm 119.89 &
	{\databar{0.32}}	0.23 \pm 0.14 &
	{\databar{0.68}}	23.71 \pm 3.22 &
	{\databar{0.02}}	56 %
\\
KPIECE                                   & %
	{\hspace{-1.5cm}\databartwo{0.94}{0.84}\makebox[0pt][c]{\hspace{1cm}43 / 48}} &
	{\databar{0.09}}	252.00 \pm 563.32 &
	{\databar{1.00}}	3533.61 \pm 693.62 &
	{\databar{0.50}}	0.36 \pm 0.17 &
	{\databar{0.88}}	30.79 \pm 3.94 &
	{\databar{0.99}}	2476 %
\\
PDST                                     & %
	{\hspace{-1.5cm}\databartwo{0.88}{0.71}\makebox[0pt][c]{\hspace{1cm}36 / 45}} &
	{\databar{0.38}}	1034.87 \pm 1098.62 &
	{\databar{0.66}}	2326.69 \pm 647.37 &
	{\databar{0.67}}	0.48 \pm 0.24 &
	{\databar{0.96}}	33.61 \pm 4.43 &
	{\databar{0.24}}	569 %
\\
PRM                                      & %
	{\hspace{-1.5cm}\databartwo{0.76}{0.59}\makebox[0pt][c]{\hspace{1cm}30 / 39}} &
	{\databar{0.95}}	2571.17 \pm 555.87 &
	{\databar{0.44}}	1552.30 \pm 27.06 &
	{\databar{0.87}}	0.63 \pm 0.27 &
	{\databar{0.89}}	31.07 \pm 4.68 &
	{\databar{0.23}}	478 %
\\
PRM${}^*$                                & %
	{\hspace{-1.5cm}\databartwo{0.78}{0.59}\makebox[0pt][c]{\hspace{1cm}30 / 40}} &
	{\databar{0.95}}	2574.74 \pm 549.23 &
	{\databar{0.43}}	1520.56 \pm 35.76 &
	{\databar{0.73}}	0.53 \pm 0.19 &
	{\databar{0.81}}	28.35 \pm 5.37 &
	{\databar{0.18}}	378 %
\\
RRT                                      & %
	{\hspace{-1.5cm}\databartwo{0.98}{0.51}\makebox[0pt][c]{\hspace{1cm}26 / 50}} &
	{\databar{0.01}}	37.15 \pm 79.79 &
	{\databar{0.53}}	1857.30 \pm 184.10 &
	{\databar{0.76}}	0.55 \pm 0.21 &
	{\databar{0.89}}	31.21 \pm 3.42 &
	{\databar{0.19}}	508 %
\\
RRT\#                                    & %
	{\hspace{-1.5cm}\databartwo{0.80}{0.49}\makebox[0pt][c]{\hspace{1cm}25 / 41}} &
	{\databar{0.96}}	2579.91 \pm 542.55 &
	{\databar{0.41}}	1460.74 \pm 19.50 &
	{\databar{0.33}}	0.24 \pm 0.15 &
	{\databar{0.67}}	23.49 \pm 2.38 &
	{\databar{0.03}}	73 %
\\
RRT${}^*$                                & %
	{\hspace{-1.5cm}\databartwo{0.82}{0.27}\makebox[0pt][c]{\hspace{1cm}14 / 42}} &
	{\databar{0.96}}	2580.62 \pm 536.45 &
	{\databar{0.41}}	1451.53 \pm 12.30 &
	{\databar{0.34}}	0.25 \pm 0.15 &
	{\databar{0.64}}	22.45 \pm 2.21 &
	{\databar{0.03}}	61 %
\\
SORRT${}^*$                              & %
	{\hspace{-1.5cm}\databartwo{0.18}{0.18}\makebox[0pt][c]{\hspace{1cm}9 / 9}} &
	{\databar{0.90}}	2420.75 \pm 792.13 &
	{\databar{0.33}}	1162.47 \pm 53.04 &
	{\databar{0.25}}	0.18 \pm 0.05 &
	{\databar{0.75}}	26.27 \pm 4.22 &
	{\databar{0.03}}	13 %
\\
SPARS                                    & %
	{\hspace{-1.5cm}\databartwo{0.57}{0.55}\makebox[0pt][c]{\hspace{1cm}28 / 29}} &
	{\databar{1.00}}	2702.11 \pm 2.87 &
	{\databar{0.51}}	1798.43 \pm 95.15 &
	{\databar{0.71}}	0.52 \pm 0.11 &
	{\databar{1.00}}	34.98 \pm 1.37 &
	{\databar{0.13}}	292 %
\\
SPARS2                                   & %
	0 &
	N / A&
	N / A&
	N / A&
	N / A&
	N / A%
\\
SST                                      & %
	{\hspace{-1.5cm}\databartwo{0.71}{0.71}\makebox[0pt][c]{\hspace{1cm}36 / 36}} &
	{\databar{1.00}}	2700.05 \pm 0.02 &
	{\databar{0.54}}	1898.34 \pm 130.62 &
	{\databar{1.00}}	0.73 \pm 0.37 &
	{\databar{0.74}}	25.97 \pm 3.79 &
	{\databar{1.00}}	1937 %
\\
\rowlabel{\textbf{Scenario: \texttt{Boston\_1\_1024}} (POSQ steering, \SI{60}{\minute} time limit)}
\\
BFMT                                     & %
	0 &
	N / A&
	N / A&
	N / A&
	N / A&
	N / A%
\\
BIT${}^*$                                & %
	{\hspace{-1.5cm}\databartwo{0.18}{0.14}\makebox[0pt][c]{\hspace{1cm}7 / 9}} &
	{\databar{0.46}}	1652.64 \pm 755.06 &
	{\databar{0.71}}	3177.10 \pm 686.80 &
	{\databar{0.33}}	0.36 \pm 0.21 &
	{\databar{0.93}}	33.49 \pm 3.87 &
	{\databar{0.08}}	118 %
\\
CForest                                  & %
	{\hspace{-1.5cm}\databartwo{0.84}{0.57}\makebox[0pt][c]{\hspace{1cm}29 / 43}} &
	{\databar{0.61}}	2198.02 \pm 1502.31 &
	{\databar{0.26}}	1166.51 \pm 688.03 &
	{\databar{0.41}}	0.46 \pm 0.43 &
	{\databar{0.84}}	30.03 \pm 9.67 &
	{\databar{0.16}}	231 %
\\
EST                                      & %
	{\hspace{-1.5cm}\databartwo{0.82}{0.80}\makebox[0pt][c]{\hspace{1cm}41 / 42}} &
	{\databar{0.98}}	3526.31 \pm 473.09 &
	{\databar{0.13}}	593.43 \pm 404.04 &
	{\databar{0.77}}	0.86 \pm 0.37 &
	{\databar{0.70}}	25.10 \pm 6.77 &
	{\databar{0.15}}	210 %
\\
Informed RRT${}^*$                       & %
	{\hspace{-1.5cm}\databartwo{0.88}{0.80}\makebox[0pt][c]{\hspace{1cm}41 / 45}} &
	{\databar{0.94}}	3387.10 \pm 805.45 &
	{\databar{0.20}}	901.63 \pm 381.98 &
	{\databar{0.37}}	0.41 \pm 0.35 &
	{\databar{0.84}}	30.31 \pm 8.17 &
	{\databar{0.08}}	119 %
\\
KPIECE                                   & %
	{\hspace{-1.5cm}\databartwo{1.00}{0.22}\makebox[0pt][c]{\hspace{1cm}11 / 51}} &
	{\databar{0.04}}	133.32 \pm 326.82 &
	{\databar{1.00}}	4480.23 \pm 1099.00 &
	{\databar{0.41}}	0.46 \pm 0.26 &
	{\databar{0.78}}	27.99 \pm 4.77 &
	{\databar{1.00}}	1698 %
\\
PDST                                     & %
	{\hspace{-1.5cm}\databartwo{0.96}{0.06}\makebox[0pt][c]{\hspace{1cm}3 / 49}} &
	{\databar{0.36}}	1284.96 \pm 1387.17 &
	{\databar{0.49}}	2201.81 \pm 730.55 &
	{\databar{1.00}}	1.11 \pm 0.50 &
	{\databar{0.73}}	26.07 \pm 6.34 &
	{\databar{0.60}}	978 %
\\
PRM                                      & %
	{\hspace{-1.5cm}\databartwo{0.78}{0.33}\makebox[0pt][c]{\hspace{1cm}17 / 40}} &
	{\databar{0.72}}	2603.78 \pm 1420.00 &
	{\databar{0.42}}	1895.07 \pm 1730.32 &
	{\databar{0.79}}	0.88 \pm 0.49 &
	{\databar{0.60}}	21.51 \pm 7.13 &
	{\databar{0.92}}	1377 %
\\
PRM${}^*$                                & %
	{\hspace{-1.5cm}\databartwo{0.86}{0.14}\makebox[0pt][c]{\hspace{1cm}7 / 44}} &
	{\databar{0.47}}	1705.63 \pm 1575.08 &
	{\databar{0.46}}	2046.21 \pm 1108.60 &
	{\databar{0.69}}	0.76 \pm 0.46 &
	{\databar{0.58}}	20.81 \pm 6.51 &
	{\databar{0.77}}	1172 %
\\
RRT                                      & %
	{\hspace{-1.5cm}\databartwo{0.90}{0.82}\makebox[0pt][c]{\hspace{1cm}42 / 46}} &
	{\databar{0.91}}	3286.61 \pm 916.39 &
	{\databar{0.51}}	2305.96 \pm 989.04 &
	{\databar{0.46}}	0.52 \pm 0.37 &
	{\databar{1.00}}	35.95 \pm 6.92 &
	{\databar{0.28}}	429 %
\\
RRT\#                                    & %
	{\hspace{-1.5cm}\databartwo{0.88}{0.55}\makebox[0pt][c]{\hspace{1cm}28 / 45}} &
	{\databar{0.93}}	3331.73 \pm 866.07 &
	{\databar{0.22}}	988.79 \pm 479.13 &
	{\databar{0.41}}	0.45 \pm 0.42 &
	{\databar{0.83}}	29.68 \pm 7.86 &
	{\databar{0.07}}	107 %
\\
RRT${}^*$                                & %
	{\hspace{-1.5cm}\databartwo{0.94}{0.84}\makebox[0pt][c]{\hspace{1cm}43 / 48}} &
	{\databar{0.92}}	3323.11 \pm 852.09 &
	{\databar{0.22}}	967.71 \pm 465.57 &
	{\databar{0.38}}	0.42 \pm 0.34 &
	{\databar{0.87}}	31.39 \pm 7.74 &
	{\databar{0.07}}	117 %
\\
SORRT${}^*$                              & %
	{\hspace{-1.5cm}\databartwo{0.78}{0.71}\makebox[0pt][c]{\hspace{1cm}36 / 40}} &
	{\databar{1.00}}	3600.74 \pm 0.69 &
	{\databar{0.21}}	925.67 \pm 343.30 &
	{\databar{0.39}}	0.44 \pm 0.41 &
	{\databar{0.92}}	32.97 \pm 8.48 &
	{\databar{0.08}}	104 %
\\
SPARS                                    & %
	{\hspace{-1.5cm}\databartwo{0.51}{0.00}\makebox[0pt][c]{\hspace{1cm}0 / 26}} &
	{\databar{0.16}}	588.23 \pm 833.84 &
	{\databar{0.51}}	2298.03 \pm 393.28 &
	{\databar{0.33}}	0.36 \pm 0.19 &
	{\databar{0.85}}	30.52 \pm 3.87 &
	{\databar{0.17}}	210 %
\\
SPARS2                                   & %
	{\hspace{-1.5cm}\databartwo{0.90}{0.00}\makebox[0pt][c]{\hspace{1cm}0 / 46}} &
	{\databar{0.18}}	660.50 \pm 688.18 &
	{\databar{0.56}}	2528.72 \pm 489.22 &
	{\databar{0.56}}	0.62 \pm 0.43 &
	{\databar{0.77}}	27.62 \pm 4.33 &
	{\databar{0.32}}	527 %
\\
SST                                      & %
	0 &
	N / A&
	N / A&
	N / A&
	N / A&
	N / A%
\\
\rowlabel{\textbf{Scenario: \texttt{Boston\_1\_1024}} (Dubins steering, \SI{30}{\minute} time limit)}
\\
BFMT                                     & %
	{\hspace{-1.5cm}\databartwo{0.98}{0.02}\makebox[0pt][c]{\hspace{1cm}1 / 50}} &
	{\databar{0.04}}	75.20 \pm 270.19 &
	{\databar{0.40}}	1650.78 \pm 59.43 &
	{\databar{0.78}}	0.25 \pm 0.00 &
	{\databar{0.90}}	31.92 \pm 3.01 &
	{\databar{0.15}}	303 %
\\
BIT${}^*$                                & %
	{\hspace{-1.5cm}\databartwo{0.78}{0.24}\makebox[0pt][c]{\hspace{1cm}12 / 40}} &
	{\databar{1.00}}	1800.32 \pm 1.11 &
	{\databar{0.36}}	1502.68 \pm 35.51 &
	{\databar{0.77}}	0.25 \pm 0.01 &
	{\databar{0.79}}	28.21 \pm 6.06 &
	{\databar{0.05}}	90 %
\\
CForest                                  & %
	{\hspace{-1.5cm}\databartwo{0.94}{0.02}\makebox[0pt][c]{\hspace{1cm}1 / 48}} &
	{\databar{1.00}}	1800.13 \pm 0.07 &
	{\databar{0.35}}	1446.68 \pm 20.22 &
	{\databar{0.70}}	0.22 \pm 0.05 &
	{\databar{0.61}}	21.73 \pm 1.83 &
	{\databar{0.01}}	24 %
\\
EST                                      & %
	{\hspace{-1.5cm}\databartwo{1.00}{0.75}\makebox[0pt][c]{\hspace{1cm}38 / 51}} &
	{\databar{0.14}}	249.30 \pm 586.65 &
	{\databar{0.57}}	2356.67 \pm 360.25 &
	{\databar{0.78}}	0.25 \pm 0.00 &
	{\databar{0.92}}	32.89 \pm 3.88 &
	{\databar{0.52}}	1066 %
\\
Informed RRT${}^*$                       & %
	{\hspace{-1.5cm}\databartwo{0.90}{0.16}\makebox[0pt][c]{\hspace{1cm}8 / 46}} &
	{\databar{1.00}}	1800.07 \pm 0.07 &
	{\databar{0.35}}	1453.39 \pm 57.24 &
	{\databar{0.73}}	0.23 \pm 0.03 &
	{\databar{0.68}}	24.18 \pm 4.40 &
	{\databar{0.02}}	39 %
\\
KPIECE                                   & %
	{\hspace{-1.5cm}\databartwo{1.00}{0.73}\makebox[0pt][c]{\hspace{1cm}37 / 51}} &
	{\databar{0.01}}	18.56 \pm 83.68 &
	{\databar{1.00}}	4117.14 \pm 958.46 &
	{\databar{0.75}}	0.24 \pm 0.01 &
	{\databar{1.00}}	35.66 \pm 4.56 &
	{\databar{1.00}}	2052 %
\\
PDST                                     & %
	{\hspace{-1.5cm}\databartwo{0.82}{0.49}\makebox[0pt][c]{\hspace{1cm}25 / 42}} &
	{\databar{0.03}}	50.56 \pm 166.76 &
	{\databar{0.59}}	2412.63 \pm 578.49 &
	{\databar{0.76}}	0.24 \pm 0.01 &
	{\databar{0.92}}	32.86 \pm 4.03 &
	{\databar{0.19}}	327 %
\\
PRM                                      & %
	{\hspace{-1.5cm}\databartwo{0.92}{0.00}\makebox[0pt][c]{\hspace{1cm}0 / 47}} &
	{\databar{1.00}}	1800.13 \pm 0.11 &
	{\databar{0.39}}	1620.43 \pm 75.43 &
	{\databar{0.78}}	0.25 \pm 0.00 &
	{\databar{0.90}}	31.94 \pm 5.27 &
	{\databar{0.13}}	262 %
\\
PRM${}^*$                                & %
	{\hspace{-1.5cm}\databartwo{0.86}{0.12}\makebox[0pt][c]{\hspace{1cm}6 / 44}} &
	{\databar{1.00}}	1800.22 \pm 0.27 &
	{\databar{0.39}}	1600.53 \pm 82.10 &
	{\databar{0.78}}	0.25 \pm 0.00 &
	{\databar{0.86}}	30.65 \pm 5.99 &
	{\databar{0.11}}	216 %
\\
RRT                                      & %
	{\hspace{-1.5cm}\databartwo{1.00}{0.71}\makebox[0pt][c]{\hspace{1cm}36 / 51}} &
	{\databar{0.01}}	16.19 \pm 58.95 &
	{\databar{0.53}}	2175.83 \pm 303.94 &
	{\databar{0.77}}	0.25 \pm 0.01 &
	{\databar{0.94}}	33.57 \pm 4.17 &
	{\databar{0.25}}	511 %
\\
RRT\#                                    & %
	{\hspace{-1.5cm}\databartwo{0.90}{0.00}\makebox[0pt][c]{\hspace{1cm}0 / 46}} &
	{\databar{1.00}}	1800.15 \pm 0.53 &
	{\databar{0.35}}	1433.26 \pm 67.70 &
	{\databar{0.75}}	0.24 \pm 0.02 &
	{\databar{0.61}}	21.88 \pm 2.21 &
	{\databar{0.01}}	22 %
\\
RRT${}^*$                                & %
	{\hspace{-1.5cm}\databartwo{0.96}{0.02}\makebox[0pt][c]{\hspace{1cm}1 / 49}} &
	{\databar{1.00}}	1800.06 \pm 0.05 &
	{\databar{0.34}}	1417.91 \pm 95.88 &
	{\databar{0.73}}	0.24 \pm 0.03 &
	{\databar{0.63}}	22.52 \pm 1.71 &
	{\databar{0.02}}	31 %
\\
SORRT${}^*$                              & %
	{\hspace{-1.5cm}\databartwo{0.90}{0.20}\makebox[0pt][c]{\hspace{1cm}10 / 46}} &
	{\databar{1.00}}	1800.06 \pm 0.06 &
	{\databar{0.35}}	1452.59 \pm 57.02 &
	{\databar{0.74}}	0.24 \pm 0.02 &
	{\databar{0.70}}	24.93 \pm 4.52 &
	{\databar{0.02}}	37 %
\\
SPARS                                    & %
	{\hspace{-1.5cm}\databartwo{0.76}{0.04}\makebox[0pt][c]{\hspace{1cm}2 / 39}} &
	{\databar{1.00}}	1805.31 \pm 4.54 &
	{\databar{0.45}}	1842.57 \pm 105.67 &
	{\databar{0.77}}	0.25 \pm 0.00 &
	{\databar{0.86}}	30.57 \pm 1.37 &
	{\databar{0.10}}	198 %
\\
SPARS2                                   & %
	{\hspace{-1.5cm}\databartwo{0.82}{0.04}\makebox[0pt][c]{\hspace{1cm}2 / 42}} &
	{\databar{1.00}}	1800.01 \pm 0.01 &
	{\databar{0.42}}	1721.66 \pm 63.07 &
	{\databar{0.78}}	0.25 \pm 0.00 &
	{\databar{0.90}}	32.25 \pm 3.90 &
	{\databar{0.18}}	371 %
\\
SST                                      & %
	{\hspace{-1.5cm}\databartwo{0.88}{0.82}\makebox[0pt][c]{\hspace{1cm}42 / 45}} &
	{\databar{1.00}}	1800.05 \pm 0.04 &
	{\databar{0.45}}	1838.02 \pm 132.55 &
	{\databar{1.00}}	0.32 \pm 0.25 &
	{\databar{0.79}}	28.23 \pm 4.24 &
	{\databar{0.70}}	1263 %
\\
\bottomrule
\end{tabular}
    }
    \caption{Planning statistics using different steer functions from the \texttt{Boston\_1\_1024} scenario from the Moving AI benchmark.}
    \label{tab:moving_ai_boston}
\end{table*}

\begin{figure*}[htp]
    \centering
    \includegraphics[width=\textwidth]{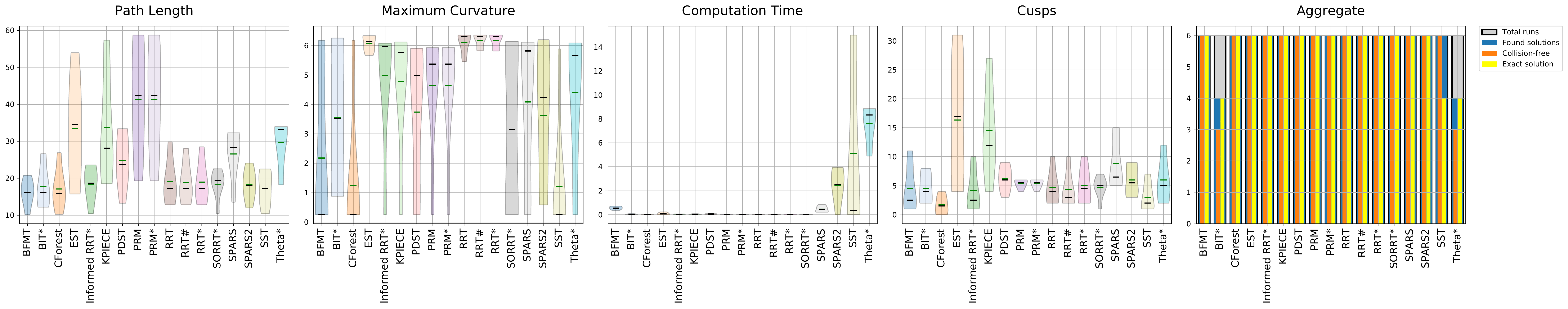}\\
    \includegraphics[width=\textwidth]{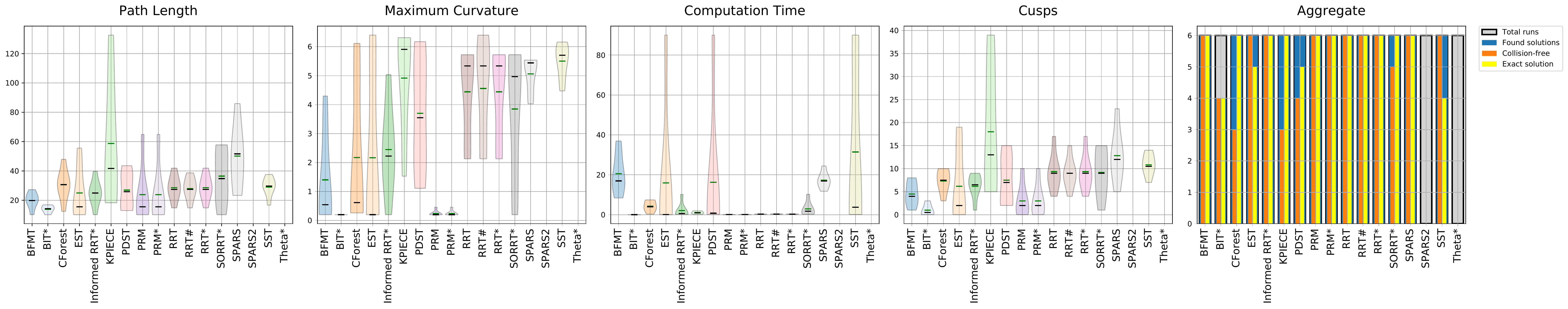}\\
    \includegraphics[width=\textwidth]{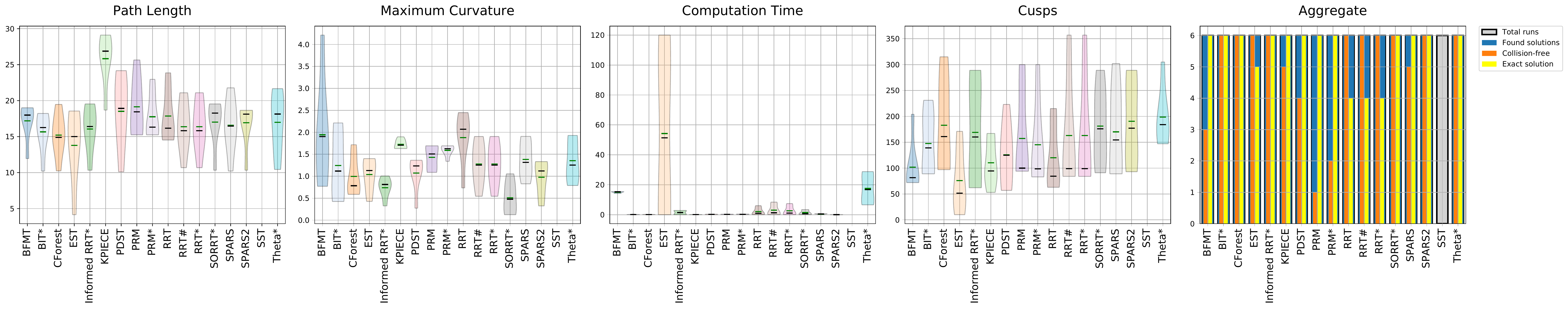}\\
    \includegraphics[width=\textwidth]{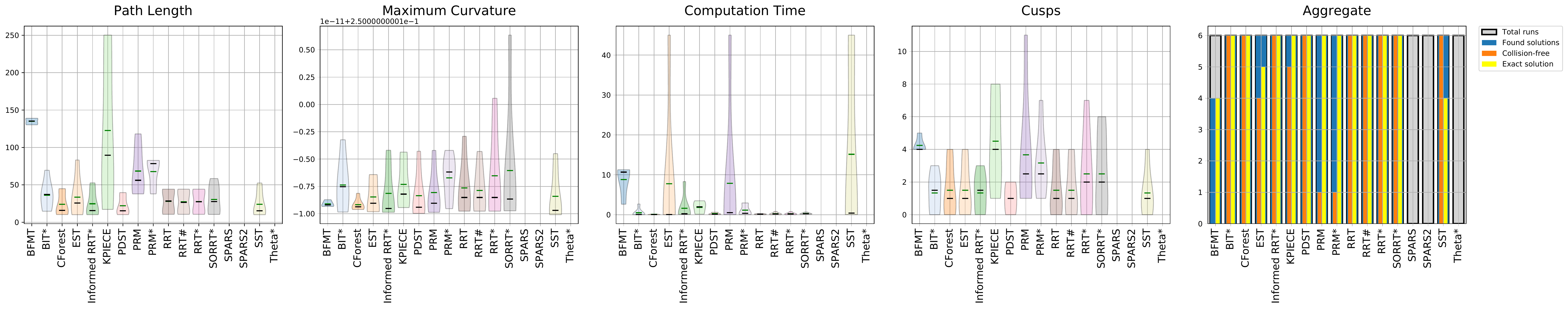}
    \caption{Statistics for the \emph{parking2} scenarios. First row: Reeds Shepp steering, second row: CC Reeds Shepp steering, third row: POSQ steering, fourth row: Dubins steering.}
    \label{fig:parking2_stats}
\end{figure*}

\begin{figure*}[htp]
    \centering
    \includegraphics[width=\textwidth]{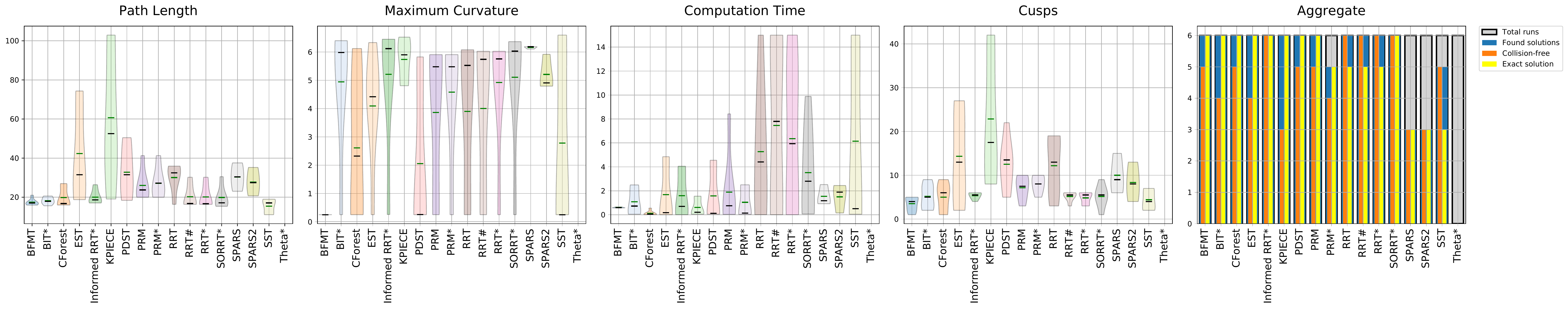}\\
    \includegraphics[width=\textwidth]{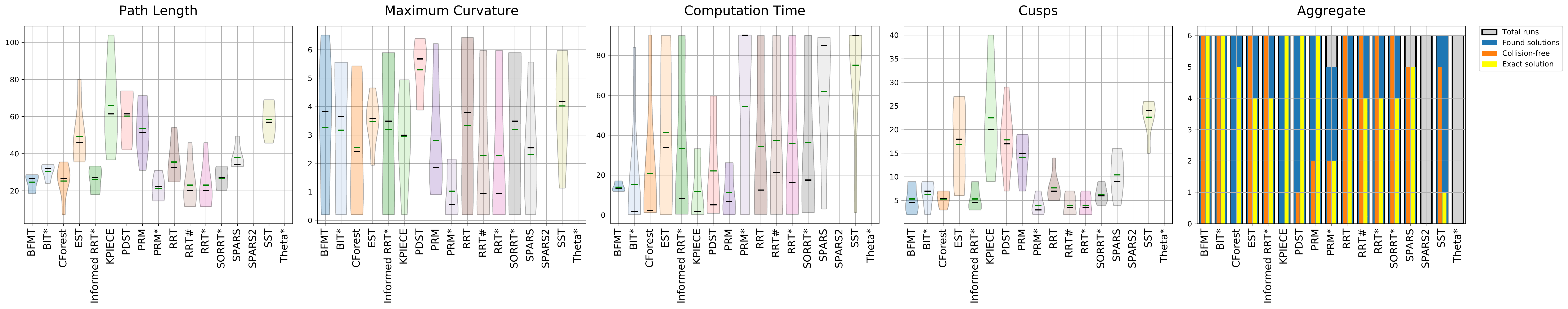}\\
    \includegraphics[width=\textwidth]{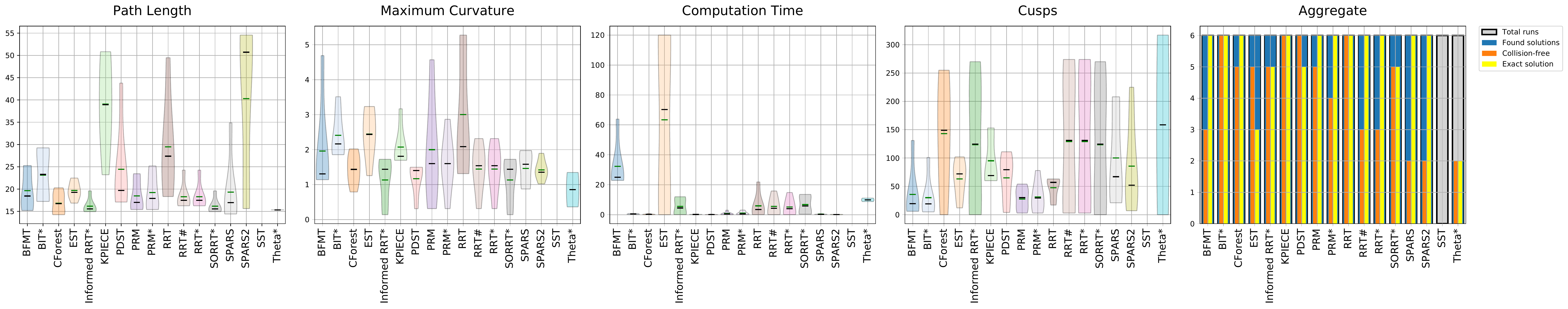}\\
    \includegraphics[width=\textwidth]{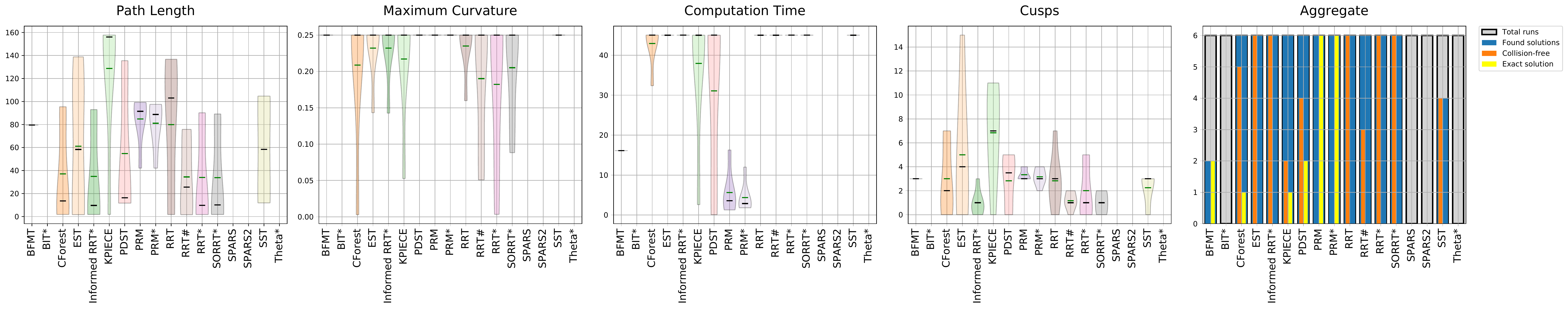}
    \caption{Statistics for the \emph{parking3} scenarios. First row: Reeds Shepp steering, second row: CC Reeds Shepp steering, third row: POSQ steering, fourth row: Dubins steering.}
    \label{fig:parking3_stats}
\end{figure*}

\fi

\end{document}